\documentclass[twoside]{article}

\usepackage[accepted]{aistats2025}

\usepackage{algorithmicx}
\usepackage{algpseudocode}
\usepackage{etoolbox}
\usepackage{cleveref}

\usepackage{multirow}
\usepackage{float}
\usepackage{overpic}

\usepackage{MnSymbol}
\usepackage{amsthm}
\usepackage{amsmath}

\usepackage{scalerel}
\usepackage{xcolor}

\usepackage{algorithm}
\usepackage{xpatch}
\usepackage{amsfonts}   

\usepackage{url}            
\usepackage{booktabs}
\usepackage{subfigure} 
\usepackage{enumitem}

\newtheorem{lemma}{Lemma}
\newtheorem{theorem}{Theorem}

\usepackage{algorithm}
\usepackage{algpseudocode}

\algrenewcommand\algorithmicrequire{\textbf{Input:}}
\algrenewcommand\algorithmicensure{\textbf{Output:}}
\usepackage[round]{natbib}

\newcommand{\ttop}{^{\top}}
\newcommand{\tr}{\textup{tr}}
\newcommand{\E}{\mathbf{E}}
\newcommand{\var}{\operatorname{var}}
\newcommand{\cov}{\textup{cov}}

\newcommand{\ts}{\textstyle}
\renewcommand{\P}{\mathbf{P}}
\newcommand{\R}{\mathbb{R}}

\newcommand{\sL}{L}
\newcommand{\sG}{G}
\newcommand{\sE}{E}
\newcommand{\sV}{V}
\newcommand{\sQ}{Q}
\newcommand{\sC}{C}
\newcommand{\sW}{W}
\newcommand{\sN}{N}
\newcommand{\sB}{B}

\newcommand{\sCc}{\mathcal{C}}

\begin{document}

%

%

\twocolumn[

\aistatstitle{Empirical Error Estimates for Graph Sparsification}

\aistatsauthor{ Siyao Wang \And Miles E. Lopes }

\aistatsaddress{ University of California, Davis \And University of California, Davis } ]

\begin{abstract}
Graph sparsification is a well-established technique for accelerating graph-based learning algorithms, which uses edge sampling to approximate dense graphs with sparse ones. Because the sparsification error is random and unknown, users must contend with uncertainty about the reliability of downstream computations. Although it is possible for users to obtain conceptual guidance from theoretical error bounds in the literature, such results are typically impractical at a numerical level. Taking an alternative approach, we propose to address these issues from a data-driven perspective by computing \emph{empirical error estimates}. The proposed error estimates are highly versatile, and we demonstrate this in four use cases: Laplacian matrix approximation, graph cut queries, graph-structured regression, and spectral clustering. Moreover, we provide two theoretical guarantees for the error estimates, and explain why the cost of computing them is manageable in comparison to the overall cost of a typical graph sparsification workflow.
\end{abstract}

\section{INTRODUCTION}\label{sec:introduction}

The scalability of graph-based algorithms in machine learning is often limited in applications that involve dense graphs with very large numbers of edges.
For this reason, \emph{graph sparsification} has become a well-established acceleration technique, which speeds up computations by replacing dense graphs with sparse approximations \citep{Benczur:1996,Spielman:2011}. Furthermore, there are myriad applications that illustrate the popularity and flexibility of graph sparsification, such as graph partitioning ~\citep{Kelner:2014,chen2022maximum}, clustering~\citep{Woodruff:2016,Agarwal:2022}, solving linear systems~\citep{Spielman:2004,Sidford:2021}, 
graph-structured regression~\citep{sadhanala2016graph,Calandriello:2018},
and deep learning~\citep{GraphSAGE:2017,Zeng:2020,Zheng:2020}.

Graph sparsification is commonly implemented in a randomized manner via edge sampling, which confronts the user with substantial uncertainty: The error produced by the sampling is both random and unknown, which raises doubts about the accuracy of results that rely on the sparsified graph. This uncertainty can also lead to less efficient computation, as users are inclined to ``hedge their bets'' with conservatively large sample sizes---undermining the benefit 
of sparsification.

To deal with these issues, it is necessary to estimate the error created by sparsification. Indeed, error estimates not only provide a gauge for the reliability of computations, but also help to avoid the inefficiency of excessive sampling. For example, error estimates can enable \emph{incremental refinement}, which involves estimating the error of an inexpensive preliminary sparsified graph, and then sampling extra edges as needed until the estimated error falls below a target threshold. Hence, such an approach can help users to sample just enough edges to suit their purpose.

Up to now, the literature has generally addressed sparsification error from a theoretical standpoint. For instance, this is often done by deriving theoretical bounds on the runtimes of sparsified graph algorithms as a function of the error. However, such results tend to be inherently conservative, as they are often designed to hold uniformly over a large class of possible inputs. Making matters worse, such results typically involve unknown parameters or unspecified constants. Consequently, it can be infeasible to use theoretical error bounds in a way that is practical on a problem-specific basis.

Based on the issues just discussed, we propose to address error estimation from a more data-driven perspective---by using bootstrap methods to compute  \emph{empirical error estimates} that only rely on the information acquired in the edge sampling process. As a result, this approach delivers error estimates that are adapted to the particular inputs at hand, avoiding the drawbacks of worst-case error analysis.

Our main contributions are summarized as follows: \textbf{(1)} To the best of our knowledge, this paper is the first to systematically develop empirical error estimates for graph sparsification.  \textbf{(2)} We illustrate the flexibility of our error estimates in four use cases, including \emph{Laplacian matrix approximation}, \emph{graph cut queries}, \emph{graph-structured regression}, and \emph{spectral clustering}. All of these examples are supported by numerical experiments under a variety of conditions. \textbf{(3)} In two different contexts, we prove that the error estimates perform correctly in the limit of large problem sizes. Because we allow the number of graph vertices and edges to diverge simultaneously with the number of sampled edges,
our theoretical results require in-depth analyses based on \emph{high-dimensional central limit theorems}.

\noindent\textbf{Preliminaries.} We consider weighted undirected graphs $\sG=(\sV,\sE,w)$, with vertex set $\sV= \{1,\ldots,n\}$, edge set $\sE \subset  \{\{i,j\}| i,j \in \sV, i\neq j\}$, and weight function \smash{$w:\sE\to[0,\infty)$.} The Laplacian matrix $\sL \in \R^{n\times n}$ of the graph $\sG$ is defined by $\sL_{ii}=\sum_{\{i,l\}\in E} w(i,l)$,  and $\sL_{ij}=-w(i,j)$ if $i\neq j$. Equivalently, if $\Delta_{e}\in\R^{n\times n}$ denotes the symmetric rank-1 matrix associated to an edge $e=\{i,j\}$ such that  $x\ttop \Delta_e x=(x_i-x_j)^2$ for all $x \in \R^n$, then $\sL$ can be represented as
\begin{equation}
    \sL \ = \ \sum_{e \in E} w(e) \Delta_{e}.
\end{equation}
With regard to graph sparsification, we focus on settings where the sparsified graph $\hat \sG=(\sV, \hat \sE, \hat w)$ is obtained by sampling $\sN$ edges from $\sG$ in an i.i.d.~manner. On each sample, an edge $e$ appears with a probability denoted by $p(e)$, and the sampled edge is incorporated into $\hat \sG$ with weight $w(e)/(\sN p(e))$. (If an edge is sampled more than once, then the weights are added.) There are many choices of interest for the sampling probabilities, such as edge-weight sampling with $p(e)\propto w(e)$, and effective-resistance sampling with $p(e)\propto w(e)\tr(\sL^+\Delta_e)$, where $\sL^+$ is the Moore-Penrose inverse of $\sL$~\citep{srivastava2011}. Importantly, \emph{our proposed algorithms can be applied in practice without restricting the user's choice of sampling probabilities for generating the sparsified graph}.

The Laplacian matrix associated with $\hat \sG$ is denoted by $\hat \sL$, and is referred to as a sparsified Laplacian. It should be emphasized that $\hat \sL$ is a random matrix that can be interpreted as a sample average in the following way: If we define the collection of rank-1 matrices $\mathcal{Q} = \{(w(e)/p(e))\Delta_{e}| e \in \sE\}$, and let $\sQ_1,\dots,\sQ_N\in\R^{n\times n}$ be i.i.d.~samples from $\mathcal{Q}$ such that $(w(e)/p(e))\Delta_{e}$ appears on each draw with probability $p(e)$, then $\hat \sL$ can be represented as
\begin{equation}
\small
    \hat \sL \ = \ \frac{1}{N}\sum_{i=1}^N \sQ_i. 
\end{equation}
Furthermore, it can be checked that $\E(\sQ_1)=\sL$, ensuring unbiasedness, $\E(\hat \sL)=\sL$.

\noindent\textbf{Problem setting.} Graph sparsification is often intended for settings where $\sG$ is so large or dense that accessing it incurs high communication costs, and only $\hat\sG$ or $\hat\sL$ can be stored in fast memory. For this reason, our error estimates will only rely on the sampled matrices $\sQ_1,\dots,\sQ_N$, and will not require access to $\sG$ or $\sL$. Likewise, we view quantities depending on $\sG$ and $\sL$ as fixed unknown parameters. 

\noindent \textbf{Error functionals.} To measure how well $\hat \sL$ approximates $\sL$, we will consider a variety of scalar-valued error functionals, denoted $\psi(\hat \sL,\sL)$. For example, $\psi$ could correspond to error in the Frobenius norm $\psi(\hat \sL, \sL)=\|\hat \sL-\sL\|_F$ or operator (spectral) norm $\psi(\hat \sL,\sL)=\|\hat \sL-\sL\|_{\textup{op}}$. 
More generally, users can select $\psi$ to suit their preferred notion of error in specific applications, as illustrated in Section~\ref{sec:method:functional}.

For a given choice of $\psi$, our goal is to estimate the tightest possible upper bound on the \emph{unobserved} random variable $\psi(\hat \sL,\sL)$ that holds with a prescribed probability, say $1-\alpha$ with $\alpha\in(0,1)$. Although this optimal bound is unknown, it can be defined precisely as the $(1-\alpha)$-quantile of $\psi(\hat \sL,\sL)$,  denoted
\begin{equation}
    q_{1-\alpha} \ = \ \inf\Big\{t\in \R\,\Big|\, \P(\psi\big(\hat \sL,\sL)\leq t\big) \, \geq \, 1-\alpha\Big\}.
\end{equation}
Accordingly, we aim to develop algorithms that can compute estimates $\hat q_{1-\alpha}$ of $q_{1-\alpha}$. Furthermore, these estimates are intended to perform well in three respects: (I) They should be flexible enough to handle many choices of $\psi$. (II) They should nearly match $q_{1-\alpha}$, so that the event $\{\psi\big(\hat \sL,\sL)\leq \hat{q}_{1-\alpha}\}$ holds with probability close to $1-\alpha$. (III) They should be affordable to compute, so that the extra step of error estimation only modestly increases the overall cost of the user's workflow. In Sections~\ref{sec:method}-\ref{sec:expt}, we demonstrate that all three desiderata are achieved by our proposed estimates.

\noindent \textbf{Simultaneous confidence intervals.} In addition to measuring error through various choices of $\psi(\hat \sL,\sL)$, it is natural in many applications to develop simultaneous confidence intervals (CIs) for unknown quantities depending on $\sL$. Denoting these quantities as $\theta_1(\sL),\dots,\theta_k(\sL)$, some important examples include graph cut values, and eigenvalues of $\sL$ that are relevant in spectral clustering. (See Section~\ref{sec:method:CI} and Appendix~\ref{sec:clustering}.) In such contexts, our approach can be extended to construct CIs that simultaneously cover $\theta_1(\sL),\dots,\theta_k(\sL)$, while enjoying properties analogous to (I)-(III) above.

\noindent\textbf{Related work and novelty.} 
Over the last 15 years, randomized approximation algorithms have been widely adopted in many applications of machine learning and large-scale computing~\citep{Cormode:2011,Mahoney:2011,Woodruff:2014,Tropp:2020,RASC}. However, the research on empirical error estimation for these algorithms
is still at a relatively early stage, and it has only just begun to accelerate within the last few years. A notable theme in this recent work is that statistical resampling techniques---such as the bootstrap, jackknife, and subsampling---have 
proven to be key ingredients in estimating the errors of many types of randomized algorithms.
Examples of randomized algorithms for which statistical error estimation methods have been developed include low-rank approximation~\citep{Epperly}, regression~\citep{lopes2018error,lopes2020measuring,Dobriban:2023}, matrix multiplication~\citep{lopes2019bootstrap,lopes2023bootstrapping}, trace estimation~\citep[][]{Tropp:2020}, Fourier features~\citep{Lopes:2023}, and PCA~\citep{lunde2021bootstrapping,Lopes:SVD,Dobriban:2024}.

Within this growing line of research, the current paper is novel in several ways. Most importantly, our work is the first to specifically target graph sparsification, \emph{which demands methodology and theory that are both new}.
At a more technical level, our work is also differentiated in the way that we adapt resampling methods to our setting. In particular, for certain applications, we leverage a specialized type of resampling known as a ``double bootstrap''~\citep{Chernick:2011,Hall:2013}. In many classical statistical problems, it is known that a double bootstrap can substantially improve upon more basic bootstrap methods,
but up to now, \emph{its advantages have not been considered in the contemporary line of work on error estimation for randomized algorithms}. Our choice to use this approach in Section~\ref{sec:method:functional} is based on practical necessity, as we found that simpler resampling techniques led to unsatisfactory error estimates. Lastly, it is worth clarifying that although the enhancements provided by double bootstrapping do require a more technical implementation,
the computational cost is not an obstacle in modern computing environments that are relevant to graph sparsification, as explained in Section~\ref{sec:computation}.

\noindent\textbf{Notation and terminology.} If $A$ is a finite set of real numbers and $\alpha\in (0,1)$, then the \emph{empirical $(1-\alpha)$-quantile} of $A$ is denoted as $\textup{quantile}(A;1-\alpha)$\label{page:q}, which is the smallest $a_0\in A$ such that $|\{a\in A: a\leq a_0\}|/|A| \geq 1-\alpha$, where $|\cdot|$ refers to cardinality. If $q\geq 1$, then the $\ell_q$ norm of $v\in\R^d$ is  $\|v\|_q = (\sum_{j=1}^d |v_{j}|^q)^{1/q}$, and $\|v\|_{\infty}=\max_{1\leq j\leq d}|v_j|$.
If $M$ is a symmetric real matrix, then $\lambda_1(M)\leq \lambda_2(M)\leq \cdots$ refer to the sorted eigenvalues.  To refer to the multinomial distribution based on tossing $\sN$ balls into $\sN$ bins with probabilities $p_1,\dots,p_N$, we write Mult.$(\sN;p_1,\dots,p_N)$.

\section{METHODS}\label{sec:method}

In this section, we present two algorithms and explain how they quantify the errors that arise from $\hat \sL$ in several tasks. Section~\ref{sec:method:functional} focuses on quantile estimates for error functionals $\psi(\hat \sL,\sL)$, which can be used in Laplacian matrix approximation and graph-structured regression. Section~\ref{sec:method:CI} develops simultaneous CIs for the values of graph cuts and eigenvalues of $\sL$.

\subsection{Error functionals}\label{sec:method:functional}
Recall that $\hat \sL$ can be represented as $\hat \sL=\frac{1}{N}\sum_{i=1}^N \sQ_i$, where $\sQ_1,\dots,\sQ_N$ are i.i.d.~random matrices such that $\E(\hat\sL)=\E(\sQ_1)=\sL$.
Letting $\psi(\hat\sL,\sL)$ denote a generic error functional, and letting $q_{1-\alpha}$ denote its $(1-\alpha)$-quantile, \emph{our goal is to compute an estimate $\hat q_{1-\alpha}$ using only knowledge of $\sQ_1,\dots,\sQ_N$}.
To develop the estimate, a bootstrap approach relies on a mechanism for generating approximate samples of the random variable $\psi(\hat\sL,\sL)$, so that $\hat q_{1-\alpha}$ can be constructed as the empirical $(1-\alpha)$-quantile of those approximate samples. But it turns out that even for simple choices of $\psi$, this approach can sometimes produce poor estimates of $q_{1-\alpha}$. In such situations, it is known in the bootstrap literature that better performance can often be achieved by using approximate samples of a suitably standardized version of $\psi(\hat\sL,\sL)$~\citep[][Ch.3]{Hall:2013}. For this reason, we aim to generate approximate samples of the random variable $\zeta=(\psi(\hat\sL,\sL)-\hat\mu)/\hat\sigma$, where $\hat\mu$ and $\hat\sigma^2$ denote estimates of $\mu=\E(\psi(\hat\sL,\sL))$ and $\sigma^2=\var(\psi(\hat\sL,\sL))$ that will be defined later. Specifically, if the approximate samples are denoted $\zeta_1^*,\dots,\zeta_B^*$, then they can be used to define the estimate $\hat q_{1-\alpha}=\textup{quantile}(\hat\mu+\hat\sigma \zeta_1^*,\dots,\hat\mu+\hat\sigma \zeta_B^*;1-\alpha)$.

The main ideas for generating approximate samples of $\zeta$ are as follows. Since $\zeta$ can be viewed a function of $\sQ_1,\dots,\sQ_N$,  denoted $\zeta=\varphi(\sQ_1,\dots,\sQ_N)$, the standard bootstrap approach would be to randomly sample matrices $\sQ_1^*,\dots,\sQ_N^*$ with replacement from $\sQ_1,\dots,\sQ_N$, and then define $\zeta^*=\varphi(\sQ_1^*,\dots,\sQ_N^*)$ as an approximate sample of $\zeta$. However, this is not directly applicable in our context with a generic choice of $\psi$, because there are generally no explicit formulas for computing  $\hat\mu$ and $\hat\sigma$ in terms of $\sQ_1,\dots,\sQ_N$, and hence, there are generally no explicit formulas for computing $\varphi(\sQ_1,\dots,\sQ_N)$. Nevertheless, an approximation $\hat\varphi$ to the function $\varphi$ can also be developed via bootstrap sampling, and approximate samples of $\zeta$ can be defined as $\zeta^*=\hat\varphi(\sQ_1^*,\dots,\sQ_N^*)$.

From an algorithmic standpoint, this way of defining $\zeta^*$ is more intricate than it might appear at first sight. A particularly important point is that computing $\zeta^*$ actually involves a ``second level'' of bootstrap sampling. This is because the quantity $\hat\varphi(\sQ_1^*,\dots,\sQ_N^*)$ will be computed by sampling from the (already resampled) matrices $\sQ_1^*,\dots,\sQ_N^*$ with replacement. Another consideration is that even though it is natural to think about the proposed method in terms of sampling from sets of matrices with replacement, it is possible to implement this more efficiently by reweighting matrices with coefficients drawn from certain multinomial distributions, as shown in Algorithm~1 below.

\smash{\textbf{Algorithm \!1} \!(Quantile estimate for error functionals)}\\
\vspace{-0.2cm}
\hrule
\textbf{Input:} Number of bootstrap samples $\sB\geq 1$, a number $\alpha\in(0,1)$, and the matrices $\hat \sL, \sQ_1, \ldots, \sQ_N$.

\textbf{for $b=1,\dots,\sB$ in parallel do:}
\vspace{-0.2cm}
\begin{itemize}[leftmargin=*]
    \item Generate $(\sW_1^*,\dots,\sW_N^*)\sim\text{Mult.}(\sN; \ts \frac{1}{N},\dots,\frac{1}{N}).$
    \item Compute  $\varepsilon_b^*=\psi(\hat\sL^{*},\hat \sL)$, where $\hat\sL^{*}=\frac{1}{N}\sum_{i=1}^N \sW_i^* \sQ_i$.
    \item[]\textbf{for $b'=1,\dots,B$ in parallel do:}
    \begin{itemize}[leftmargin=*]
    \item[$\bullet$] \smash{\!Generate\!
$(\sW_1^{**},\dots,\sW_N^{**})\!\sim\!\textup{Mult.}(\sN;\!\ts \frac{W_1^*}{N},\dots,\!\frac{W_N^*}{N}).$}\\[-0.2cm]
\item[$\bullet$] \smash{\!Compute $\varepsilon_{b'}^{**}\!\!=\psi(\hat \sL^{**}\!\!\!,\hat\sL^{*})$, where $\hat \sL^{**}\!\!\!=\!\frac{1}{N}\!\sum_{i=1}^N \!\sW_i^{**}\sQ_i$.}
\end{itemize}
\item[]\textbf{end for}
\item Compute $\hat\mu_b^*=\frac{1}{B}\sum_{b'=1}^{B} \varepsilon_{b'}^{**}$ as well as
\begin{align*}
\small
\hat\sigma_b^*&=\sqrt{\ts \frac{1}{B}\sum_{b'=1}^{B} (\varepsilon_{b'}^{**}-\hat\mu_b^*)^2},\\
\zeta_b^* & = \frac{1}{\hat\sigma_b^*}(\varepsilon_b^*-\hat\mu_b^*). \text{ \ \ \ \ (If $\hat\sigma_b^*=0$, put $\zeta_b^*=0$.)
}
\end{align*}
\end{itemize}
\vspace{-0.3cm}
\textbf{end for}

Compute $\hat\mu=\frac{1}{B}\sum_{b=1}^B \varepsilon_b^*$ \,and\, $\hat\sigma=\sqrt{\frac{1}{B}\sum_{b=1}^B (\varepsilon_b^*-\hat\mu)^2}$.

\textbf{Output:} $\hat q_{1-\alpha}=\textup{quantile}(\hat\mu+\hat\sigma \zeta_1^*,\dots,\hat\mu+\hat\sigma \zeta_B^*; 1-\alpha)$.
\vspace{0.1cm}
\hrule

~\\
\noindent\textbf{Graph-structured regression.}\label{page:regression} To illustrate other choices of error functionals beyond norms such as $\psi(\hat \sL,\sL)=\|\hat\sL-\sL\|_F$ or $\psi(\hat \sL,\sL)=\|\hat\sL-\sL\|_{\textup{op}}$, we now discuss an application to graph-structured regression~\citep{sadhanala2016graph,Calandriello:2018}. In this context, the user has a vector of observations $y=(y_1,\dots,y_n)$ associated with the $n$ vertices of $\sG$, and the unsparsified version of the task is to use $y$ and $\sL$ to estimate a vector of unknown parameters $\beta^{\circ}\in\R^n$. Ordinarily, the estimate $r(\sL)$ for $\beta^{\circ}$ is computed as a solution to an optimization problem of the form $r(\sL)=\textup{argmin}_{\beta\in \R^n}\{\ell(y,\beta)+\tau\beta\ttop \sL\beta\}$. Here, $\ell(y,\beta)$ measures the goodness of fit between $y$ and a candidate vector $\beta$, and $\tau \beta\ttop \sL\beta$ penalizes vectors $\beta$ that do not respect the structure of $\sG$, with $\tau\geq0$ being a tuning parameter.
In situations where $\sL$ is very large or dense, the previously cited works have proposed approximating $r(\sL)$ with $r(\hat\sL)=\textup{argmin}_{\beta\in \R^n}\{\ell(y,\beta)+\tau\beta\ttop \hat\sL\beta\}$. However, the accuracy of $r(\hat\sL)$ as an approximation to $r(\sL)$ is unknown. To address this issue, Algorithm~1 can be applied with an error functional such as $\psi(\hat\sL,\sL)=\|r(\hat\sL)-r(\sL)\|_2$,
and we illustrate this empirically in Section~\ref{sec:expt}.

\subsection{Simultaneous confidence intervals }\label{sec:method:CI}
The second aspect of our proposed methodology deals with CIs for various quantities
associated with $\sL$. We discuss this first in the context of graph cut values, and then explain how the same approach can be extended to the eigenvalues of $\sL$.

 \noindent\textbf{Background on graph cuts.} By definition, a \emph{cut} in a graph $\sG=(\sV,\sE,w)$ is a partition of the vertex set $\sV$ into two disjoint subsets, and the \emph{value} of the cut is the sum of the weights of the edges that connect vertices in the two subsets.
 Recalling the notation $\sV=\{1,\dots,n\}$, every cut can be identified with a binary vector $x\in\{0,1\}^n$, where the two subsets of vertices are $\{i\in \sV|x_i=0\}$ and $\{i\in \sV|x_i=1\}$. This representation of a cut allows its value, denoted $\sC(x)$, to be computed as 
 $\sC(x)=x\ttop \sL x=\sum_{\{i,j\}\in E} w(i,j) (x_i-x_j)^2$. 
 
Because many fundamental characteristics of graphs can be computed in terms of cut values, it is common for algorithms to be formulated in terms of a collection of ``cut query'' vectors $\sCc\subset\{0,1\}^n$, which is specific to the user's task. Moreover, there is a well-established line of research on using edge sampling to efficiently approximate the cut values of large or dense graphs~\citep[][]{Benczur:1996,Benczur:2015,Woodruff:2016:cuts,Arora:2019:cut}. Hence, this amounts to approximating $\{\sC(x)|x\in\sCc\}$ using
the cut values of a sparsified graph $\{\hat \sC(x) |x\in\sCc\}$, where we define $\hat \sC(x)=x\ttop\hat\sL x$. To quantify the approximation error, we propose an algorithm that uses $\{\hat\sC(x)|x\in\sCc\}$ to build simultaneous CIs for $\{\sC(x)|x\in \sCc \}$.

\noindent\textbf{Simultaneous CIs for graph cut values.} 
The starting point for our approach is to consider the $(1-\alpha)$-quantile $q_{1-\alpha}$ of the unobserved random variable 
$$\xi=\max_{x\in\mathcal{C}}\frac{|\hat\sC(x)-\sC(x)|}{\hat\sigma(x)},$$
where $\hat\sigma^2(x)$ is an estimate of $\var(\hat\sC(x))$ to be detailed shortly. It is straightforward to check that if $q_{1-\alpha}$ were known, then the (theoretical) CIs defined by \smash{$\mathcal{I}_{1-\alpha}(x)=[\hat\sC(x)\pm \hat\sigma(x)q_{1-\alpha}]$} would have a simultaneous coverage probability $\P(\bigcap_{x\in\mathcal{C}}\{\sC(x)\in\mathcal{I}_{1-\alpha}(x)\})$ that is at least $1-\alpha$.
The crux of the problem is to construct a quantile estimate $\hat q_{1-\alpha}$, which will allow us to use practical intervals defined by $\hat{\mathcal{I}}_{1-\alpha}(x)=[\hat\sC(x)\pm \hat\sigma(x)\hat q_{1-\alpha}]$. Despite the seeming simplicity of this definition, the theoretical problem of demonstrating that these intervals have a simultaneous coverage probability close to $1-\alpha$ is quite involved when the number of queries $|\sCc|$ is large.
Nevertheless, we will show in Theorem~\ref{thm:cutquery} that the intervals can succeed even when $|\sCc|$ is allowed to diverge asymptotically.

Analogously to Algorithm 1, the main idea for constructing $\hat q_{1-\alpha}$ here is to generate approximate samples $\xi_1^*,\dots,\xi_B^*$ of $\xi$, and then define $\hat q_{1-\alpha}=\textup{quantile}(\xi_1^*,\dots,\xi_B^*;1-\alpha)$. 
For this purpose, it is natural to define the quantities $\hat \sC_i(x)=x\ttop \sQ_i x$ so that we have $\hat \sC(x) =\frac{1}{N}\sum_{i=1}^N \hat \sC_i(x)$, and we may estimate $\textup{var}(\hat\sC(x))$ using \smash{$\hat\sigma^2(x)=\frac{1}{N}\sum_{i=1}^N (\hat\sC_i(x)-\hat\sC(x))^2$}.
In this notation, Algorithm~2 generates approximate samples having the form
$$\xi^*=\max_{x\in\mathcal{C}}\frac{|\hat\sC^{*}(x)-\hat\sC(x)|}{\hat\sigma(x)},$$
where $\hat\sC^{*}(x)=\frac{1}{N}\sum_{i=1}^N \hat \sC_i^{*}(x)$ and $\hat \sC_1^{*}(x),\dots,\hat\sC_N^{*}(x)$ are drawn with replacement from $\hat \sC_1(x),\dots,\hat\sC_N(x)$. 
(The exceptional case that $\hat\sigma(x)=0$ for some $x\in\sCc$ is handled by treating $|\hat\sC^{*}(x)-\hat\sC(x)|/\hat\sigma(x)$ as 0, because in this case we must have  $\hat\sC^{*}(x)=\hat\sC(x)$.)

\smash{\textbf{Algorithm 2}~(Simultaneous CIs for graph cut values)}\\
\vspace{-0.2cm}
\hrule
\textbf{Input:} Number of bootstrap samples $\sB\geq 1$, a number $\alpha\in(0,1)$, and the set $\{\hat \sC_i(x)\,|\, x\in\sCc, 1\leq i\leq N\}$.

Compute the estimates $\hat \sC(x)=\frac{1}{N}\sum_{i=1}^N \hat \sC_i(x)$ and $\hat\sigma^2(x)=\frac{1}{N}\sum_{i=1}^N (\hat \sC_i(x)-\hat \sC(x))^2$ for each $x\in\sCc$.

\textbf{for $b=1,\dots,B$ in parallel do:}
\vspace{-0.2cm}
\begin{itemize}[leftmargin=*]
    \item Generate  $(\sW_1^*,\dots,\sW_N^*) \sim$ Mult.$(\sN; \ts \frac{1}{N},\dots,\frac{1}{N})$.
    \item Compute $    \xi_b^*= \max_{x\in\mathcal{C}} \frac{1}{\hat\sigma(x)}|\frac{1}{N}\sum_{i=1}^N (\sW_i^*-1)\hat \sC_i(x)|.$
\end{itemize}
\vspace{-0.2cm}
\textbf{end for}

Compute  $\hat q_{1-\alpha}=\textup{quantile}( \xi_1^*,\dots \xi_B^*; 1-\alpha)$.

\textbf{Output:} The collection of CIs $\{\hat{\mathcal{I}}_{1-\alpha}(x)\,|\,x\in\sCc\}$ defined by $\hat{\mathcal{I}}_{1-\alpha}(x)=[\hat \sC(x)\pm \hat\sigma(x)\hat q_{1-\alpha}]$.
\vspace{0.1cm}
\hrule

\vspace{0.1cm}

\noindent\textbf{Remarks.} Notably, this algorithm does not require a second level of bootstrap sampling, which is an important contrast with Algorithm~1. 
The main reason for this simplification is that we can estimate $\E(\hat\sC(x))$ and $\var(\hat\sC(x))$ using explicit functions of $\hat\sC_1(x),\dots,\hat\sC_N(x)$, whereas it was not possible to estimate $\E(\psi(\hat\sL,\sL))$ and $\var(\psi(\hat\sL,\sL))$ in the same manner for a general choice of $\psi$. One more significant point is that $\hat q_{1-\alpha}$ in Algorithm~2 can be used to extract information about the maximal cut query value $\sC_{\max}=\max_{x\in \mathcal{C}}\sC(x)$ and minimal cut query value $\sC_{\min}=\min_{x\in \mathcal{C}}\sC(x)$, which are of interest in many graph partitioning problems.
Specifically, $\sC_{\max}$ is covered by  $[\max_{x\in\mathcal{C}}\{\hat\sC(x)-\hat\sigma(x)\hat q_{1-\alpha}\}, \max_{x\in\mathcal{C}}\{\hat\sC(x)+\hat\sigma(x)\hat q_{1-\alpha}\}]$ with a probability at least as large as the simultaneous coverage probability of $\{\sC(x)\in \hat{\mathcal{I}}_{1-\alpha}(x)\,|\,x\in\sCc\}$. The same holds, mutatis mutandis, for $\sC_{\min}$.

\noindent\textbf{Simultaneous CIs in spectral clustering.} 
One of the most well known machine learning tasks involving graph Laplacians is spectral clustering~\citep{von2007tutorial}, which uses Laplacian eigenvectors to construct low-dimensional representations of data that allow clusters to be distinguished more effectively. Because the Laplacians in spectral clustering tend to be dense, sparsification has been advocated as a way to improve computational efficiency~\citep{chakeri2016spectral,Woodruff:2016,sun:2019:cluster}. On the other hand, sparsification can also distort the clustering results.

As an illustration of how Algorithm~2 can be adapted to address this issue, we focus on one of the most pivotal steps in clustering: the selection of the number of clusters. Often, this choice is made by searching for a prominent gap among the bottom eigenvalues of a Laplacian, and then choosing the number of clusters to be the number of eigenvalues that fall below that gap~\citep{von2007tutorial}. However, when a sparsified Laplacian is used, this selection technique becomes more nuanced, 
because if the gaps between eigenvalues are too sensitive to the chance variation from sparsification, then they may be unreliable indicators for the correct number of clusters.

To quantify the uncertainty, it is possible to construct simultaneous CIs, say $\hat{\mathcal{I}}_1,\dots,\hat{\mathcal{I}}_r$, for the eigenvalues $\lambda_1(\sL)\leq \cdots\leq\lambda_r(\sL)$, where $r\geq 2$ is a number that the user believes is safely above the correct number of clusters.  If there is an index $j\in\{1,\dots,r\}$ such that a clear gap exists between the upper endpoint of $\hat{\mathcal{I}}_{j}$ and the lower endpoint of $\hat{\mathcal{I}}_{j+1}$, then this gives more credible evidence that $j$ clusters are present, because in this case, the gap cannot be easily explained away by the sparsification error. 

The intervals $\hat{\mathcal{I}}_1,\dots,\hat{\mathcal{I}}_r$ are constructed as follows. First, we put $\hat{\mathcal{I}}_1=\{0\}$, since $\lambda_1(\sL)$ is always 0. Next, the definition of  $\xi_b^*$ in Algorithm~2 can simply be replaced by
$\xi_b^*= \max_{2 \leq j \leq r} |\lambda_j(\hat{\sL}^*)/\lambda_j(\hat\sL)-1|$, where the $j$th quantity in the max is set to 0 when $\lambda_j(\hat\sL)=0$, because this implies $\lambda_j(\hat\sL^{*})=0$.  In turn, $\hat q_{1-\alpha}$ in Algorithm~2 can be used to define the CIs $\hat{\mathcal{I}}_j = [\lambda_j( \hat \sL)/(1+\hat q_{1-\alpha}),\lambda_j( \hat \sL)/(1-\hat q_{1-\alpha})]$ for $j\in\{ 2,\dots,r\}$, with the upper endpoint interpreted as $\infty$ in the unlikely case $\hat q_{1-\alpha}\geq 1$. Lastly, in  Appendix~\ref{sec:clustering}, we present several empirical examples showing that these CIs provide effective guidance in selecting the number of clusters.

\subsection{Computational efficiency}\label{sec:computation}

We now address the computational efficiency of the proposed algorithms. Given that Algorithm~1 uses a double bootstrap, it is important to begin by providing some historical context.
Because double bootstrapping was first developed in the 1980s~\citep{Efron:1983,Beran:1988}, the computing
environments of that time were ill-suited to its structure, and it acquired a long-held reputation of being computationally intensive. However, due to major technological shifts, this perception is becoming increasingly outdated. In particular, there are three aspects of our algorithms that make them affordable in modern computing environments: (1) low communication cost, (2) high parallelism, and (3) incremental refinement.

\noindent\textbf{Low communication cost.} As was discussed in the introduction, graph sparsification is often intended for settings where $\sG$ is too large or dense to be stored in fast memory. In these situations, the communication cost of accessing $\sG$ in order to generate $\hat \sL$ is often of greater concern than the flop count of subsequent computations on $\hat\sL$~\citep[][\textsection 16.2]{Tropp:2020}.  (This is sometimes also referred to as an instance of the ``memory wall'' problem~\citep{memory:wall}.) 
Meanwhile, it is crucial to recognize that Algorithms~1 and~2 \emph{do not require any additional access to $\sG$}, since they only rely on the samples used to produce  $\hat\sL$. 
Hence, when the communication cost to access $\sG$ is high, it is less likely that error estimation will be a bottleneck.

\noindent\textbf{High parallelism.} Another factor that counts in favor of Algorithms~1 and~2 is that bootstrap sampling is ``embarrassingly parallel'', which is to say that all of the samples within a given loop can be computed independently. Moreover, this is especially favorable as cloud and GPU computing are becoming ubiquitous. In fact, the Python Package Index now includes a GPU-compatible package that is specifically designed to perform bootstrap sampling~\citep{Recombinator}.

With regard to Algorithm~1, some additional attention should be given to the fact that its two loops are nested---which might appear to restrict the benefit of parallelism.
However, the nested structure is manageable for two reasons. First, in many settings, it is sufficient to take only  $\sB\sim 50$ bootstrap samples in each loop, and this is demonstrated empirically in Section~\ref{sec:expt}. Second, there are established techniques in GPU computing for parallelizing nested loops.

\noindent\textbf{Processing cost and incremental refinement.} 
Whereas the communication cost of Algorithms~1 and~2 is likely to be much less than that of the overall graph sparsification workflow, a comparison of processing cost (e.g.~flop count) involves more considerations. Due to the high parallelism of Algorithms~1 and~2, the main driver of their runtimes will be the processing cost of one iteration of each loop. Often, this cost will be similar to that of the main task involving $\hat\sL$. For example, in graph-structured regression, where the main task is to compute $r(\hat\sL)$, the cost of computing $\psi(\hat\sL^{*},\hat\sL)=\|r(\hat\sL^{*})-r(\hat\sL)\|_2$ and $\psi(\hat\sL^{**},\hat\sL^{*})=\|r(\hat\sL^{**})-r(\hat\sL^{*})\|_2$ in Algorithm~1 will be dominated by the cost of computing $r(\hat\sL^{**})$ and $r(\hat\sL^{*})$, which is proportional to the cost of computing $r(\hat\sL)$. Similarly, for cut queries, if the user's main task with $\hat\sL$ is to compute the maximal approximate cut value $\max_{x\in\mathcal{C}}\hat \sC(x)$, then this will be similar to the cost of computing $\xi_b^*$ in Algorithm~2.

Based on the reasoning above, the runtimes of Algorithms~1 and~2 are expected to be similar to the runtime of the main task involving $\hat\sL$, which in turn, is expected to be less than the communication time of accessing $\sG$. So, from this standpoint, error estimation is not expected to substantially increase the overall cost of the workflow. But as it turns out, there is one further technique that can be used to make the cost of error estimation even lower---which is \emph{incremental refinement}. The first step of this technique is to generate a ``rough'' preliminary instance of $\hat\sL$ based on a small sample size, say $\sN_0$. If we let $q_{1-\alpha}(\sN)$ denote the $(1-\alpha)$-quantile of $\psi(\hat\sL,\sL)$ based on a generic sample size $\sN$, then the key idea is that an estimate $\hat q_{1-\alpha}(\sN_0)$ can be obtained inexpensively, and then it can be used to ``forecast'' what larger sample size $\sN_1\gg \sN_0$ is needed to \emph{refine} the sparsified Laplacian so that $q_{1-\alpha}(\sN_1)$ is below a target threshold.  In other words, the error estimation is accelerated because it is faster to run Algorithms~1 and~2 when there are $\sN_0$ sampled edges, rather than $\sN_1$. This process of ``forecasting'' $\sN_1$ is based on an easily implemented type of extrapolation that is well established in the bootstrap literature~\citep{Bickel:Extrapolation}, and is detailed in Appendix~\ref{app:extrap}. In particular, we show empirically that the rule is effective when \emph{$\sN_0$ is 10 times smaller than $\sN_1$}, enabling substantial speedups.

\section{THEORETICAL RESULTS}\label{sec:theory}
In this section, we present two results that establish the theoretical validity of Algorithms~1 and~2 in the limit of large graphs with a diverging number of vertices, $n\to\infty$.
The first result shows that Algorithm~2 produces CIs that simultaneously cover the exact cut values $\{\sC(x)|x\in \sCc\}$ with a probability that is asymptotically correct. Likewise, the second result shows that when $\psi(\hat\sL,\sL)=\|\hat\sL-\sL\|_F^2$, Algorithm~1 produces a quantile estimate that upper bounds $\|\hat \sL-\sL\|_F^2$ with a probability converging to the correct value. The proofs of both results require extensive theoretical analysis based on \emph{high-dimensional central limit theorems}, and are deferred to Appendices~\ref{app:cut} and~\ref{app:Frobenius}.

\noindent\textbf{Setting for theoretical results.}  Our theoretical results are framed in terms of a sequence of weighted undirected graphs $\sG_n=(\sV_n,\sE_n,w_n)$ indexed by the number of vertices $n=1,2,\dots$, such that $\sV_n$, $\sE_n$, and $w_n$ are allowed to vary as functions of $n$. For each $n$, we assume that the sparsified Laplacian $\hat \sL_n$ is obtained by drawing $\sN_n$ edges from $\sG_n$ in an i.i.d~manner via edge-weight sampling. Lastly, the number of bootstrap samples $\sB_n$ and the set of cut queries $\sCc_n$ may also vary with $n$.

\noindent{\textbf{Simultaneous CIs for cut values.}} Some notation is needed for our first result. Let $w_n(\sE_n)=\sum_{e\in E_n}w_n(e)$ denote the total weight of $\sG_n$, and for any binary cut vector $x$, let $\underline{\sC}(x)=x\ttop \sL_n x/w_n(\sE_n)$ be its standardized value, which satisfies $0\leq \underline{\sC}(x)\leq 1$.
Lastly, for a set $\sCc_n\subset\{0,1\}^n$, define the theoretical quantity $\eta(\sCc_n) = \min_{x \in \mathcal{C}_n}\{\underline{\sC}(x) (1-\underline{\sC}(x))\}$.

\begin{theorem}\label{thm:cutquery}
    As $n\to\infty$, suppose that $\sN_n\to\infty$ and \smash{$\sB_n\to\infty$,} as well as $ \log(\sN_n |\sCc_n|)^5=o(\sqrt{\sN_n}   \eta(\sCc_n))$.
  Then, for any fixed $\alpha\in (0,1)$, the confidence intervals $\{\hat{\mathcal{I}}_{1-\alpha}(x)|x\in \sCc_n\}$ produced by Algorithm~2 have a simultaneous coverage probability that   satisfies the following limit as $n\to\infty$,
    \begin{equation}
         \small
        \P\bigg(\bigcap_{x\in  \mathcal{C}_n} \Big\{\sC(x)\in \hat{\mathcal{I}}_{1-\alpha}(x) \Big\}\bigg) \ \to \  1-\alpha.
    \end{equation}
\end{theorem}
\noindent\textbf{Remarks.} A valuable feature of this result is that it can handle situations where $\sCc_n$ is a large set, since the cardinality $|\sCc_n|$ is only constrained through a polylogarithmic function, $\log(\sN_n|\sCc_n|)^5=o(\sqrt{\sN_n}\eta(\sCc_n))$. This means that Algorithm~2 can succeed in high-dimensional inference problems, because $|\sCc_n|$ (i.e.~the number of unknown parameters) may diverge.

With regard to the role of $\eta(\sCc_n)$, a notable point is that its value is allowed to approach 0 as $n\to\infty$, as long as $\eta(\sCc_n)$ is of larger order than $\log(\sN_n|\sCc_n|)^5/\sqrt{\sN_n}$. In essence, values of $\eta(\sCc_n)$ near 0 occur when $\sCc_n$ contains a cut $x$ whose value is negligible compared to $w_n(\sE_n)$, or when the two graph components induced by $x$ have negligible weight compared to $w_n(\sE_n)$. The reason that such cuts need to be excluded is technical, because if $\underline{\sC}(x)$ is close to 0 or 1, then the random variable $x\ttop \hat \sL_n x$ is nearly degenerate---which interferes with establishing limiting distributions for statistics that depend on $x\ttop \hat \sL_n x$. To briefly mention some explicit examples that are covered by Theorem~\ref{thm:cutquery}, it is known that for Erd\H{o}s-Renyi graphs with average degree $\gamma$, our assumption involving $\eta(\sCc_n)$ holds with high probability as $n\to\infty$, provided that all $x\in\sCc_n$ are bisections (i.e.~$\|x\|_1=n/2$), $ \gamma$ is sufficiently large, and $\log(|\sCc_n|)^5=o(\sqrt{N_n})$~\citep[][]{Dembo}.
We will also show empirically that Algorithm~2 can work well for natural graphs when all cuts in $\sCc_n$ are drawn uniformly at random and $|\sCc_n|= n$.

\noindent\textbf{Error estimates for the Frobenius norm.} Our next result provides a guarantee on the performance of Algorithm~1 when $\psi(\hat \sL_n, \sL_n)=\|\hat \sL_n-\sL_n\|_F^2$. Here, the choice to use $\|\cdot\|_F^2$ rather than $\|\cdot\|_F$ is essentially a matter of mathematical convenience, because if $\hat q_{1-\alpha}$ is an estimated quantile for $\|\hat \sL_n-\sL_n\|_F^2$, then $\sqrt{\hat q_{1-\alpha}}$ has equivalent performance for $\|\hat \sL_n-\sL_n\|_F$.
\begin{theorem} \label{thm:Frobenius}
As $n \to \infty$, suppose that \smash{$\sN_n \to \infty$,} \smash{$\sB_n\to\infty$}, $n/\sN_n\to 0$, and $\|\mathsf{d}_n\|_{\infty}/\|\mathsf{d}_n\|_2 \to 0$ hold, where $\mathsf{d}_n\in\R^n$ contains the diagonal entries of $L_n$. 
Then, for any fixed $\alpha\in(0,1)$, the quantile estimate $\hat q_{1-\alpha}$ produced by Algorithm~1 satisfies the following limit as $n\to\infty$,
\begin{equation}
    \P\big( \|\hat L_n-L_n\|_F^2 \leq \hat q_{1-\alpha} \big) \ \to \ 1- \alpha.
\end{equation}
\end{theorem}
\noindent\textbf{Remarks.} The condition that the sample size $\sN_n$ be of larger order than the number of vertices $n$ is typical in the analysis of graph sparsification algorithms. As for the vector of degrees $\mathsf{d}_n$, the condition $\|\mathsf{d}_n\|_{\infty}/\|\mathsf{d}_n\|_2 \to 0$ has the interpretation that no single vertex dominates the entire graph with respect to degrees.

\begin{table*}[h!]
\caption{Results for Algorithms~1 and~2 in several error estimation tasks. Under the heading of `graph cuts', we report the observed value of the simultaneous coverage probability $\P(\bigcap_{x\in\mathcal{C}}\{\sC(x)\in\hat{\mathcal{I}}_{1-\alpha}(x)\})$, with $1-\alpha$ being 90\% or 95\%. In the columns to the right of `graph cuts', we report the observed value of $\P(\psi(\hat\sL,\sL)\leq \hat q_{1-\alpha})$ for the three choices of $\psi$ in a similar manner.\\[-0.2cm]}
\centering
\small
\setlength\tabcolsep{3pt}
\begin{tabular}{lcccccccccccccc}
                   &             &             & &\multicolumn{2}{c}{graph cuts} &  &\multicolumn{2}{c}{$\|\hat \sL-\sL\|_F$} &  & \multicolumn{2}{c}{$\|\hat\sL-\sL\|_{\textup{op}}$} &  & \multicolumn{2}{c}{\!$\|r(\hat \sL) - r( \sL) \|_{2}$}  \\ \cline{5-6} \cline{8-9} \cline{11-12} \cline{14-15} \specialrule{0em}{1pt}{1pt}
\ \ \  $\sG$  & $|\sE|$ & sampling  & & 90\%  & 95\% && 90\%  & 95\% && 90\%  & 95\% && 90\%  & 95\%  \\ \hline
\specialrule{0em}{1pt}{1pt}
\multirow{3}{*}{{\tt{Citations}}}                   &\multirow{3}{*}{ 218,835 }& EW &   &          88.3 &93.2 &&89.6 &94.1 & &89.4 &93.3 & &90.1 &94.3\\  && ER& &87.8 &92.8&   &92.0 &95.7 & &91.6 &95.7 & &93.3 &96.8\\  && AER& &88.1 &93.4&   &90.1 &94.6 & &89.5 &94.7 & &92.4 &96.0\\ \specialrule{0em}{1pt}{1pt} \hline
\specialrule{0em}{1pt}{1pt}
\multirow{3}{*}{{\tt{DIMACS}}}                   &\multirow{3}{*}{ 1,799,532 }& EW & &          90.1 &94.9 &&88.6 &94.9 & &89.4 &93.9 & &89.3 &94.8\\  && ER & & 90.4 &94.6& &89.0 &93.3 & &87.9 &93.7 & &90.3 &94.4\\  && AER & &89.3 &93.5& &88.8 &93.5 & &88.5 &93.0 & &90.0 &94.7\\ \specialrule{0em}{1pt}{1pt} \hline
\specialrule{0em}{1pt}{1pt}
\multirow{3}{*}{{\tt{Genes}}}                   & \multirow{3}{*}{ 743,712 }&  EW      &   &         87.7 &93.4& &     87.1 &93.2 & &88.2 &93.5 & &88.2 &94.7\\  &  &  ER &   &87.7 &93.7 & &90.4 &94.3 & &89.7 &94.4 & &87.0 &92.6\\  &  &  AER & &87.5 &93.4 &   &88.8 &94.5 & &88.9 &93.3 & &89.8 &95.8\\ \specialrule{0em}{1pt}{1pt}\hline
\specialrule{0em}{1pt}{1pt}
\multirow{3}{*}{{\tt{Howard}}}                   &\multirow{3}{*}{ 107,264 }& EW &          &88.7 &94.6&   &89.8 &94.2 & &90.0 &94.7 & &88.4 &94.5\\  && ER& &87.3 &92.2&   &90.9 &94.5 & &93.1 &96.5 & &87.7 &93.9\\  && AER& &89.7 &94.5&   &90.7 &94.5 & &91.5 &94.8 & &89.3 &93.6\\ \specialrule{0em}{1pt}{1pt} \hline
\specialrule{0em}{1pt}{1pt}
\multirow{3}{*}{{\tt{M14}}}                   & \multirow{3}{*}{ 172,195 }& EW &        &90.3 &94.7&   &91.2 &96.1 & &90.8 &94.6 & &87.3 &93.7\\  && ER& &89.2 &94.0&   &90.9 &95.2 & &92.5 &95.9 & &88.4 &93.5\\  && AER& &89.6 &94.0&   &89.0 &94.2 & &92.0 &95.3 & &89.4 &94.2\\ \specialrule{0em}{1pt}{1pt} \hline
\end{tabular}
\label{table:cutfunctionLap}
\end{table*}
\normalsize

\section{EMPIRICAL RESULTS}\label{sec:expt}

This section investigates the empirical performance of our proposed error estimation methods in three applications: graph cut queries, Laplacian matrix approximation, and graph-based regression. A fourth application to spectral clustering is covered in Appendix~\ref{sec:clustering}.

\noindent\textbf{Graphs.}
The experiments were based on five graphs:  {\tt{Citations}}, {\tt{DIMACS}}, {\tt{Genes}}, {\tt{Howard}}, and {\tt{M14}}, which are detailed in Appendix~\ref{sec:Additionaldetails}, along with information about the computing resources used in our experiments. {\tt{Citations}} represents the co-citation graph of scientific papers from a section of arXiv~\citep{rossi2015network}.
{\tt{DIMACS}} is a benchmarking graph from the DIMACS Implementation Challenge~\citep{bader201110th}.
{\tt{Genes}} is a human gene regulatory network~\citep{davis2011university}.  
 {\tt{Howard}} is a student social network~\citep{traud2012social,rossi2015network}. {\tt{M14}} is the Mycielskian14 graph, which is part of a test suite for benchmarking graph algorithms~\citep{davis2011university}.

From each of the graphs mentioned above, we constructed a corresponding graph $\sG$ by randomly sampling $n=2,000$ vertices according to their degrees (without replacement) and retaining all edges among the sampled vertices. The resulting number of edges $|\sE|$ for each graph $\sG$ is reported in Table \ref{table:cutfunctionLap}, where we use the same name to refer to $\sG$ and the original graph it was drawn from.

\noindent\textbf{Experiment settings.}
We considered three sampling schemes for sparsifying each $\sG$: edge-weight sampling (EW), effective-resistance sampling (ER), and approximate effective-resistance sampling (AER)~\citep{srivastava2011,EstimateER}. Whereas EW and ER were defined in the introduction, the definition of AER is more involved and is discussed in Appendix~\ref{sec:Additionaldetails}. For the five choices of $\sG$ and three choices of edge sampling scheme, we generated 1,000 sparsified Laplacians $\hat \sL$, yielding 15,000 in total. The number of sampled edges $\sN$ for constructing $\hat\sL$ was chosen to be 10\% of the total number of edges, $\sN = |\sE|/10$.

For every realization of $\hat\sL$, we applied Algorithms 1 and 2 in four error estimation tasks: simultaneous CIs for graph cut values, as well as quantile estimation for $\psi(\hat\sL,\sL) = \|\hat\sL-\sL\|_F$, $\psi(\hat\sL,\sL) = \|\hat\sL-\sL\|_{\textup{op}}$, and $\psi(\hat\sL,\sL)=\|r(\hat\sL)-r(\sL)\|_2$.
With regard to the number of bootstrap samples in Algorithm~1, we used $\sB=50$ for the outer loop and $\sB=30$ for the inner loop. For Algorithm~2, we used $\sB=50$. 
Under the heading of `graph cuts' in Table~\ref{table:cutfunctionLap}, we report the observed value of the simultaneous coverage probability $\P(\bigcap_{x\in\mathcal{C}}\{\sC(x)\in\hat{\mathcal{I}}_{1-\alpha}(x)\})$, and in a similar manner, we report the observed value of $\P(\psi(\hat\sL,\sL)\leq \hat q_{1-\alpha})$ for the three choices of $\psi$, where the desired confidence level $1-\alpha$ is either 90\% or 95\%. The observed probabilities were computed by averaging over the 1,000 trials in each setting. 

There are a few more details to mention about cut queries and graph-structured regression. The set of cuts $\sCc\subset\{0,1\}^n$ was selected by independently generating 2,000 random vectors whose entries were i.i.d. Bernoulli(1/2) random variables.
Next, for the graph-structured regression task on page \pageref{page:regression}, we adopted the following setting considered in \cite{sadhanala2016graph}: The vector of observations $y\in\R^n$ was generated from the Gaussian distribution $N(\beta^{\circ},  \varsigma^2 I)$, where the mean $\beta^{\circ}\in\R^n$ was obtained by averaging 20 (unit norm) eigenvectors of $\sL$ corresponding to the smallest 20 eigenvalues, and the scalar variance parameter was $\varsigma^2=\frac{1}{n}\sum_{i=1}^n (\beta_i^{\circ}-\bar\beta^{\circ})^2$, with $\bar\beta^{\circ}=\frac{1}{n}\sum_{i=1}^n \beta_i^{\circ}$.  Also, the loss function was taken as $\ell(y,\beta)=\|y-\beta\|_2^2$, and the tuning parameter was set to $\tau=0.01$.

\noindent\textbf{Discussion of empirical results.}  Table~\ref{table:cutfunctionLap} captures the performance of our proposed algorithms in 120 distinct settings---corresponding to five choices of $\sG$, three choices of edge sampling, four choices of task, and two choices of confidence level. Thus, both the quality and consistency of the empirical results are excellent, as the observed probabilities match the desired confidence level $1-\alpha$ to within about 2\% in all but a few settings. We also show in  Appendix~\ref{sec:clustering} that simultaneous CIs for the eigenvalues of $\sL$ exhibit similar performance in the context of spectral clustering.

\noindent\textbf{Computational efficiency.} The sizes of the five graphs used in the previous experiments were limited by a number of factors, such as the need to perform thousands of Monte Carlo trials, and compute ground truth errors involving unsparsified Laplacians. To assess the computational efficiency of error estimation, it is of interest to consider a runtime experiment involving a much larger graph. For this purpose, we used the {\tt{M20}} (Mycielskian20) graph \citep{davis2011university}, which is part of the same benchmarking suite as {\tt{M14}}, and contains $n=786,431$ nodes and $|\sE|=2,710,370,560$ edges. Storing this graph in a 3-column CSV file with $|\sE|$ rows requires more than 42 GB, where an edge connecting nodes $i$ and $j$ with weight $w(i,j)$ is saved as the row vector $(i,j, w(i,j))$. In particular, this graph is too large to be stored in the RAM of a typical laptop, and presents a situation where graph sparsification is a practical option for dealing with limited memory.

Taking the approach of incremental refinement described in Section~\ref{sec:computation}, we generated an initial sparsified Laplacian with $\sN_0\approx 0.02|\sE|$ sampled edges. (See Appendix~\ref{app:extrap} for additional experiments demonstrating the effectiveness of incremental refinement with such a choice of $\sN_0$.) Due to the large size of {\tt{M20}}, we used an approximate form of EW sampling that takes advantage of the fact that all the edge weights of {\tt{M20}} are equal. Under exact EW sampling, the counts for the sampled edges would be a random vector drawn from a Multinomial distribution, corresponding to tossing $\sN_0$ balls into $|\sE|$ bins, each with probability $1/|\sE|$. Since the entries of such a random vector are approximately independent Poisson(0.02) random variables, it is computationally simpler to divide the full graph into a series of small ``blocks'' that can fit into RAM, and independently sample the edge counts as independent Poisson(0.02) random variables within each block.

The blockwise edge sampling was performed on a laptop with 16 GB of RAM by referring to the full graph as a \texttt{tabularTextDatastore} object in MATLAB, which is a type of object that allows for the blocking to be automated. After this was done, the initial sparsified Laplacian $\hat\sL$, and the matrices $\sQ_1,\dots,\sQ_{N_0}$ were stored implicitly using sampled edge counts, so that memory need not be allocated for  $n\times n$ matrices. Next, we applied Algorithm 1 to estimate the 90\% quantile of $\psi(\hat\sL, \sL) = \|\hat\sL- \sL\|_F$, with $B=30$ iterations for the inner loop and $B=50$ iterations for the outer loop. Without using any parallelization for these loops, the overall runtime to obtain the quantile estimate was approximately 7 hours.

To place this runtime into context, we proceeded to the second stage of the incremental refinement approach, which involved generating a ``refined'' sparsified Laplacian based on $\sN_1\approx 0.1|\sE|$ sampled edges. The edge sampling in this stage was performed in the same manner as in the previous stage and took approximately 25 hours. We did not perform any additional tasks with this refined sparsified Laplacian, but if we did, it would have clearly increased the overall runtime of the workflow beyond 25 hours. This shows that the error estimation process increased the runtime of the workflow by at most $7/25= 28\%$. Moreover, this does not reflect the straightforward speedup that could be obtained by running either of the loops in Algorithm 1 in parallel. For instance, if the outer loop were distributed across 8 processors with the inner loop still being run sequentially, the error estimation process would only increase the runtime of the workflow by at most $(7/8)/25= 3.5\%$.

\noindent\textbf{Code.} The code for Algorithms~1 and~2 is available at the repository \url{https://github.com/sy-wwww/Error-Estimates-Graph-Sparsification}.

\section{CONCLUSION}
Due to the fact that graph sparsification has had far-reaching impact in machine learning and large-scale computing, our work has the potential to enhance many applications by providing users with practical error estimates. Indeed, considering that this is the first paper to develop a systematic way to estimate graph sparsification error, there is a substantial opportunity to adapt our approach to applications beyond the four that we have already presented here. Furthermore, the possibility of such extensions is underscored by our empirical results, which show that the error estimates perform reliably across a substantial range of conditions, corresponding to different graphs, edge sampling schemes, and error metrics. Lastly, we have also provided two theoretical performance guarantees that hold in a high-dimensional asymptotic setting where $n$, $|\sE|$, and $\sN$ diverge simultaneously.

\section*{Acknowledgements}
The authors gratefully acknowledge partial support from DOE grant DE-SC0023490.

\bibliography{citation}

\newpage
\onecolumn

\appendix
\section*{Supplementary material}

The appendices are organized as follows: Appendix \ref{sec:clustering} covers an application to spectral clustering. Appendix \ref{app:extrap} presents experiments illustrating the performance of incremental refinement. Appendix~\ref{sec:Additionaldetails} provides additional details about the design of the experiments and computing resources. Appendix \ref{sec:notation} presents the notation necessary for the proofs. Appendices \ref{app:cut} and \ref{app:Frobenius} contain the proofs of Theorems~\ref{thm:cutquery} and~\ref{thm:Frobenius} respectively. Appendix \ref{sec:background} provides background results used in the proofs.

\section{Empirical results on spectral clustering}
\label{sec:clustering}
In this section, we examine the performance of the simultaneous CIs for the eigenvalues of $L$ in spectral clustering. 

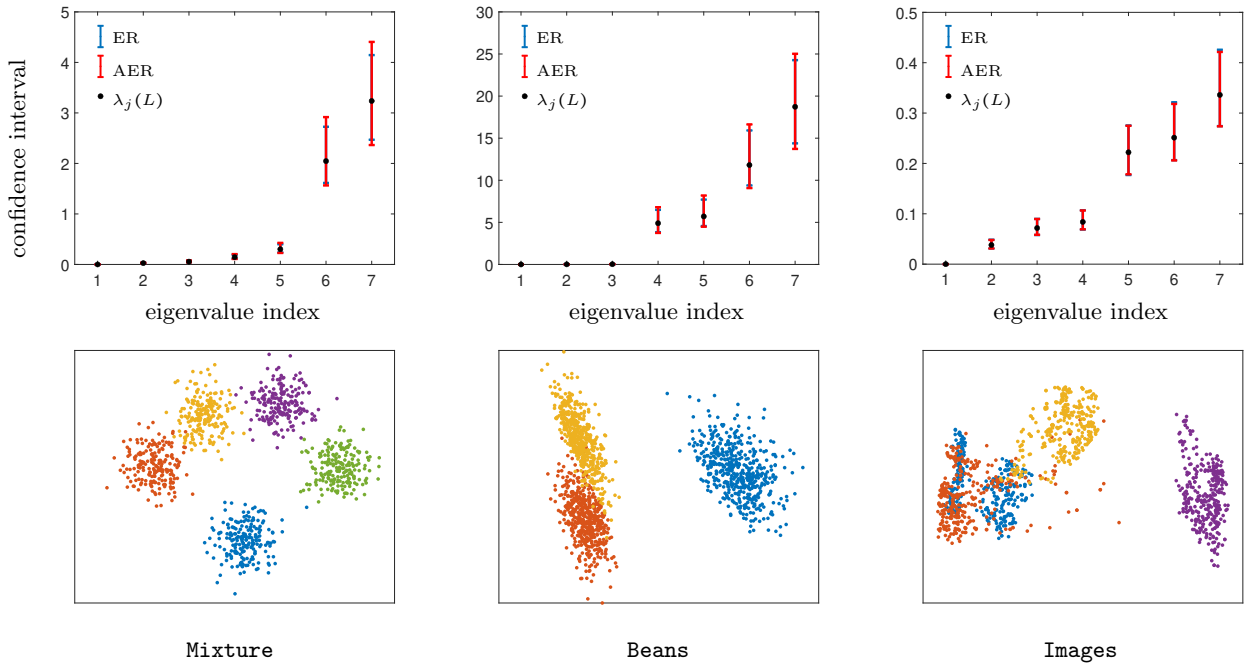
\begin{figure}[H]
\setlength{\abovecaptionskip}{10pt}
\centering 
\subfigure{
\begin{overpic}[width=0.32\textwidth]{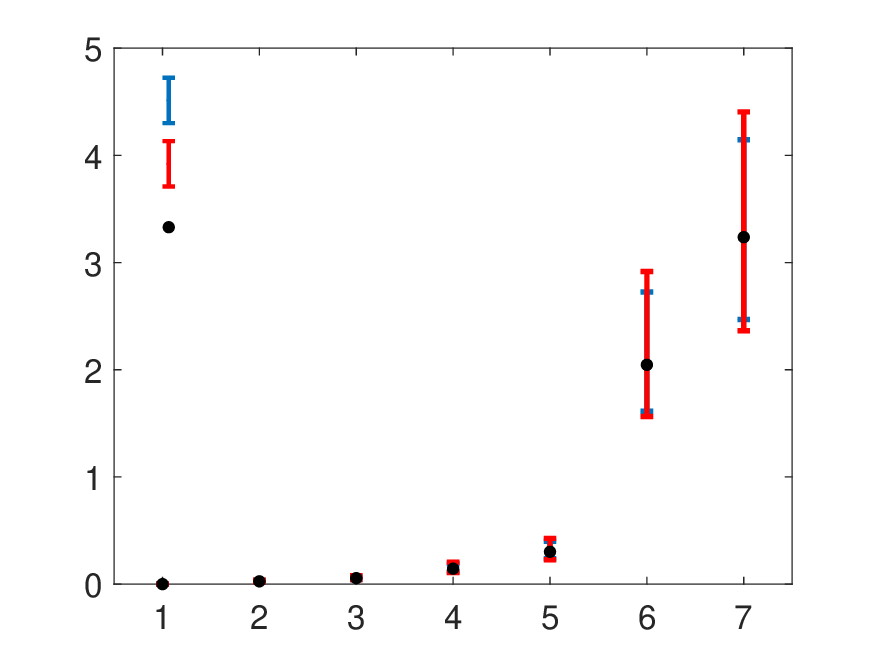}
\put(20,62){ \tiny \tiny ER}
\put(20,54){ \tiny \tiny AER}
\put(20,47){ \tiny \tiny $\lambda_j(L)$}
\put(-3,-3){\rotatebox{90}{ {  \ \ \ \ \small  confidence interval  \ \ }}}
\put(28,-5){ \footnotesize eigenvalue index }
\end{overpic}}
\hspace{-2mm}
\subfigure{    
\begin{overpic}[width=0.32\textwidth]{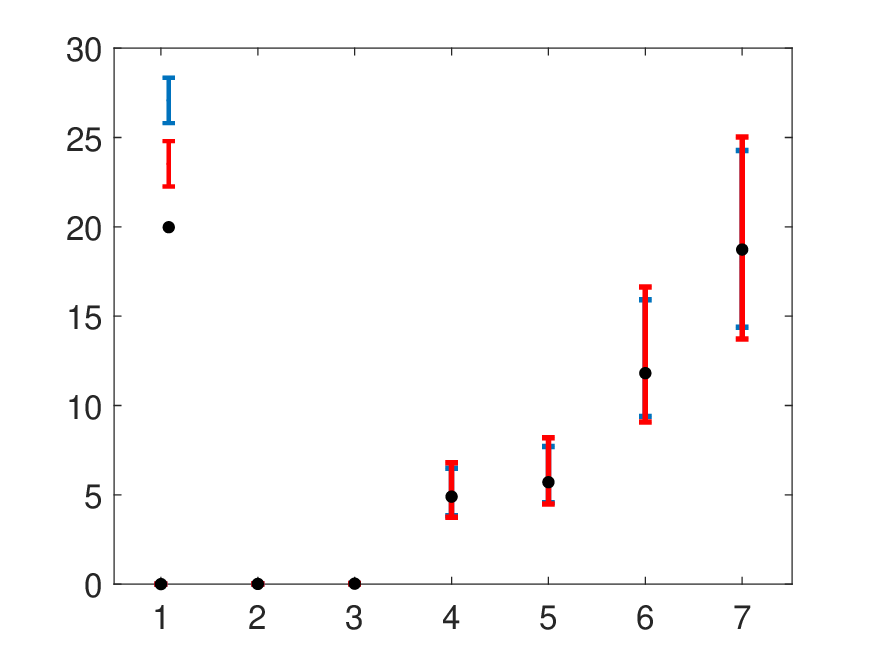}
\put(20,62){ \tiny \tiny ER}
\put(20,54){ \tiny \tiny AER}
\put(20,47){ \tiny \tiny $\lambda_j(L)$}
\put(28,-5){ \footnotesize eigenvalue index }
\end{overpic}}
\hspace{-2mm}
\subfigure{
\begin{overpic}[width=0.32\textwidth]{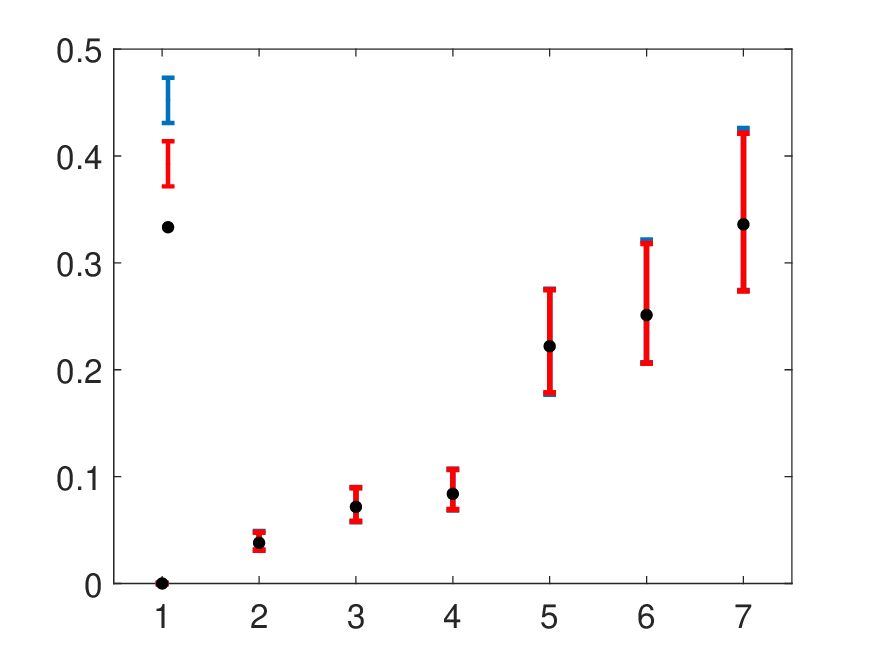}
\put(20,62){ \tiny \tiny ER}
\put(20,54){ \tiny \tiny AER}
\put(20,47){ \tiny \tiny $\lambda_j(L)$}
\put(28,-5){ \footnotesize eigenvalue index }
\end{overpic}}
\vspace{0.1cm}
\subfigure{
\begin{overpic}[width=0.32\textwidth]{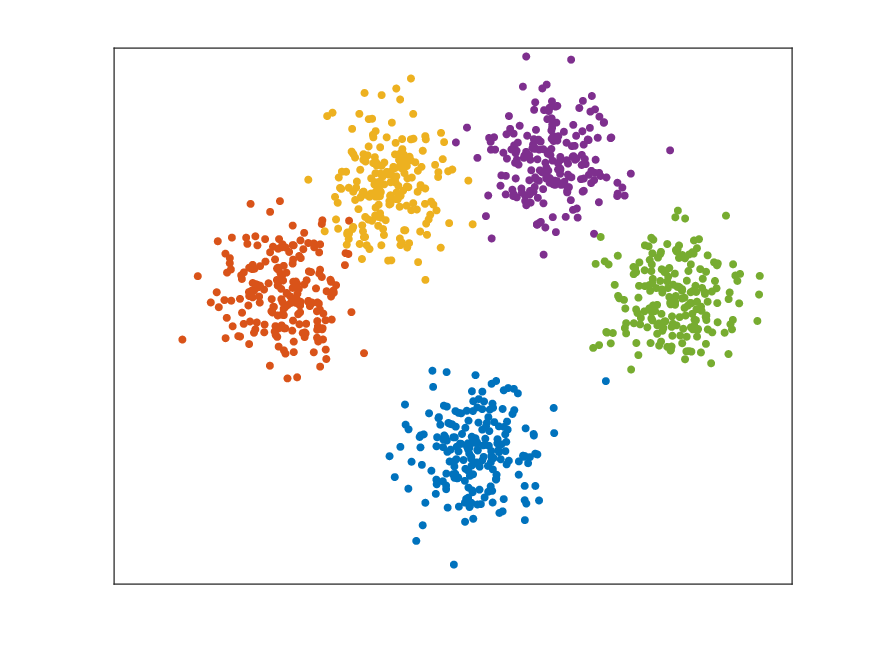}
\put(38,-5){ \small {\tt{Mixture}}}
	\end{overpic}}
\hspace{-2mm}
 \subfigure{
\begin{overpic}[width=0.32\textwidth]{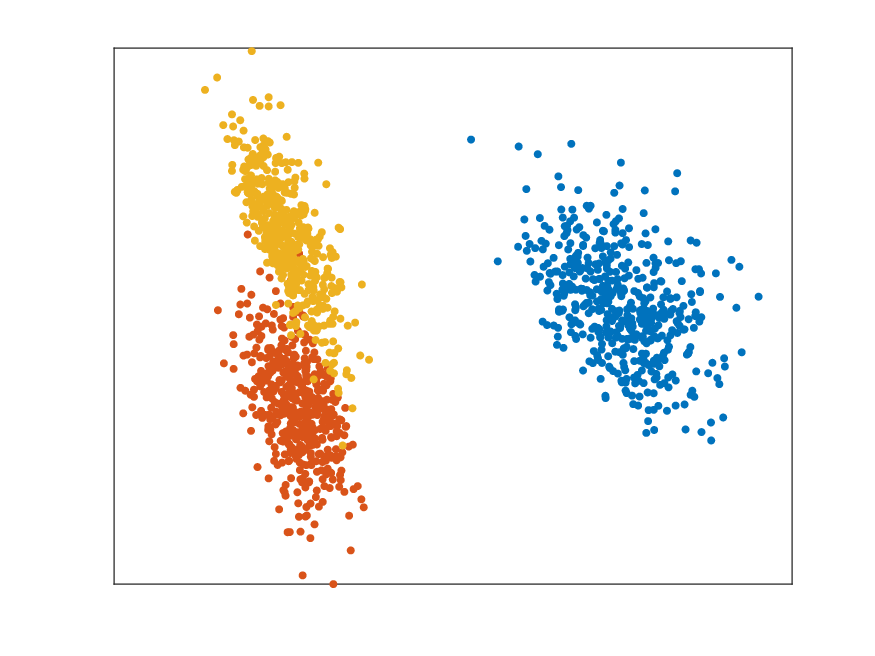}
\put(42,-5){ \small {\tt{Beans}} }
\end{overpic}}
\hspace{-2mm}
\subfigure{
\begin{overpic}[width=0.32\textwidth]{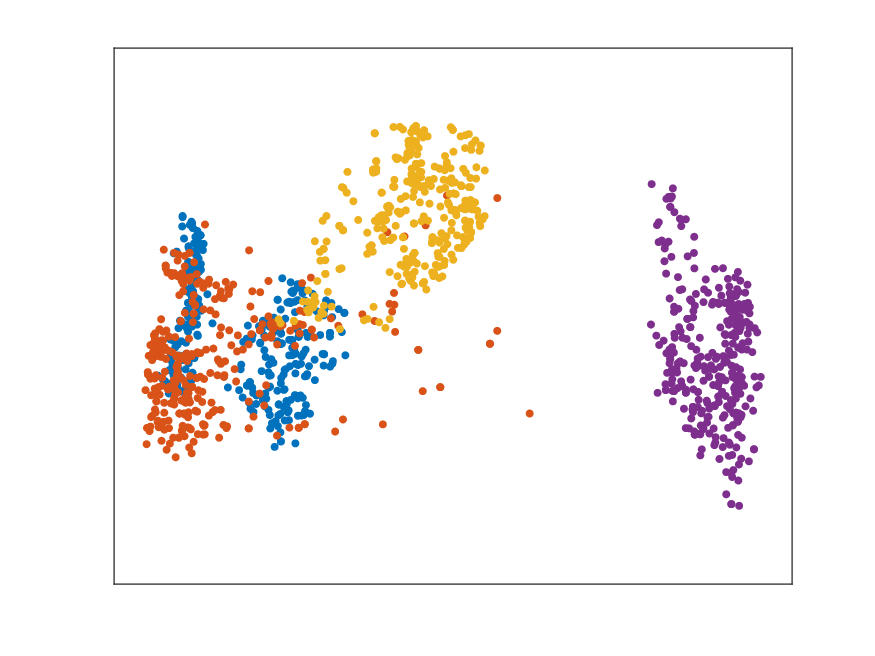}
\put(40,-5){ \small {\tt{Images}} }
\end{overpic}}
\caption{Scatter plots and simultaneous CIs.} 
\label{fig:DryBean}
\end{figure}

\noindent\textbf{Data.} The results are based on a synthetic dataset labeled as {\tt{Mixture}} and two natural datasets labeled as {\tt{Beans}} and {\tt{Images}}.  {\tt{Mixture}} was constructed from 1000 total samples, with 200 being drawn from each of five Gaussian distributions in $\R^6$ whose covariance matrices were all equal to the identity matrix, and whose mean vectors were $(0,0,0,0,0,0)$, $(5,5,5,0,0,0)$, $(0,5,5,5,0,0)$, $(0,0,5,5,5,0)$  and $(0,0,0,5,5,5)$.
Regarding the natural datasets, {\tt{Beans}} and {\tt{Images}} correspond to the Dry Bean and Image Segmentation datasets from the UCI Machine Learning Repository~\citep{uci_ml_repo}. {\tt{Beans}} is derived from a set of
beans (observations) in 7 categories, with all beans having 16 associated features. We extracted the observations from 3 categories, ``Bombay'', ``Dermason'' and ``Seker'', and uniformly sampled 500 observations from each of these categories. Lastly,  {\tt{Images}} is based on a collection of outdoor images in 7 categories, with each image having 19 features. We extracted all the observations from 4 categories, ``brickface'', ``foliage'', ``path'' and ``sky'' images, with all of these categories having the same number of 330 observations. For each dataset, we applied the rescale() function in MATLAB with the default settings. Lastly, we calculated the two top eigenvectors from the sample covariance matrix of each dataset, and plotted the observations in 2 dimensions based on their coordinates with respect to these eigenvectors, as shown in Figure~\ref{fig:DryBean}.

\noindent\textbf{Graphs.} We adopted a commonly used approach to spectral clustering that involves assigning a vertex to each observation, and assigning a weighted edge to each pair of observations $x$ and $x'$, where the weight value is given by the Gaussian kernel $\exp(-\frac{1}{2 \delta^2}\|x-x'\|_2^2)$. (The bandwidth parameter $\delta$ was set to 0.2 for {\tt{Mixture}} and 0.3 for both {\tt{Beans}} and {\tt{Images}}.) In this way, each of the three datasets above induces a fully connected graph $G$ with associated Laplacian $L$. In particular, the pairs of values $(n,|E|)$ for the numbers of vertices and edges are  (1000,499500) for {\tt{Mixture}}, (1500,1124250) for {\tt{Beans}}, and (1320,870540) for {\tt{Images}}. 

\noindent\textbf{Experiment design.} 
Since ER sampling and AER sampling are specifically designed to preserve spectral properties of $L$~\citep{srivastava2011}, whereas EW sampling is not, we focused on ER sampling and AER sampling. (See Appendix~\ref{sec:Additionaldetails} for background on AER.)

For each of the graphs associated with {\tt{Mixture}}, {\tt{Beans}}, and {\tt{Images}}, we generated 1000 realizations of $\hat L$ using both ER and AER sampling, and employed the method outlined in Section \ref{sec:method:CI} to construct simultaneous CIs for $\lambda_1(L)  \leq  \cdots \leq \lambda_{15}(L)$. Table \ref{table:clustering} shows the observed simultaneous coverage probabilities $\P(\bigcap_{j=1}^{15}\{\lambda_j(L)\in\hat{\mathcal{I}}_j\})$ based on desired confidence levels of 90\% and 95\%, which were computed by averaging over the 1000 trials in each setting. A display of the CIs and eigenvalues of $L$ are given in Figure~\ref{fig:DryBean}. For clarity of presentation, we only plotted a single representative CI at each index, corresponding to one whose center was nearly equal to the median of the centers of all the 1000 intervals. Also, for clarity, Figure~\ref{fig:DryBean} only displays the intervals corresponding to the 7 bottom Laplacian eigenvalues and a confidence level of 95\%.

\noindent\textbf{Discussion of empirical results.} An intended feature of the experiments is that the clustering problems corresponding to {\tt{Mixture}}, {\tt{Beans}}, and {\tt{Images}} have increasing levels of difficulty (as can be seen in Figure~\ref{fig:DryBean}), which allows us to see the performance of the CIs in a range of conditions. Table \ref{table:clustering} shows that the observed simultaneous coverage probabilities agree well with the desired confidence levels in all three problems. Also, the largest gaps among the CIs coincide with the correct number of clusters in all three problems---which demonstrates that the intervals can provide practical guidance to users in selecting the number of clusters. An especially good illustration of this occurs in the case of {\tt{Beans}}, where there are large gaps between the \emph{centers} of the 5th, 6th, and 7th CIs, but the gaps between their relevant \emph{endpoints} are much smaller. In other words, this is a case where a user might be tempted to conclude that 5 or 6 clusters are present based only on the eigenvalues of $\hat L$ (i.e.~when error estimation is not used), whereas the CIs guard against these incorrect conclusions.

\begin{table}[H]
\caption{Observed simultaneous coverage probabilities $\P(\bigcap_{j=1}^{15}\{\lambda_j(L)\in\hat{\mathcal{I}}_j\})$.}
\centering
\setlength\tabcolsep{3pt}
\begin{tabular}{lllllll}
\hline
                   &                           \multicolumn{2}{c}{ER} &  & \multicolumn{2}{c}{AER}  \\ \cline{2-3} \cline{5-6}  
Dataset & 90th  & 95th && 90th  & 95th \\ \hline
\specialrule{0em}{1pt}{1pt}
\multirow{1}{*}{{\tt{Mixture}}}                         &        90.3 &95.7 & &91.6 &95.0 \\{\tt{Beans}}&90.2 &95.0&& 93.3 &97.2          \\{\tt{Images}}&90.0 &94.2   &&89.2 &94.3         \\
\specialrule{0em}{1pt}{1pt}
\hline
\end{tabular}
\label{table:clustering}
\end{table}

\normalsize

\section{Empirical results on incremental refinement}
\label{app:extrap}
In this section, we illustrate the performance of the incremental refinement technique discussed in Section \ref{sec:computation}.  The experiments are based on the graphs {\tt{Genes}}, {\tt{M14}}, {\tt{Howard}}, {\tt{Citations}}, and {\tt{DIMACS}}, as well as the same choices of $\psi$ and sampling schemes covered in Section \ref{sec:expt}. To reduce the number of plots, only the confidence level $1-\alpha=95\%$ was considered.

\noindent\textbf{Experiment design.} Here, we follow the notation introduced in Section~\ref{sec:computation}. For each graph and sampling scheme, we generated 1000 sparsified Laplacians $\hat L$ based on $N_0 = 0.02|E|$ sampled edges, and applied Algorithm 1 to obtain 1000 corresponding quantile estimates $\hat q_{1-\alpha}(N_0)$. To construct estimates $\hat q_{1-\alpha}(N)$ for all $N\geq N_0$ by extrapolating from $\hat q_{1-\alpha}(N_0)$, we used the rule defined by $\hat q_{1-\alpha}(N)=\sqrt{N_0/N}\hat q_{1-\alpha}(N_0)$, which is based on the intuition that fluctuations of the entries of $\hat L$ should have a $1/\sqrt{N}$ scaling with respect to $N$. We refer to~\citep{Bickel:Extrapolation} for further background on the use of extrapolation rules to reduce the cost of bootstrapping.

\begin{figure}[H]
\setlength{\abovecaptionskip}{10pt}
\centering 
\subfigure{
\begin{overpic}[width=0.32\textwidth]{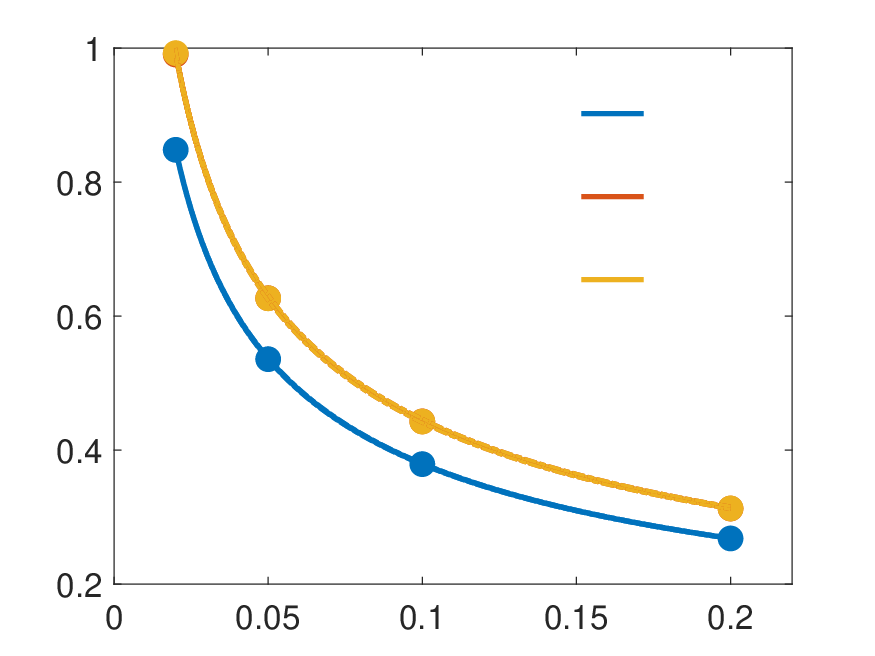}
\put(-4,12){\rotatebox{90}{ {  \ \ \ \ \small  $\hat q_{1-\alpha}(N)$  \ \ }}}
\put(74,60){ \tiny \tiny  EW}
\put(74,51){ \tiny \tiny  ER}
\put(74,42){ \tiny \tiny  AER}
\put(35,74){ \small $\|\hat L - L \|_F$}
\put(40,-5){ $N/|E|$}
\end{overpic}}
\hspace{-2mm}
\subfigure{
\begin{overpic}[width=0.32\textwidth]{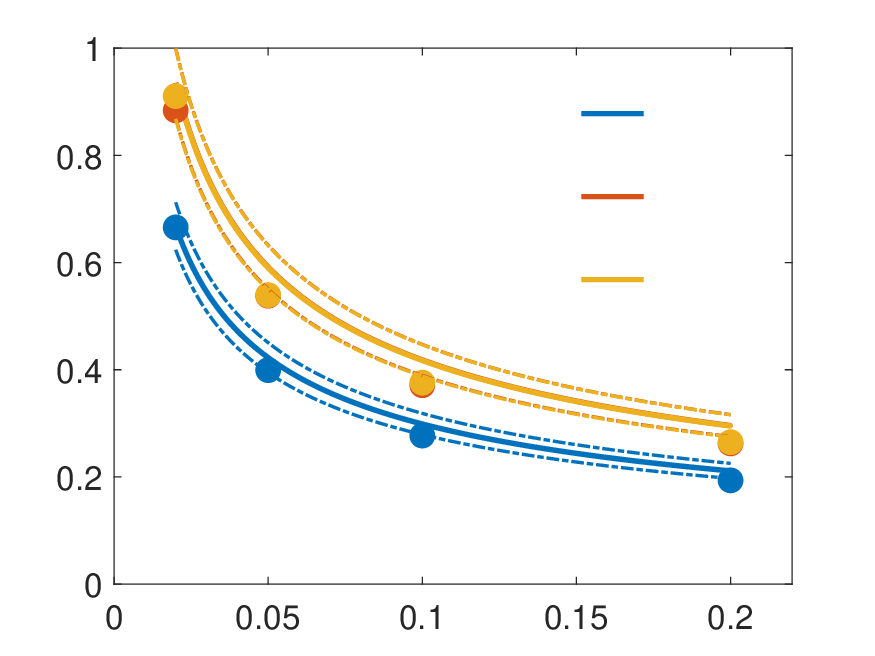}

\put(74,60){ \tiny \tiny  EW}
\put(74,51){ \tiny \tiny  ER}
\put(74,42){ \tiny \tiny  AER}
\put(35,74){ \small $\|\hat L - L \|_{\textup{op}}$}
\put(40,-5){ $N/|E|$}
\end{overpic}}
\hspace{-2mm}
\subfigure{
\begin{overpic}[width=0.32\textwidth]{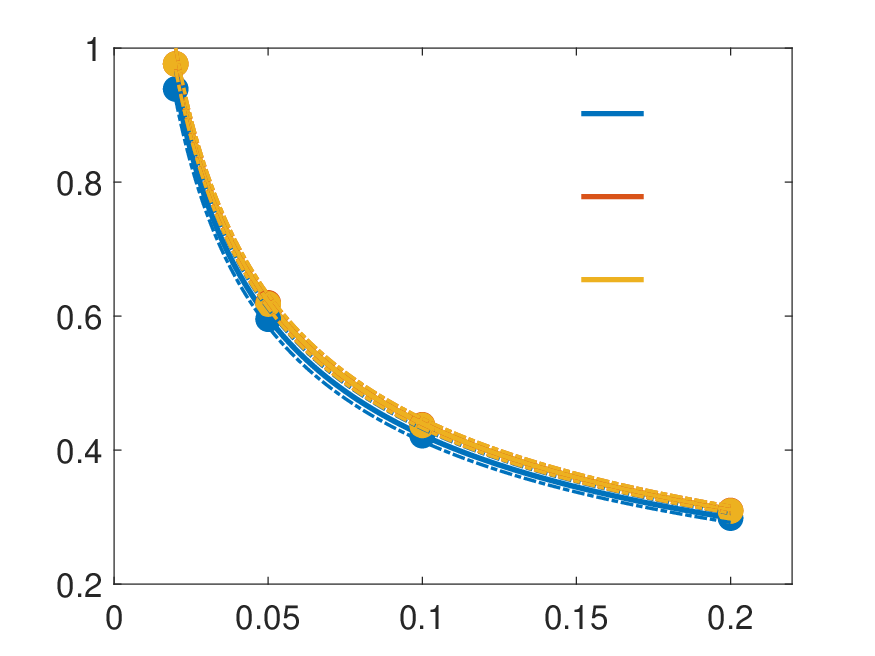}

\put(74,60){ \tiny \tiny  EW}
\put(74,51){ \tiny \tiny  ER}
\put(74,42){ \tiny \tiny  AER}
\put(30,74){ \small $\|r(\hat L) - r( L) \|_{2}$}
\put(40,-5){ $N/|E|$}
\end{overpic}}
\caption{Results on incremental refinement for {\tt{Genes}}.} 
\label{fig:Genes}
\end{figure}

\begin{figure}[H]
\setlength{\abovecaptionskip}{10pt}
\centering 
\subfigure{
\begin{overpic}[width=0.32\textwidth]{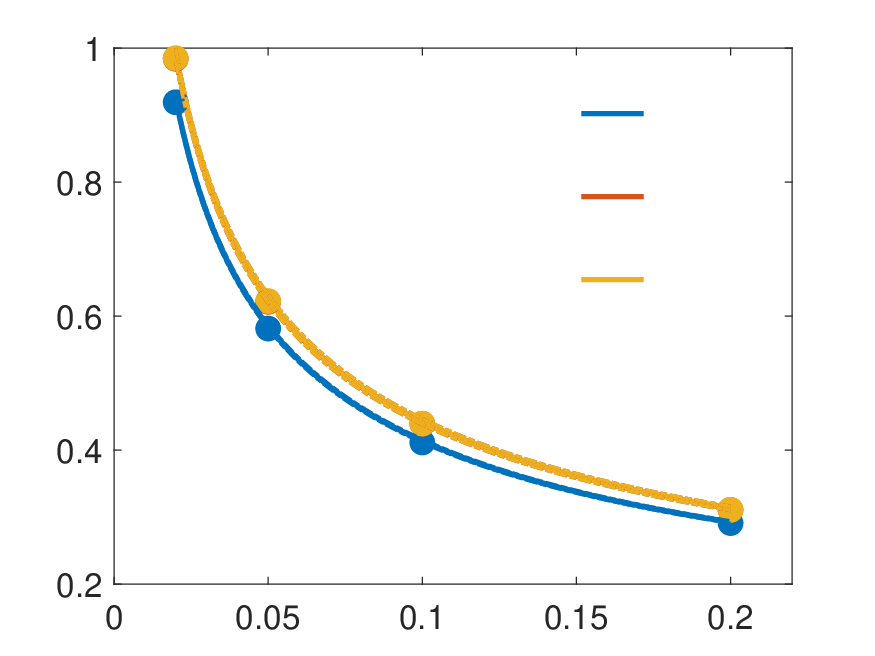}
\put(-4,12){\rotatebox{90}{ {  \ \ \ \ \small  $\hat q_{1-\alpha}(N)$  \ \ }}}
\put(74,60){ \tiny \tiny  EW}
\put(74,51){ \tiny \tiny  ER}
\put(74,42){ \tiny \tiny  AER}
\put(35,74){ \small $\|\hat L - L \|_F$}
\put(40,-5){ $N/|E|$}
\end{overpic}}
\hspace{-2mm}
\subfigure{
\begin{overpic}[width=0.32\textwidth]{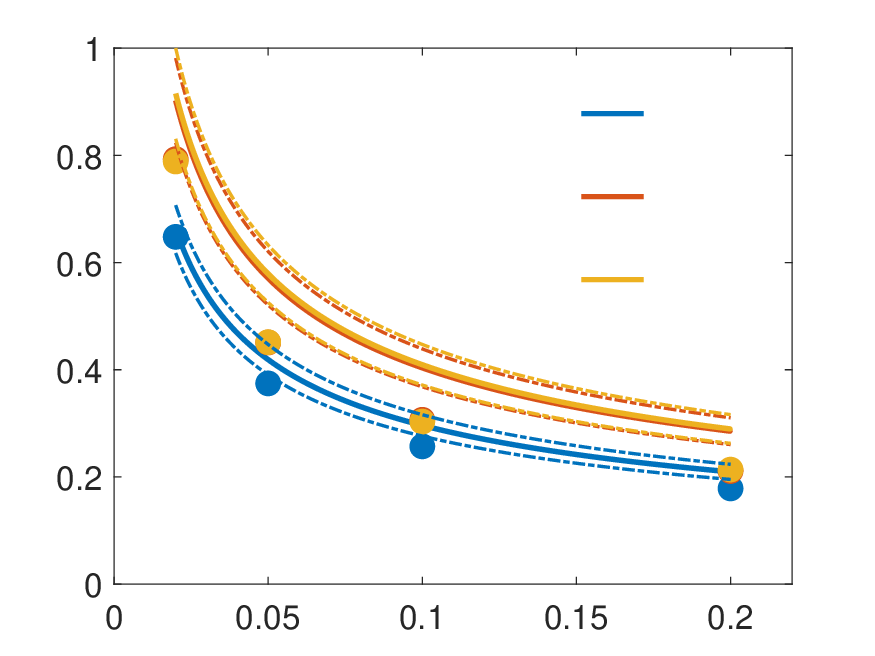}
\put(74,60){ \tiny \tiny  EW}
\put(74,51){ \tiny \tiny  ER}
\put(74,42){ \tiny \tiny  AER}
\put(35,74){ \small $\|\hat L - L \|_{\textup{op}}$}
\put(40,-5){ $N/|E|$}
\end{overpic}}
\hspace{-2mm}
\subfigure{
\begin{overpic}[width=0.32\textwidth]{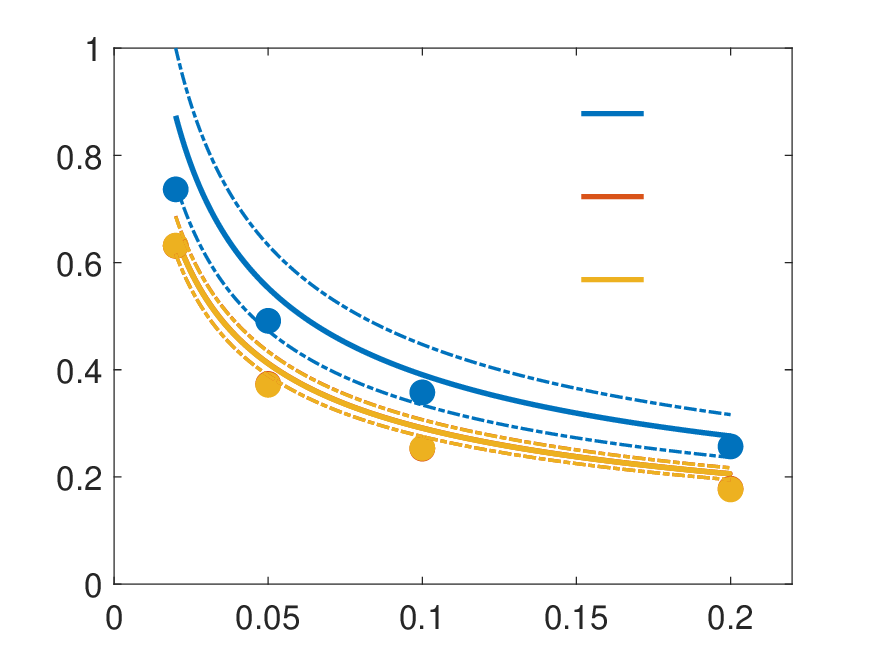}
\put(74,60){ \tiny \tiny  EW}
\put(74,51){ \tiny \tiny  ER}
\put(74,42){ \tiny \tiny  AER}
\put(30,74){ \small $\|r(\hat L) - r( L) \|_{2}$}
\put(40,-5){ $N/|E|$}
\end{overpic}}
\caption{Results on incremental refinement for  {\tt{M14}}.} 
\label{fig:M14}
\end{figure}

\begin{figure}[H]
\setlength{\abovecaptionskip}{10pt}
\centering 
\subfigure{
\begin{overpic}[width=0.32\textwidth]{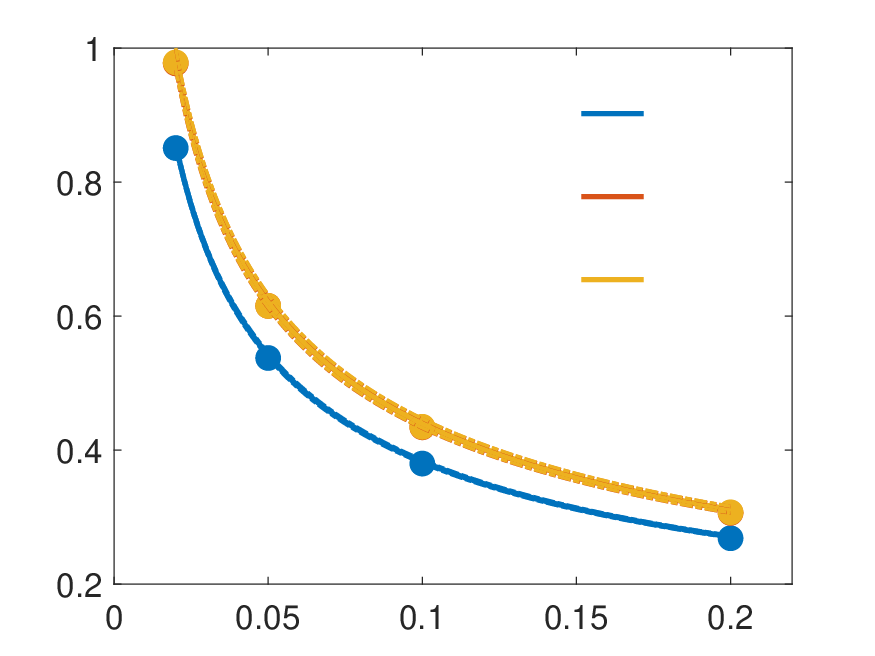}
\put(-4,12){\rotatebox{90}{ {  \ \ \ \ \small  $\hat q_{1-\alpha}(N)$  \ \ }}}
\put(74,60){ \tiny \tiny  EW}
\put(74,51){ \tiny \tiny  ER}
\put(74,42){ \tiny \tiny  AER}
\put(35,74){ \small $\|\hat L - L \|_F$}
\put(40,-5){ $N/|E|$}
\end{overpic}}
\hspace{-2mm}
\subfigure{
\begin{overpic}[width=0.32\textwidth]{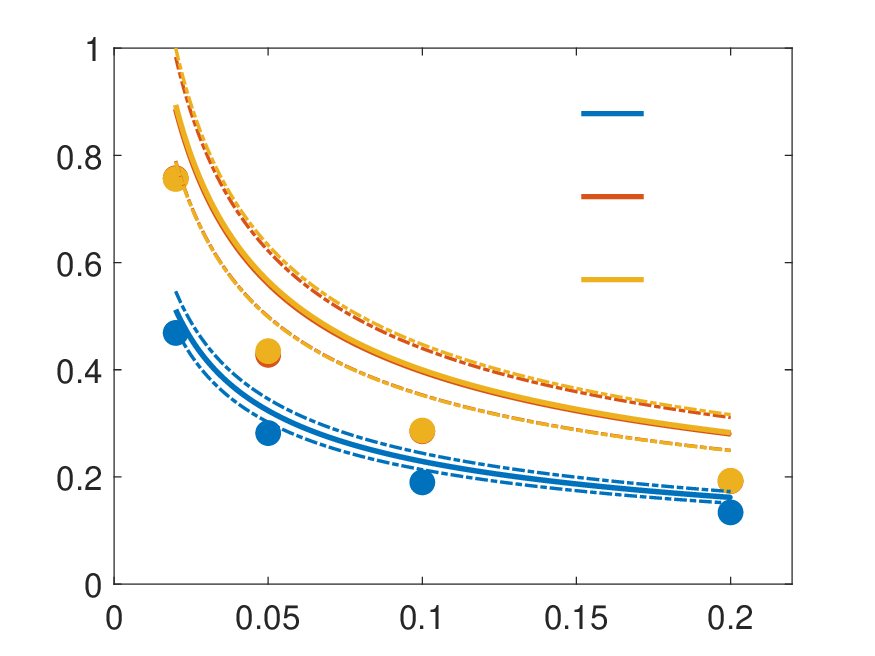}
\put(74,60){ \tiny \tiny  EW}
\put(74,51){ \tiny \tiny  ER}
\put(74,42){ \tiny \tiny  AER}
\put(35,74){ \small $\|\hat L - L \|_{\textup{op}}$}
\put(40,-5){ $N/|E|$}
\end{overpic}}
\hspace{-2mm}
\subfigure{
\begin{overpic}[width=0.32\textwidth]{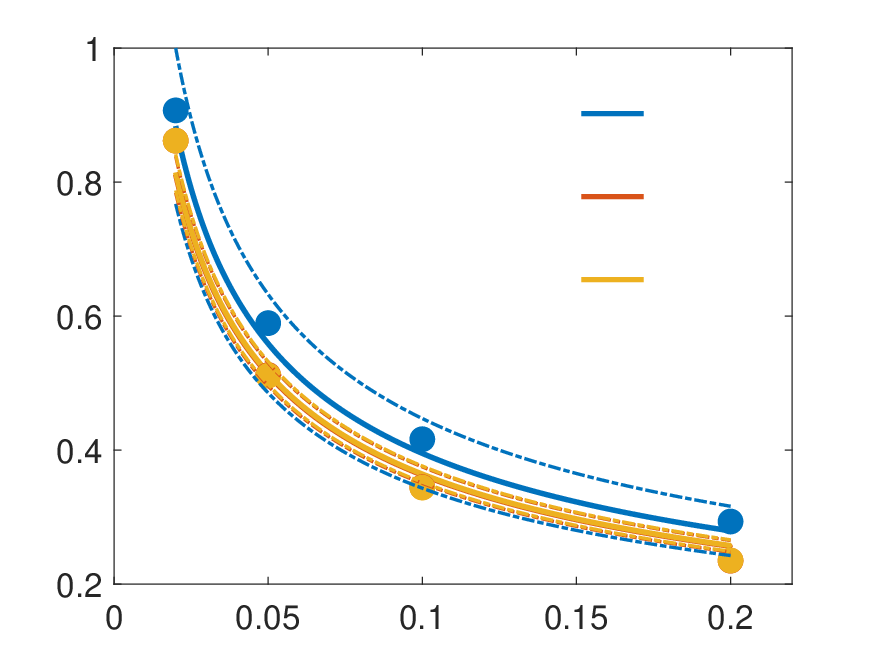}
\put(74,60){ \tiny \tiny  EW}
\put(74,51){ \tiny \tiny  ER}
\put(74,42){ \tiny \tiny  AER}
\put(30,74){ \small $\|r(\hat L) - r( L) \|_{2}$}
\put(40,-5){ $N/|E|$}
\end{overpic}}
\caption{Results on incremental refinement for  {\tt{Howard}}.} 
\label{fig:Howard}
\end{figure}

\begin{figure}[H]
\setlength{\abovecaptionskip}{10pt}
\centering 
\subfigure{
\begin{overpic}[width=0.32\textwidth]{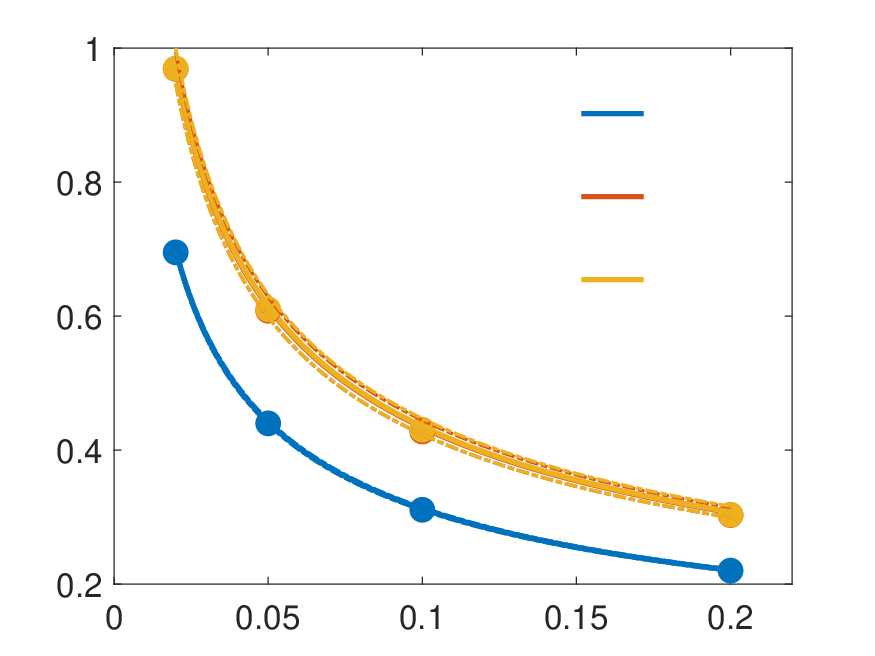}
\put(-4,12){\rotatebox{90}{ {  \ \ \ \ \small  $\hat q_{1-\alpha}(N)$  \ \ }}}
\put(74,60){ \tiny \tiny  EW}
\put(74,51){ \tiny \tiny  ER}
\put(74,42){ \tiny \tiny  AER}
\put(35,74){ \small $\|\hat L - L \|_F$}
\put(40,-5){ $N/|E|$}
\end{overpic}}
\hspace{-2mm}
\subfigure{
\begin{overpic}[width=0.32\textwidth]{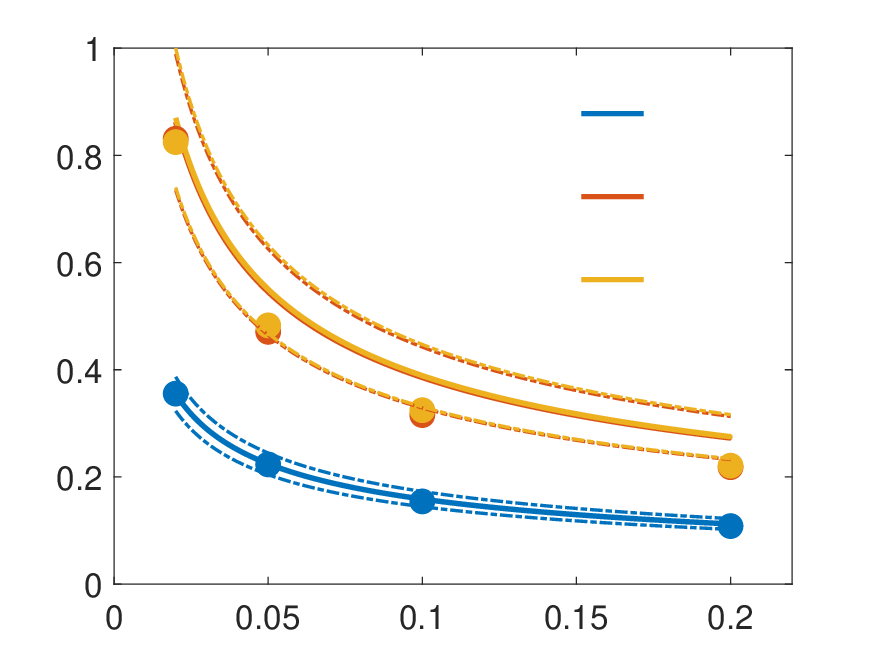}
\put(74,60){ \tiny \tiny  EW}
\put(74,51){ \tiny \tiny  ER}
\put(74,42){ \tiny \tiny  AER}
\put(35,74){ \small $\|\hat L - L \|_{\textup{op}}$}
\put(40,-5){ $N/|E|$}
\end{overpic}}
\hspace{-2mm}
\subfigure{
\begin{overpic}[width=0.32\textwidth]{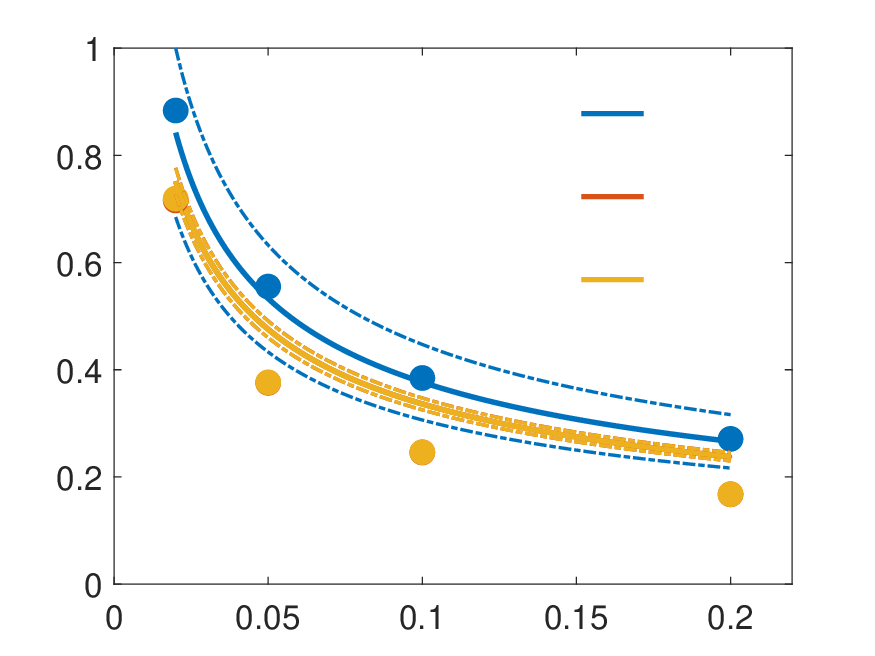}
\put(74,60){ \tiny \tiny  EW}
\put(74,51){ \tiny \tiny  ER}
\put(74,42){ \tiny \tiny  AER}
\put(30,74){ \small $\|r(\hat L) - r( L) \|_{2}$}
\put(40,-5){ $N/|E|$}
\end{overpic}}
\caption{Results on incremental refinement for {\tt{Citation}}.} 
\label{fig:citation}
\end{figure}

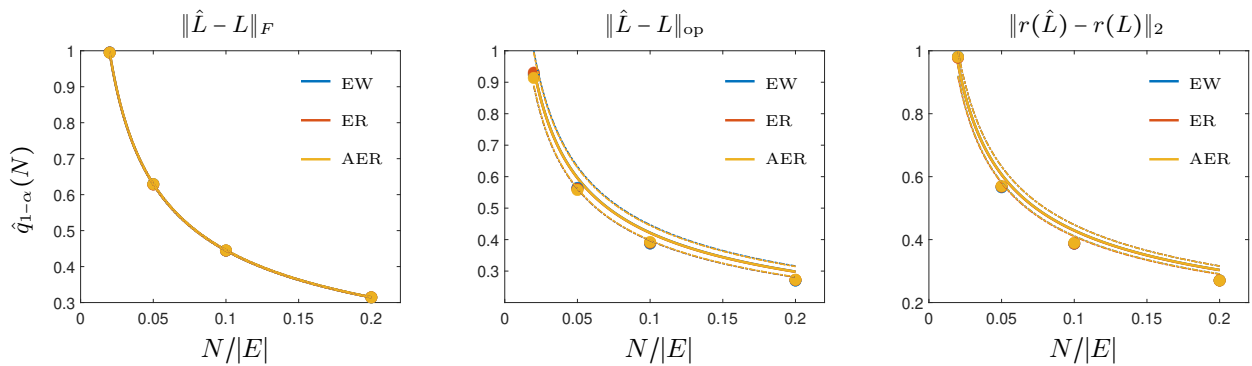
\begin{figure}[H]
\setlength{\abovecaptionskip}{10pt}
\centering 
\subfigure{
\begin{overpic}[width=0.32\textwidth]{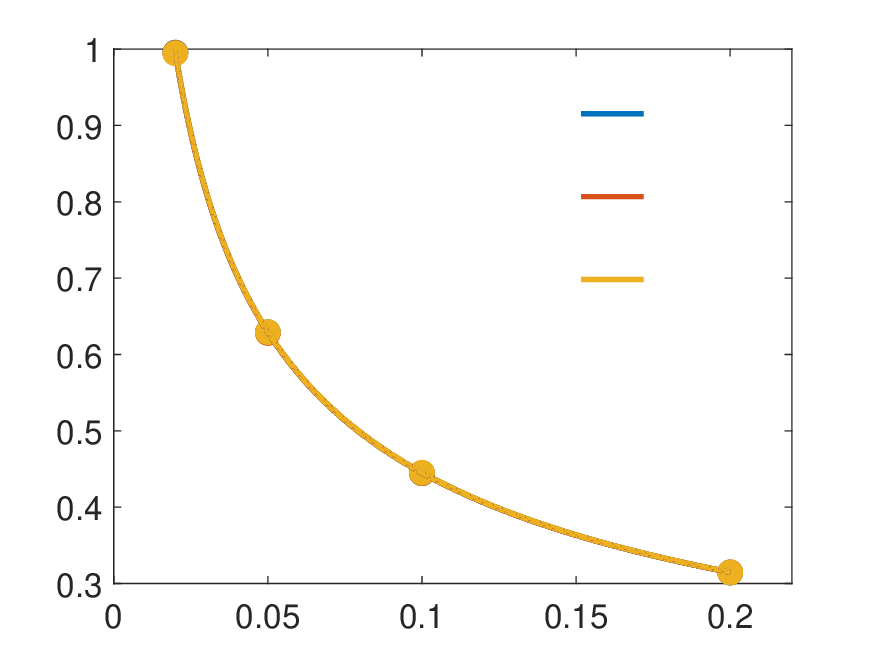}
\put(-4,12){\rotatebox{90}{ {  \ \ \ \ \small  $\hat q_{1-\alpha}(N)$  \ \ }}}
\put(74,60){ \tiny \tiny  EW}
\put(74,51){ \tiny \tiny  ER}
\put(74,42){ \tiny \tiny  AER}
\put(35,74){ \small $\|\hat L - L \|_F$}
\put(40,-5){ $N/|E|$}
\end{overpic}}
\hspace{-2mm}
\subfigure{
\begin{overpic}[width=0.32\textwidth]{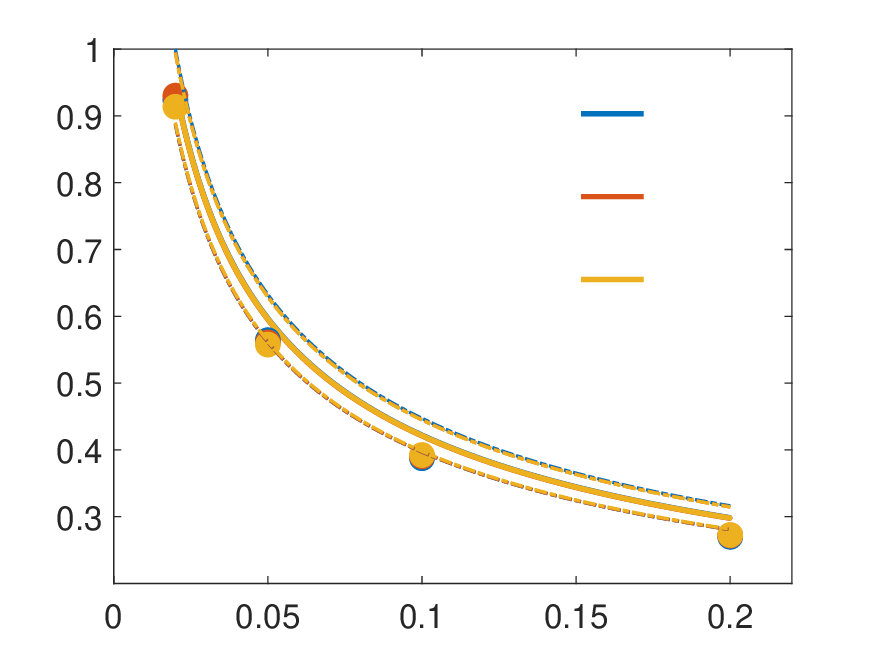}
\put(74,60){ \tiny \tiny  EW}
\put(74,51){ \tiny \tiny  ER}
\put(74,42){ \tiny \tiny  AER}
\put(35,74){ \small $\|\hat L - L \|_{\textup{op}}$}
\put(40,-5){ $N/|E|$}
\end{overpic}}
\hspace{-2mm}
\subfigure{
\begin{overpic}[width=0.32\textwidth]{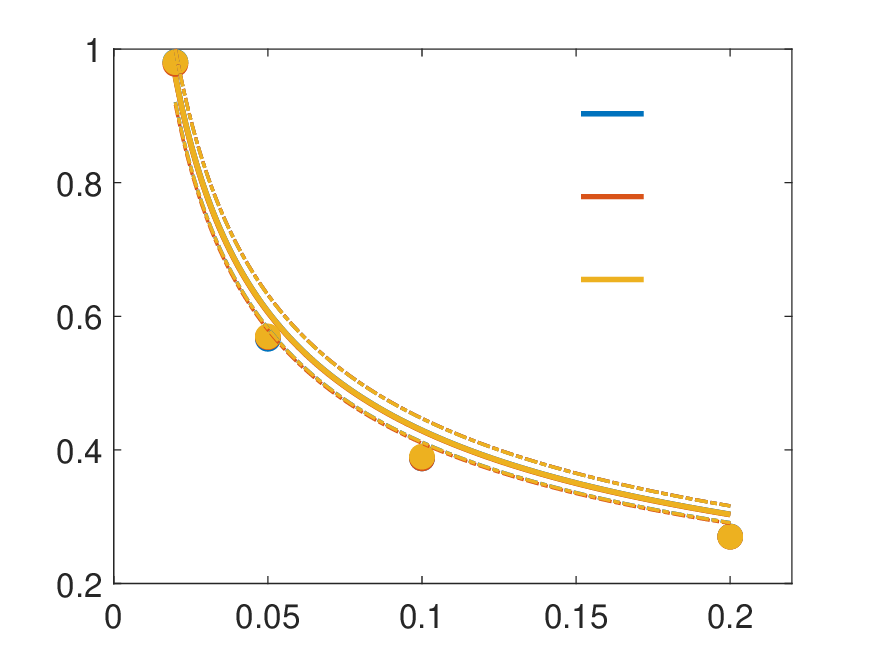}
\put(74,60){ \tiny \tiny  EW}
\put(74,51){ \tiny \tiny  ER}
\put(74,42){ \tiny \tiny  AER}
\put(30,74){ \small $\|r(\hat L) - r( L) \|_{2}$}
\put(40,-5){ $N/|E|$}
\end{overpic}}
\caption{Results on incremental refinement for {\tt{DIMACS}}.} 
\label{fig:DIMACS}
\end{figure}

In all cases, the average of $\hat q_{1-\alpha}(N)$ over all 1000 trials is plotted in Figures~\ref{fig:Genes}-\ref{fig:DIMACS} as a function of $N$ using a solid line, where $N$ ranges between $0.02|E|$ and $0.2|E|$. The variability $\hat q_{1-\alpha}(N)$ is indicated by dashed lines, which are plotted 1 standard deviation above and below the solid curve. (Note that for the ER and AER sampling schemes, the curves tend to overlap in many cases, making only the curves for AER visible.) Also, all the plots were put on a common scale by dividing all curves in a given plot by the value of the highest curve at $N_0=0.02|E|$. 
Lastly, as a substitute for the ground truth value of $q_{1-\alpha}(N)$, we computed the empirical 95\% quantile of the 1000 values of $\psi(\hat L, L)$ at $N\in\{0.05|E|,0.1|E|,0.2|E|\}$, and these values are marked with large dots.

\noindent\textbf{Discussion of empirical results.} 
The accuracy of the extrapolated estimates $\hat q_{1-\alpha}(N)$ is judged by how well the curves agree with the large dots of the same color. Overall, Figures~\ref{fig:Genes}-\ref{fig:DIMACS} show that the estimates perform well, considering that in most cases the dots are within about one standard deviation of the corresponding solid curve. The stability of the estimates is also notable, as the standard deviation is generally small in proportion to the height of the solid curve. Lastly, and perhaps most importantly, the curves remain accurate up to $N=0.2|E|$ even though they were extrapolated from a sample size $N_0=0.02|E|$ that is \emph{10 times smaller}. This indicates that the incremental refinement technique has the potential to substantially improve computational efficiency, because error estimation can be performed more quickly when the number of sampled edges is small.

\normalsize

\section{Additional details on experiments}
\label{sec:Additionaldetails}

\noindent\textbf{Details of the graphs used in Section \ref{sec:expt}.} The edges of {\tt{Genes}} and {\tt{Citations}} have varying weights, whereas the edges of the other graphs all have equal weights.

\begin{itemize}[leftmargin=*]

    \item ca-cit-HepTh ({\tt{Citations}}) \citep{rossi2015network}:  
    This graph represents the co-citation of scientific papers from arXiv's high energy physics-theory (HEP-TH) section, involving 22,908 vertices and 2,444,798 edges. An edge between two papers means that both papers have been cited by a common third paper ~\citep{konect}. 
    \item C2000-9 ({\tt{DIMACS}}) \citep{rossi2015network}: This graph is from the DIMACS Implementation Challenge ~\citep{bader201110th}, 
    consisting of 2,000 nodes and 1,799,532 edges.
        \item human-gene2 ({\tt{Genes}}) \citep{davis2011university}: This graph is a human gene regulatory network, with 14,340 vertices and 9,041,364 edges.
        \item FB-Howard90 ({\tt{Howard}}) \citep{rossi2015network}: 
    This graph is a social network graph constructed based on Howard University Facebook data~\citep{red2011comparing,traud2012social}. All friendships are represented as undirected links. The graph contains 4,047 vertices and 204,850 edges.
    \item Mycielskian14 ({\tt{M14}}) \citep{davis2011university}: 
    This graph is part of a test suite for benchmarking graph algorithms. It is triangle free with a chromatic number of 14, and contains 12,287 vertices and 1,847,756 edges~\citep{mycielski1955coloriage}.
    \item Mycielskian20 ({\tt{M20}}) \citep{davis2011university}: 
    This graph is part of a test suite for benchmarking graph algorithms. It is triangle free with a chromatic number of 20, and contains 786,431 vertices and 2,710,370,560 edges~\citep{mycielski1955coloriage}.
\end{itemize}

\noindent\textbf{Discussion of AER sampling.} 
Recall that effective-resistance sampling is based on edge probabilities of the form $p(e)\propto w(e)\tr(L^+\Delta_e)$, where $L^+$ is the Moore-Penrose inverse of $L$. Because it is often costly or infeasible to compute $L^+$, there has been substantial research interest in developing efficient ways to approximate these probabilities. Here, we discuss one approach that was proposed in~\cite{srivastava2011} and implemented in~\cite{EstimateER}. In a nutshell, the approximate effective-resistance sampling probabilities are of the form $p(e)\propto w(e)\tr(S\ttop S \Delta_e)$,
where $S$ is a $k\times n$ random matrix and $k$ is of order $\log(n)/\epsilon^2$ for some accuracy parameter $\epsilon>0$. The computation of the matrix $S$ combines random projections with repeated use of the Spielman and Teng solver~\citep{Spielman:2004}, but the precise details are beyond the scope of our work here. In Section~\ref{sec:expt} we set $\epsilon=0.01$, and in Appendix~\ref{sec:clustering} we set $\epsilon=1$. The reason for using $\epsilon=1$ in the second case is that it was the smallest choice for which some difference in the results for ER and AER could be observed.

\noindent\textbf{License information.} The University of Florida sparse matrix collection \citep{davis2011university} is under the CC BY 4.0 License, and the graphs from \cite{rossi2015network} are under a CC BY-SA License. {\tt{Beans}} and {\tt{Images}} from the UCI machine learning repository are both available under the CC BY 4.0 License.

\noindent\textbf{Computing resources.} The results presented in Table \ref{table:cutfunctionLap}, Appendix \ref{sec:clustering}, and Appendix \ref{app:extrap} were obtained using MATLAB on servers equipped with 32 CPUs and 216 GB of RAM. All of the experiments together consumed roughly 200 hours of computing time. The results in Section \ref{sec:expt} in the discussion of computational efficiency were obtained using a laptop with approximately 16 GB of RAM, 8 physical cores, and 16 logical cores.

\section{Notation and conventions in proofs}
\label{sec:notation}
 The $L^q$ norm of a scalar random variable $V$ is denoted as $\|V\|_{L^q}= (\E(|V|^q))^{1/q}$. For any random object $V$, we use $\mathcal{L}(V)$ to refer to its distribution, while $\mathcal{L}(\cdot|Q)$, $\P(\cdot\,|Q)$ and $\E(\cdot\,|Q)$ refer to conditional distributions, probabilities and expectations given the random matrices $Q_1,\dots, Q_N$. Convergence in probability and convergence in distribution are respectively denoted by $\xrightarrow{\P}$ and $\xrightarrow{\mathcal{L}}$. The Kolmogorov metric is defined as $d_{\textup{K}}(\mathcal{L}(V),\mathcal{L}(W)) = \sup_{t\in \mathbb{R} } | \P(V\leq t) - \P(W \leq t)|$. In connection with this metric, we will sometimes use P\'olya's theorem \citep[Theorem B.7.7]{bickel2015mathematical}, which implies that if $\{V_n\}$ is a sequence of random variables satisfying $V_n\xrightarrow{\mathcal{L}}Z$ as $n\to\infty$ for a standard normal random variable $Z$, then $d_{\textup{K}}(\mathcal{L}(V_n),\mathcal{L}(Z))\to 0$ as $n\to\infty$.

For matrices $A, B \in \mathbb{R}^{n \times n}$, let  $\llangle A,B\rrangle = \tr(A\ttop B)$. For  $A \in \mathbb{R}^{n \times n}$, define $\|A\|_{\infty} = \max_{1\leq i,j \leq n} |A_{ij}|$.
For two sequences of non-negative real numbers ${a_n}$ and ${b_n}$, we write $a_n\lesssim b_n$ if there exists a constant $C>0$, independent of $n$, such that $a_n\leq Cb_n$ holds for all large $n$. If both $a_n\lesssim b_n$ hold and $b_n\lesssim a_n$ hold, then we write $a_n\asymp b_n$. The relation $a_n=o(b_n)$ means $a_n/b_n\to 0$ as $n\to\infty$, while $a_n=\mathcal{O}(b_n)$ is equivalent to $a_n\lesssim b_n$. For two sequences of random variables $\{U_n\}$ and $\{V_n\}$, the relation $U_n=o_{\P}(V_n)$ means that $U_n/V_n \xrightarrow{\P} 0$, and the relation $U_n= \mathcal{O}_{\P}(V_n)$ means that for every $\epsilon >0$, there exists a positive constant $C$ not depending on $n$ such that the inequality $\P(|U_n|/|V_n| \geq C) \leq \epsilon$ holds for all large $n$.
The indicator function for a condition $\cdots$ is represented as $1\{\cdots\}$.

Because the probabilities $\P(\bigcap_{x\in\mathcal{C}} \{C(x)\in \hat{\mathcal{I}}_{1-\alpha}(x)\})$ and $\P(\|\hat L-L\|_F^2\leq \hat q_{1-\alpha})$ in Theorems~\ref{thm:cutquery} and~\ref{thm:Frobenius} are both invariant to rescaling $L$ by a positive constant, we may always assume without loss of generality that all the edge weights sum to 1,
$$\sum_{e\in E}w(e)=1.$$
In this case, the edge weights $w(e)$ and sampling probabilities $p(e)$ are the same, and so the collection $\mathcal{Q}=\{(w(e)/p(e))\Delta_e|e\in E\}$ is the same as $\{\Delta_e|e\in E\}$. Furthermore, this means that the i.i.d.~random matrices $Q_1,\dots,Q_N$ satisfy $\P(Q_1=\Delta_e)=p(e)$ for all $e\in E$. As another simplification, the proofs will generally omit the subscript $n$ that was used in the statements of the theorems.

To introduce some further notation that will be used in the proofs, note that the symmetric rank-1 matrix $\Delta_e$ associated to an edge $e = \{i,j\}$ with $i<j$ can be written as $\Delta_e = \delta_e\delta_e\ttop$, where $\delta_e = u_i -u_j$, and $u_i$ denotes the $i$th standard basis vector. Also, let $\hat e_1, \ldots, \hat e_N$ be i.i.d samples drawn from $ E$ via edge-weight sampling, and define $D_{i} \ = \ \delta_{\hat e_i}$ for simplicity. In this case, the random matrix $\hat L$ can be represented as
\begin{align*}
    \hat L = \frac{1}{N} \sum_{i=1}^N D_i D_i\ttop.
\end{align*}
Let $\hat e_1^*, \ldots, \hat e_N^*$ be i.i.d samples uniformly drawn from $\hat e_1, \ldots, \hat e_N$, and $\hat e_1^{**}, \ldots, \hat e_N^{**}$ be i.i.d samples uniformly drawn from $\hat e_1^*, \ldots, \hat e_N^*$. Define 
\begin{align*}
    D_{i}^* \ = \ \delta_{\hat e_{i}^*}  \ \ \ \ \ \ \ \ \ \ \ \ \text{ and } \ \ \ \ \ \ \ \ \ \ \ \
    D_{i}^{**} \ = \ \delta_{\hat e_{i}^{**}}
\end{align*}
for $i=1,\ldots,N$, so that the random matrices $\hat L^*$ and $\hat L^{**}$ can be represented as 
\begin{align*}
    \hat L^* = \frac{1}{N} \sum_{i=1}^N D_{i}^*(D_{i}^*)\ttop  \ \ \ \ \ \ \text{ and }  \ \ \ \ \ \  \hat L^{**} = \frac{1}{N} \sum_{i=1}^N D^{**}_i(D^{**}_i)\ttop.
\end{align*}

\section{Proof of Theorem~\ref{thm:cutquery}}
\label{app:cut}
\setcounter{lemma}{0}
\renewcommand{\thelemma}{E.\arabic{lemma}}

Let the set of cut vectors be enumerated as $\sCc = \{x_1, \ldots, x_{|\mathcal{C}|}\}$. (Note that in the main text, we used $x_i$ to refer to the $i$th coordinate of a single vector $x\in\mathcal{C}$, but  that earlier usage will no longer be needed for cut vectors.) Also, for  $i,j \in\{ 1,\dots,|\sCc|\}$, define
\begin{equation}\label{eq:defofsij}
    \begin{split}
        &s_{ij} = \cov\Big(x_i\ttop Q_1 x_i\,,\,x_j\ttop Q_1 x_j\Big) \ = \ \sum_{e \in E} w(e) x_i\ttop \Delta_{e} x_i x_j\ttop \Delta_{e} x_j - x_i\ttop L x_i x_j\ttop L x_j\\
        &\hat s_{ij} \ = \ \cov\Big(x_i\ttop Q_{1}^* x_i\,,\,x_j\ttop  Q_{1}^* x_j\Big |Q\Big) \ = \ \frac{1}{N} \sum_{k=1}^N x_i\ttop  Q_{k} x_i x_j\ttop Q_{k} x_j - x_i\ttop \hat L x_i x_j\ttop \hat L x_j.
    \end{split}
\end{equation}
Next, define the random variables
\begin{align*}
    M \ = \  &\displaystyle \sqrt{N} \max_{ 1 \leq i \leq |\mathcal{C}|} |x_i \ttop \hat L x_i-x_i\ttop L x_i|/\sqrt{s_{ii}}, \\  
    \hat M \ = \  &\displaystyle \sqrt{N} \max_{ 1 \leq i \leq |\mathcal{C}|}   |x_i \ttop \hat L x_i-x_i\ttop L x_i|/\sqrt{\hat s_{ii}}, \\
    M^* \ = \ &\sqrt{N} \max_{1\leq i \leq |\mathcal{C}|} |x_i \ttop \hat L^* x_i - x_i\ttop \hat L x_i|/\sqrt{\hat s_{ii}}.
\end{align*}
Let $G =(G_1, \ldots G_{|\mathcal{C}|})$ be a Gaussian vector drawn from $N({0}, R)$, where $R_{ij} = \frac{s_{ij}}{\sqrt{s_{ii} s_{jj}}}$, and define 
\begin{equation*}
M( G) = \max_{1\leq i \leq |\mathcal{C}|} |G_i|.
\end{equation*}
Also, let $\hat G = (\hat G_1, \ldots \hat G_{|\mathcal{C}|})$ be a random vector that is drawn $N({0}, \hat R)$ conditionally on $Q_1,\dots,Q_N$, where $\hat R_{ij} = \frac{\hat s_{ij}}{\sqrt{ \hat s_{ii}\hat s_{jj}}}$, and define 
\begin{equation*}
M(\hat G)= \max_{1\leq i \leq |\mathcal{C}|}| \hat G_i|.
\end{equation*}

It follows from standard arguments in the bootstrap literature (e.g.~the proof of~\citep[][Lemma 10.4]{lopes2022central}) that Theorem~\ref{thm:cutquery} reduces to showing the following two limits as $n\to\infty$,
\begin{align*}
    & d_{\mathrm{K}}\big(\mathcal{L}( \hat M), \mathcal{L}(M(G))\big) \ \xrightarrow{} \ 0\\
    & d_{\mathrm{K}}\big(\mathcal{L}( M^*|Q), \mathcal{L}(M(G))\big) \ \xrightarrow{\P} \ 0.
\end{align*}
These two statements are shown in Lemmas \ref{lem:MG} and \ref{lem:M*G} respectively.
\begin{lemma}\label{lem:MG}
    If the conditions in Theorem~\ref{thm:cutquery} hold, then as $n\to\infty$,
    \begin{align*}
        d_{\mathrm{K}}\big(\mathcal{L}( \hat M), \mathcal{L}(M( G))\big) \ \xrightarrow{} \ 0.
    \end{align*}
\end{lemma}
\proof
Since the triangle inequality gives
\begin{equation*}
    \begin{split}
        d_{\mathrm{K}}\big(\mathcal{L}( \hat M), \mathcal{L}(M( G))\big) 
    \ \leq \
    & d_{\mathrm{K}}\big(\mathcal{L}( \hat M), \mathcal{L}(M)\big) \ + \  d_{\mathrm{K}}\big(\mathcal{L}( M), \mathcal{L}(M( G))\big),
    \end{split}
\end{equation*}
we will handle the terms on the right side separately. To handle the second term on the right, we will apply Lemma \ref{lem:CCK} 
to establish a Gaussian approximation for $M$. Specifically, we will apply this lemma to a set of i.i.d.~random vectors $X_1,\dots,X_N\in\R^{|\mathcal{C}|}$, where the $j$th component of the $i$th vector is defined by $X_{ij}=(x_j\ttop Q_ix_j - x_j\ttop L x_j)/\sqrt{s_{jj}}$. Also note that $x_j\ttop Q_1 x_j$ is a Bernoulli$(x_j\ttop L x_j)$ random variable for all $j=1,\dots,|\mathcal{C}|$. In the notation of Lemma \ref{lem:CCK}, we will put $p=|\mathcal{C}|$,
$b_1 =b_2 = 1$ and $c_n = 2/(\log(2)\sqrt{\eta(\sCc)})$. By noting the inequalities $0 \leq x_j\ttop Q_ix_j \leq 1$, and $0 \leq x_j\ttop L x_j \leq 1$, as well as the fourth central moment formula
\begin{align}\label{eq:4thcentralmoment}
    \E\big((x_i\ttop Q_1 x_i - x_i\ttop L x_i)^4\big) 
    \ = \ x_i\ttop L x_i( 1 - x_i\ttop L x_i )( 1- 3x_i\ttop L x_i( 1- x_i\ttop L x_i) ),
\end{align}
 it is possible to check that the three conditions in Lemma \ref{lem:CCK} hold under these choices of $b_1$, $b_2$, and $c_n$. Consequently, the lemma gives
\begin{align*}
    d_{\mathrm{K}}\big(\mathcal{L}( M), \mathcal{L}(M( G))\big) \ \lesssim \
    \Big( \frac{\log(N|\sCc|)^5}{N \eta(\sCc) }\Big)^{1/4}.
\end{align*}
So, under the conditions in Theorem~\ref{thm:cutquery}, it follows that as $n\to\infty$,
\begin{align}\label{eq:D1}
    d_{\mathrm{K}}\big(\mathcal{L}( M), \mathcal{L}(M(G))\big) \ = \ o(1).
\end{align}
To analyze $d_{\mathrm{K}}\big(\mathcal{L}(M), \mathcal{L}(\hat M )\big)$, we will use the basic fact that the Kolmogorov distance between any two random variables $U$ and $V$ can be bounded as
\begin{equation}\label{eqn:dKdecomp}
    d_{\textup{K}}(\mathcal{L}(U),\mathcal{L}(V)) \ \leq \ \P\big(|U-V|>\epsilon\big) \ + \ \sup_{r\in\R}\P\big(|U-r|\leq \epsilon\big)
\end{equation}
for any $\epsilon>0$. Specifically, we will take $U=M$, $V = \hat M$, and $\epsilon=\log(N|\mathcal{C}|)^2/ \sqrt{N\eta(\mathcal{C})}$.
To handle the second term on the right side of~\eqref{eqn:dKdecomp}, we will use the assumptions in Theorem~\ref{thm:cutquery}, in conjunction with Lemmas \ref{lem:Nazarov} and \ref{lem:anticon} as well as the limit \eqref{eq:D1} to conclude that
\begin{align*}
     \sup_{r \in \mathbb{R} }\P\Big(|M -r| \leq \ts   \frac{\log(N|\mathcal{C}|)^2}{ \sqrt{N\eta(\mathcal{C})}} \Big) \ \leq \  &\sup_{r \in \mathbb{R} }\P\Big(|M(G) -r| \leq \ts \frac{\log(N|\mathcal{C}|)^2}{ \sqrt{N\eta(\mathcal{C})}} \Big) +  o(1) \\ 
     \ \lesssim \ &  \frac{\log(N|\mathcal{C}|)^{5/2}}{ \sqrt{N\eta(\mathcal{C})}} +  o(1)\\
     \ = \ & o(1).
\end{align*}
For the first term on the right side of~\eqref{eqn:dKdecomp}, note that 
\begin{equation}\label{eq:coupling}
    \begin{split}
        \P\Big(|M   -  \hat M|> \ts \frac{\log(N|\mathcal{C}|)^2}{ \sqrt{N\eta(\mathcal{C})}} \Big) \ \leq \ 
        & \ \P\Big(M  \displaystyle \max_{ 1 \leq i \leq |\mathcal{C}| } | \ts \frac{\sqrt{s_{ii}}}{\sqrt{\hat s_{ii}}} -1| >  \frac{\log(N|\mathcal{C}|)^2}{ \sqrt{N\eta(\mathcal{C})}} \Big)\\
        \ \leq \ 
        & \ \P\big( M  > 4\sqrt{\log(N |\mathcal{C}|)}  \big)  +  \P\Big( \displaystyle \max_{ 1 \leq i \leq |\mathcal{C}| } | \ts \frac{\sqrt{s_{ii}}}{\sqrt{\hat  s_{ii}}} -1|  > \frac{\log(N|\mathcal{C}|)^{3/2}}{ 4\sqrt{N\eta(\mathcal{C})}}\Big).
    \end{split}
\end{equation}
Combining the limit \eqref{eq:D1} with a union bound, we have
\begin{equation}\label{eq:OP}
    \begin{split}
        \P\big( M  > 4\sqrt{\log(N |\mathcal{C}|)}  \big) 
         \ \leq \  & \ \P\Big(M(G) > 4\sqrt{\log(N |\mathcal{C}|)}\Big)  + o(1) \\
         \ \leq \ & \sum_{i=1}^{|\mathcal{C}|} \P\Big(|G_i|>4\sqrt{\log(N|\mathcal{C}|})\Big) \ + \ o(1).\\
         \ \lesssim \  & \frac{1}{N}+o(1).
    \end{split}
\end{equation}
To handle the second term on the right side of \eqref{eq:coupling}, note that  $\hat s_{ii}$ can be represented as
\begin{align*}
    \hat s_{ii} \ = \ &\frac{1}{N^2} \sum_{1 \leq k < j \leq N}  (x_i \ttop Q_k x_i - x_i \ttop Q_j x_i)^2
\end{align*}
and $\E((x_i \ttop Q_k x_i - x_i \ttop Q_j x_i)^2) =s_{ii} =x_i\ttop L x_i(1-x_i\ttop L x_i)$. Since $\var( x\ttop Q_1 x(1- x\ttop L x))\leq 1$ holds for any $x \in \mathcal{C}$, a concentration inequality for U statistics~\cite[Theorem 2]{arcones1995bernstein} can be used to obtain
\begin{align*}
    \P\Big( \Big|\ts \frac{ \sum_{1 \leq k < j \leq N}  (x \ttop Q_k x - x \ttop Q_j x)^2}{N(N-1) x\ttop L x(1-x\ttop L x)} -1 \Big | \geq \epsilon\Big) \ \leq \ 4 \exp\Big( -  \frac{ N \epsilon^2 (x\ttop L x(1-x\ttop L x))^2}{8 + 128 x\ttop L x(1-x\ttop L x)\epsilon}\Big).
\end{align*}
Hence, for any $i = 1,\ldots, |\mathcal{C}|$ and $\epsilon \in (0,1)$, we have 
\begin{align*}
    \P\Big( | \ts \frac{ \sqrt{s_{ii}}}{\sqrt{\hat s_{ii}}} -1| \geq \epsilon\Big) \ \leq \ &\P\Big( | \ts \frac{ s_{ii}}{\hat s_{ii}} -1| \geq \epsilon\Big) \\
     \ \leq \ &\P\Big(| \ts \frac{\hat s_{ii}}{ s_{ii}} -1| \geq \frac{\epsilon}{2}\Big)\\
     \ \leq \ &\P\Big( \Big|\ts \frac{ \sum_{1 \leq k < j \leq N}  (x_i \ttop Q_k x_i - x_i \ttop Q_j x_i)^2}{N(N-1) x_i\ttop L x_i(1-x_i\ttop L x_i)} -1 \Big | \geq \frac{\epsilon}{4}\Big)\\
     \ \leq \ & 4 \exp\Big( \ts -  \frac{ N \epsilon^2   (x_i\ttop L x_i(1-x_i\ttop L x_i))^2}{128 (1+ 4\epsilon)}\Big),
\end{align*}
and a union bound implies
\begin{align}\label{eq:sl}
    \P\big( \max_{ 1 \leq i \leq |\mathcal{C}| } | \ts \frac{ \sqrt{s_{ii}}}{\sqrt{\hat s_{ii}}} -1| \geq \epsilon\big) \ \leq \ 4|\mathcal{C}| \exp\big( -  \frac{ N   \eta(\mathcal{C})^2\epsilon^2}{128 (1+ 4\epsilon)}\big),
\end{align}
when $N$ is large. Taking $\epsilon=\frac{\log(N|\mathcal{C}|)^{3/2}}{\sqrt{N}\eta(\mathcal{C})}$, the conditions in Theorem~\ref{thm:cutquery} show that as $n\to\infty$
\begin{align*}
    \P\big( \max_{ 1 \leq i \leq |\mathcal{C}| } | \ts \frac{ \sqrt{s_{ii}}}{\sqrt{\hat s_{ii}}} -1| \geq \frac{\log(N|\mathcal{C}|)^{3/2}}{ 4\sqrt{N}\eta(\mathcal{C})} \big)
    \ = \ &o(1),
\end{align*}
which proves that the left side of~\eqref{eq:coupling} is $o(1)$, completing the proof.
\qed
~\\

\begin{lemma}\label{lem:M*G}
    If the conditions in Theorem~\ref{thm:cutquery} hold, then as $n\to\infty$,
    \begin{align*}
        d_{\mathrm{K}}\big(\mathcal{L}( M^*|Q ), \mathcal{L}(M( G))\big) \ \xrightarrow{\P} \ 0.
    \end{align*}
\end{lemma}
\proof
By the triangle inequality, we have
\begin{equation}\label{eq:triangleineq}
    \begin{split}
        d_{\mathrm{K}}\big(\mathcal{L}( M^*|Q ), \mathcal{L}(M( G))\big) 
    \ \leq \
    & d_{\mathrm{K}}\big(\mathcal{L}(M^*|Q ), \mathcal{L}(M(\hat G)|Q)\big)  \ + \  d_{\mathrm{K}}\big(\mathcal{L}( M( G)), \mathcal{L}(M(\hat G)|Q)\big).
    \end{split}
\end{equation}
With regard to the second term on the right side, Lemma \ref{lem:harR-R} and the Gaussian comparison inequality in Lemma \ref{lem:gaussianappro} imply
\begin{align*}
    d_{\mathrm{K}}\Big(\mathcal{L} (M( G)  ), \mathcal{L}(M(\hat G)|Q)\big) \ \lesssim \ &\big( \|\hat R-R\|_{\infty} \log(|\mathcal{C}|)^2\Big)^{1/2}\\
    \ = \ &o_{\P}(1).
\end{align*}

To handle the first term on the right side of~\eqref{eq:triangleineq}, we will follow the argument used in deriving~\eqref{eq:D1}.
The calculation in \eqref{eq:4thcentralmoment} yields
\begin{align*}
    \E\big((x_i\ttop Q_1^* x_i - x_i  \ttop \hat L x_i)^4|Q\big) 
    \ = \ x_i\ttop \hat L x_i( 1 - x_i\ttop  \hat L x_i )( 1- 3x_i\ttop \hat L x_i( 1- x_i\ttop \hat L x_i) ).
\end{align*}
We will apply Lemma \ref{lem:CCK} (conditionally on $Q_1,\dots,Q_N$) to a set of random vectors $X_1,\dots,X_N\in\R^{|\mathcal{C}|}$, where the $j$th component of the $i$th vector is defined by $X_{ij}=(x_j\ttop Q_i^* x_j - x_j\ttop \hat L x_j)/\sqrt{\hat s_{jj}}$. Also, in the notation of that lemma, we will take $b_1 = b_2 =1$ and $c_n^2 =\frac{4}{\log(2)^2}/ \min_{x \in \mathcal{C}} \{x\ttop \hat L x - (x\ttop \hat L x)^2\}$, which implies that the following bound holds with probability 1,
\begin{align}\label{eqn:itermed}
    d_{\mathrm{K}}\Big(\mathcal{L}(M^*|Q ), \mathcal{L}(M(\hat G)|Q)\Big)
    \ \lesssim \  \Big( \frac{ \log(N|\mathcal{C}|)^5}{N\min_{x \in \mathcal{C}} x\ttop \hat L x (1-x\ttop \hat L x)}\Big)^{1/4}.
\end{align}
 Also, for any numbers $a,b\in [0,1]$, we have $|a(1-a)-b(1-b)|\leq 2|a-b|$ and so 
\begin{align*}
    \min_{x \in \mathcal{C}} x\ttop \hat L x (1-x\ttop \hat L x) \ \geq \   \eta(\mathcal{C})  - 2\max_{x \in \mathcal{C}} |x\ttop \hat L x-x\ttop L x |.
\end{align*}
To demonstrate the right side of~\eqref{eqn:itermed} is $o_{\P}(1)$, it suffices to show
\begin{equation*}
\max_{x \in \mathcal{C}} |x\ttop \hat L x-x\ttop L x | \ = \ o_{\P}( \eta(\mathcal{C})),
\end{equation*}
and then combining with the conditions in Theorem \ref{thm:cutquery} will give the right side of~\eqref{eqn:itermed} is $ o_{\P}(1)$.
The bound \eqref{eq:OP} implies $M = \mathcal{O}_{\P}(\sqrt{\log(N|\mathcal{C}|)})$, and so the conditions in Theorem \ref{thm:cutquery} give
 \begin{equation}\label{eq:maxL}
     \begin{split}
         \max_{x \in \mathcal{C}} |x\ttop L x - x\ttop \hat L x| 
        \  \leq \ & \frac{M}{\sqrt{\log(N|\mathcal{C}|)}}    \sqrt{\frac{\log(N|\mathcal{C}|)}{N}}\\
        \ = \ & \mathcal{O}_{\P}(1) \cdot o(  \eta(\mathcal{C}))\\
        \ = \ & o_{\P}\big( \eta(\mathcal{C})\big).
     \end{split}
 \end{equation}
%
\qed

\begin{lemma}
    \label{lem:harR-R}
    If the conditions in Theorem~\ref{thm:cutquery} hold, then as $n\to\infty$,
    \begin{align}\label{eqn:firststepcomp}
        \|\hat R-R\|_{\infty} \log(|\mathcal{C}|)^2 \ \xrightarrow{\P} \ 0.
    \end{align}
\end{lemma}
\proof
Observe that 
\begin{align}\label{eqn:firstcomparison}
    \|\hat R-R\|_{\infty} 
    \ \leq \ &\max_{ 1 \leq i, j \leq |\mathcal{C}| } \frac{|\hat s_{ij}  - s_{ij} |}{\sqrt{s_{ii} s_{jj}} } + \displaystyle \max_{ 1 \leq i, j \leq |\mathcal{C}| } \frac{|\hat s_{ij} |}{ \sqrt{ \hat s_{ii}\hat s_{jj}} } \Big|1 - \frac{\sqrt{ \hat s_{ii}\hat s_{jj}}}{ \sqrt{s_{ii} s_{jj}}}\Big|.
\end{align}
For the second term on the right side, by noting that $\frac{|\hat s_{ij} |}{ \sqrt{ \hat s_{ii}\hat s_{jj}} } \leq 1$, we can obtain
\begin{align*}
    \max_{ 1 \leq i, j \leq |\mathcal{C}| } \frac{|\hat s_{ij} |}{\sqrt{ \hat s_{ii}\hat s_{jj}} } \Big|1 -  \frac{\sqrt{ \hat s_{ii}\hat s_{jj}}}{ \sqrt{s_{ii} s_{jj}}}\Big|  \ \leq \ & \displaystyle \max_{ 1 \leq i\leq |\mathcal{C}| } \Big|1 -  \frac{ \sqrt{\hat s_{ii}}}{   \sqrt{s_{ii}}}\Big| + \displaystyle \max_{ 1 \leq i, j \leq |\mathcal{C}| }   \frac{ \sqrt{\hat s_{ii}}}{   \sqrt{s_{ii}}} \Big|1 -  \frac{\sqrt{\hat s_{jj}}}{ \sqrt{s_{jj}}}\Big| \\ 
    \ \lesssim \ &   \max_{ 1 \leq i\leq |\mathcal{C}| } \Big|1 -  \frac{ \sqrt{\hat s_{ii}}}{   \sqrt{s_{ii}}}\Big| + \Big( \max_{ 1 \leq i\leq |\mathcal{C}| } \Big|1 -  \frac{ \sqrt{\hat s_{ii}}}{   \sqrt{s_{ii}}}\Big|\Big)^2.
\end{align*}
Applying the inequality in \eqref{eq:sl} with $\epsilon= 1/\log(|\mathcal{C}|)^2$, we have
\begin{align*}
    \log(|\mathcal{C}|)^2 \max_{ 1 \leq i, j \leq |\mathcal{C}| } \frac{|\hat s_{ij} |}{ \sqrt{ \hat s_{ii}\hat s_{jj}} } \Big|1 -  \frac{\sqrt{ \hat s_{ii}\hat s_{jj}}}{ \sqrt{s_{ii} s_{jj}}}\Big| \ = \ o_{\P}(1).
\end{align*}
For the first term on the right side of~\eqref{eqn:firstcomparison}, combining the definitions of $s_{ij}$ and $\hat s_{ij}$ in \eqref{eq:defofsij} with the facts that $s_{ii}\geq \sqrt{\eta(\mathcal{C}) }$, $x_i\ttop \hat L x_i \leq 1$, and $x_i\ttop L x_i \leq 1$, we have
\begin{equation}\label{eqn:severalterms}
\small
\begin{split}
    \log(|\mathcal{C}|)^2 \max_{ 1 \leq i, j \leq |\mathcal{C}| } \frac{|\hat s_{ij}  - s_{ij} |}{\sqrt{s_{ii} s_{jj}} }
    \ \lesssim \ &  \frac{\log(|\mathcal{C}|)^2}{\eta(\mathcal{C})} \displaystyle \max_{ 1 \leq i, j \leq |\mathcal{C}| } \bigg| \frac{1}{N} \sum_{k=1}^N x_i\ttop  Q_{k} x_i x_j\ttop Q_{k} x_j   - \E(x_i\ttop  Q_{1} x_i x_j\ttop Q_{1} x_j) \bigg| \\
    &+  \frac{\log(|\mathcal{C}|)^2}{\sqrt{\eta(\mathcal{C}) }}\displaystyle \max_{1 \leq i \leq |\mathcal{C}|}  \frac{| x_i\ttop \hat L x_i  - x_i\ttop L x_i |}{\sqrt{s_{ii}}}.
    \end{split}
\end{equation}
Following a similar argument to~\eqref{eq:maxL}, the second term on the right side of~\eqref{eqn:severalterms} is $o_{\P}(1)$.  
It remains to show the first term on the right side of~\eqref{eqn:severalterms} is $o_{\P}(1)$.
By noting that $x_i\ttop  Q_{1} x_i x_j\ttop Q_{1} x_j$ takes values in $\{0,1\}$, Bernstein’s inequality \citep[Proposition 2.14]{wainwright2019high} gives for any fixed $\epsilon>0$,
\begin{align*}
    \P\bigg(\ts \frac{\log(|\mathcal{C}|)^2}{\eta(\mathcal{C})} \bigg|\displaystyle \frac{1}{N} \sum_{k=1}^N x_i\ttop  Q_{k} x_i x_j\ttop Q_{k} x_j   - \E(x_i\ttop  Q_{1} x_i x_j\ttop Q_{1} x_j) \bigg| >\epsilon\bigg) \ \leq \ &2 \exp\Big(- \ts \frac{N \eta(\mathcal{C})^2 \epsilon ^2}{2\log(|\mathcal{C}|)^4 ( 1 + \epsilon)}\Big),
\end{align*}
and so applying a union bound over $1\leq i,j\leq |\mathcal{C}|$ shows that the first term on the right side of~\eqref{eqn:severalterms} is indeed $o_{\P}(1)$.
Combining the above results with the conditions in Theorem~\ref{thm:cutquery} completes the proof.
\qed

\section{Proof of Theorem \ref{thm:Frobenius}}
\label{app:Frobenius}
\setcounter{lemma}{0}
\renewcommand{\thelemma}{F.\arabic{lemma}}

Let $\mu=\E(\|\hat L-L\|_F^2)$ and $\sigma^2 = \var(\|\hat L-L\|_F^2)$. Define the statistics
\begin{align*}
    T   \ = \   \frac{\|\hat L-L\|_F^2 - \hat \mu}{\hat \sigma}\ \ \ \ \ \ \ \ \ \ \ \ \text{ and } \ \ \ \ \ \ \ \ \ \ \ \ 
    T^*  \ = \   \frac{\|\hat L^*-\hat L\|_F^2 - \hat \mu^*}{\hat \sigma^*},
\end{align*}
where $\hat \mu$, $\hat \mu^*$, $\hat \sigma$, $\hat \sigma^*$ are defined as in Algorithm 1 with $\psi$ corresponding to $\|\cdot\|_F^2$. In particular, letting $\hat L_1^*, \ldots, \hat L_B^*$ denote conditionally i.i.d.~copies of $\hat L^*$ given $Q_1,\dots,Q_N$, the quantities $\hat\mu$ and $\hat\sigma$ can be represented in distribution as
\begin{align*}
    \hat\mu \ = \ \frac{1}{B} \sum_{b=1}^B \|\hat L^*_{b} -\hat L\|_F^2 \ \ \ \ \ \ \ \ \ \ \ \ \text{ and } \ \ \ \ \ \ \ \ \ \ \ \ 
    \hat \sigma^2  \ = \  \frac{1}{B} \sum_{b=1}^B \big(\|\hat L^*_b -\hat L\|_F^2 - \hat\mu\big)^2.
\end{align*}
As in the proof of Theorem \ref{thm:cutquery}, it is sufficient to show that as $n\to\infty$,
\begin{align}
    d_{\mathrm{K}}\big(\mathcal{L}(T)\, ,\, 
    \mathcal{L}(Z)\big)
    & \ \xrightarrow{} \ 0\label{eqn:firstTZlim}\\
    d_{\mathrm{K}}\big(\mathcal{L}(T^*  |Q)\, ,\,
    \mathcal{L}(Z)\big)
    & \ \xrightarrow{\P} \ 0,
    \label{eq:normalapprox1}
\end{align}
where $Z$ denotes a standard Gaussian random variable.
Lemma \ref{lem:correctedmeanvariance} ensures that $T \xrightarrow{\mathcal{L}} Z$, and consequently,
 P\'olya's theorem \citep[Theorem B.7.7]{bickel2015mathematical} implies the limit \eqref{eqn:firstTZlim}.

To establish $d_{\mathrm{K}}\big(\mathcal{L}(T^*  |Q)\, ,\,  \mathcal{L}(Z)\big) \xrightarrow{\P}0$, we need to deal with the fact that $\mathcal{L}(T^*|Q)$ is a random probability distribution. It is enough to show that for any subsequence $J\subset\{1,2,\dots\}$, there is a further subsequence $J'\subset J$ such that the limit $d_{\mathrm{K}}\big(\mathcal{L}(T^*  |Q)\, ,\,  \mathcal{L}(Z)\big) \xrightarrow{}0$ holds almost surely as $n\to\infty$ along $n\in J'$. The key ingredient for doing this is to show that if $\hat{\mathsf{d}}\in\R^n$ contains the diagonal entries of $\hat L$, then $\|\hat{\mathsf{d}}\|_{\infty}/\|\hat{\mathsf{d}}\|_2 =o_{\P}(1)$ holds as $n\to\infty$, which is established in Lemma \ref{lem:conditionofS}. This implies that $\|\hat{\mathsf{d}}\|_{\infty}/\|\hat{\mathsf{d}}\|_2 \to 0$ holds almost surely as $n\to\infty$ along a subsequence of $J$. Because $\hat L^*$ can be viewed as being generated with $N$ edges that are drawn from $\hat G$ in an i.i.d.~manner with edge-weight sampling, analogues of the original conditions in Theorem~\ref{thm:Frobenius} hold with respect to $\hat L$ (instead of $L$), almost surely along subsequences. Therefore, the argument for proving~\eqref{eqn:firstTZlim} can be used in a completely analogous manner to prove the limit~\eqref{eq:normalapprox1}.\label{explanation}  \\

The following lemma provides some basic properties of $L$ that we need at various points in the proof of Theorem \ref{thm:Frobenius}.
\begin{lemma}
    \label{lem:rankofL}
    If $\frac{n}{N} \to 0$ and $\frac{\|\mathsf{d}\|_{\infty}^2}{\|\mathsf{d}\|_2^2} \to 0$ hold as $n\to\infty$, then the following limits also hold as $n\to\infty$,
    \begin{equation*}
        \|L\|_{\infty} \ = \ \|\mathsf{d}\|_{\infty} \ = \ o(1), \ \ \ \ \ \ \ \ \ \ \  \tr(L^2)  \ = \  o(1), \ \ \ \ \ \ \ \ \ \ \  \text{ and }  \ \ \ \ \ \ \ \ \ \ \   N\tr(L^2) \ \xrightarrow{} \ \infty.
    \end{equation*}
\end{lemma}
\proof

Note that the entry of a positive semidefinite matrix with the largest magnitude must always occur along the diagonal, and so $\|L\|_{\infty} \ = \ \|\mathsf{d}\|_{\infty}$.

To show $\|\mathsf{d}\|_{\infty}=o(1)$, we write  $\|\mathsf{d}\|_{\infty}= (\|\mathsf{d}\|_{\infty}/\|\mathsf{d}\|_2)\|\mathsf{d}\|_2$, and so the assumption $\|\mathsf{d}\|_{\infty}/\|\mathsf{d}\|_2=o(1)$ implies that it is sufficient to show $\|\mathsf{d}\|_2\lesssim 1$. For this purpose, first note that $\|\mathsf{d}\|_2^2 \leq \tr(L^2)$. Since $\tr(L)=2w(E)$ our reduction to the case when $w(E)=1$ gives $\tr(L) = 2$. Therefore, using the general inequality $1\leq \tr(A)^2/\tr(A^2)$ for any non-zero $n\times n$ positive semidefinite matrix $A$, we have $  \tr(L^2) \lesssim 1$, as needed.

To show $\tr(L^2)=o(1)$, observe that H\"older's inequality and our previous steps imply
\begin{equation*}
    \begin{split}
        \tr(L^2) & \ \leq \ \|L\|_{\infty} \sum_{1 \leq i ,j\leq n} |L_{i j}|\\
        & \ = \|L\|_{\infty} \cdot 4\sum_{e\in E} w(e)\\
        & \ = \ \|\mathsf{d}\|_{\infty} \cdot 4\\
        & \ = \ o(1).
    \end{split}
\end{equation*}

Finally, to show that $N\tr(L^2)\to\infty$ as $n\to\infty$, note that the inequality $\tr(A)^2/\tr(A^2)\leq n$ holds for any non-zero $n\times n$ positive semidefinite matrix $A$, and so $\tr(L^2)\geq \tr(L)^2/n=4/n$. So, because our assumption on $N$ and $n$ implies $N/n\to\infty$, the proof is complete.\qed

\subsection{Asymptotic normality of $\|\hat L-L\|_F^2$}

\begin{lemma}
    \label{lem:correctedmeanvariance}
    If the conditions in Theorem \ref{thm:Frobenius} hold, then as $n\to\infty$,
    \begin{equation*}
    \frac{\|\hat L-L\|_F^2 -\hat\mu}{\hat\sigma}
        \ \xrightarrow{\mathcal{L}} \ N(0,1).
    \end{equation*}
\end{lemma}
\proof
Lemmas \ref{lem:consistencyofexpectation} and \ref{lem:consistencyofvar} establish 
\begin{align*}
    \frac{\hat\mu-\mu}{\sigma} \ \xrightarrow{\P} \ 0  \ \ \ \ \ \ \ \ \ \ \ \text{ and } \ \ \ \ \ \ \ \ \ \ \ \frac{\hat \sigma^2}{\sigma^2} \ \xrightarrow{\P} \ 1.
\end{align*}
If we can show
\begin{align}
\label{eq:clt}
    \frac{\|\hat L-L\|_F^2 -   \mu }{\sigma}  \ \xrightarrow{\mathcal{L}} \ N(0,1),
\end{align}
then the proof is completed by Slutsky's lemma.
Recall $D_1,\dots,D_N$ defined in Appendix \ref{sec:notation} are independent and identically distributed random vectors with $\E(D_1 D_1\ttop)=L$, and so we have
\begin{equation}\label{eq:expectationofF2}
    \begin{split}
\mu & \ = \ \frac{1}{N^2}\sum_{i,j=1}^N \E \Big(\big\llangle D_iD_i\ttop -L \, , \, D_jD_j\ttop -L\big\rrangle\Big)\\
& \ = \  \frac{1}{N}\E\Big(\|D_1D_1\ttop -L\|_F^2\Big) \\
& \ = \  \frac{1}{N}(4-\tr(L^2)),
\end{split}
\end{equation}
where we have used the almost-sure relation $D_1\ttop D_1=2$ in the last step.  Based on this formula for $\mu$, it can be checked by a direct algebraic calculation that $\|\hat L-L\|_F^2 - \mu$  may be decomposed according to
\begin{equation*}
    \|\hat L-L\|_F^2 - \mu \  = \  \frac{N-1}{N} U \ - \ \frac{2}{N} U',
\end{equation*}
where we define the statistics
\begin{align*}
    U \ = \ &\Big(\ts \frac{1}{N(N-1)}\sum_{1 \leq i\neq j \leq N}\big(D_{i}\ttop D_{j}\big)^2\Big) - \Big(\ts \frac{2}{N}\sum_{i=1}^N  D_{i}\ttop LD_{i}\Big) + \tr(L^2) \label{eqn:Udef}\\
    U' \ = \ & \Big(\ts \frac{1}{N}\sum_{i=1}^N D_{i}\ttop LD_{i}\Big) - \tr(L^2).
\end{align*}
Lemma \ref{lem:chiapproximation} shows that $U/\sigma$ converges in distribution to $N(0,1)$ as $n\to\infty$, and so the desired limit \eqref{eq:clt} will hold if we show that $U'/(N\sigma)$ is $o_{\P}(1)$. It is simple to check $\E(D_{1}\ttop LD_{1}) =\tr(L^2)$, and so $\E(U') = 0$. Hence, it is enough to show that the variance of $U'/(N\sigma)$ is $o(1)$. Since Lemma \ref{lem:sigma2} gives $N^2 \sigma^2 \asymp \tr(L^2)$ and equation \eqref{eq:orderofD} in the proof of Lemma \ref{lem:sigma2} implies $\var(D_{1}\ttop LD_{1}) = o(\tr(L^2))$, we have 
\begin{align*}
    \var\Big(\frac{U'}{N\sigma} \Big) \ = \ \frac{1}{N^2 \sigma^2} \cdot \frac{1}{N}\var(D_{1}\ttop LD_{1})  \ = \ o(1).
\end{align*} 

\qed

\begin{lemma}
\label{lem:chiapproximation}
    If the conditions in Theorem \ref{thm:Frobenius} hold, then as $n\to\infty$,
        \begin{align*}
            d_{\mathrm{K}}\Big(\mathcal{L}(\ts \frac{ U }{ \sigma}), \mathcal{L}(Z )\Big) \ \to \ 0, 
        \end{align*}
        where $Z$ is a standard Gaussian random variable.
\end{lemma}
\proof
Observe that 
\begin{align*}
    U \ = \ \frac{1}{N(N-1)} \sum_{1 \leq i\neq j \leq N} h(\hat e_i, \hat e_j),
\end{align*}
where
\begin{equation*}
    \begin{split}
         h(\hat e_1, \hat e_2) \ = \ & \llangle D_{1} D_{1}\ttop -L ,  D_{2} D_{2}\ttop -L \rrangle\\
          \ = \ &  \sum_{1 \leq i \leq n} (D_{1} D_{1}\ttop -L)_{ii}(D_{2} D_{2}\ttop -L)_{ii} + 2 \sum_{1 \leq i < j \leq n} (D_{1} D_{1}\ttop -L)_{ij}(D_{2} D_{2}\ttop -L)_{ij}.
    \end{split}
\end{equation*}
Define the collection of ordered pairs $\mathcal{J} = \{(i,i')| 1\leq i\leq i' \leq n\}$. For each $\mathbf{i} \in \mathcal{J}$, define
\begin{equation} \label{eq:definitionofvarphi}
    \begin{split}
        &\varphi_{\mathbf{i} }(\hat e_1)  \ = \  1{\{i=i'  \in \hat e_1\}} - \sqrt{2} \cdot 1{\{i< i', \hat e_1 = \{i,i'\} \} }\\
        &\phi_{\mathbf{i} }(\hat e_1) \ = \ \varphi_{\mathbf{i}}(\hat e_1)-\E(\varphi_{\mathbf{i}}(\hat e_1)).
    \end{split}
\end{equation}
It can be checked that 
\begin{equation}\label{eq:expectaionofvarphi}
    \begin{split}
        \E(\varphi_{\mathbf{i}}(\hat e_1))  & \ = \  L_{i i'} \big(1{\{i= i'\}} + \sqrt{2}  \cdot 1{\{i< i'\} } \big)\\
        \phi_{\mathbf{i} }(\hat e_1) & \ = \ (D_{1} D_{1}\ttop -L)_{ii'} \big(1{\{i= i'\}} + \sqrt{2}  \cdot 1{\{i< i'\} } \big),
    \end{split}
\end{equation}
which leads to 
\begin{align*}
    h(\hat e_1, \hat e_2) \ = \ \sum_{ \mathbf{i} \in \mathcal{J} } \phi_{\mathbf{i}}(\hat e_1)\phi_{\mathbf{i}}(\hat e_2).
\end{align*}
Let $\varphi(\hat e_1)$ and $\phi(\hat e_1)$ respectively denote the random vectors in $\mathbb{R}^{n(n+1)/2}$ defined by
\begin{equation*}
    \begin{split}
        \varphi(\hat e_1)& \ = \ (\varphi_{\mathbf{i}}(\hat e_1))_{\mathbf{i}\in\mathcal{J}}\\
        \phi(\hat e_1)& \ = \ (\phi_{\mathbf{i}}(\hat e_1))_{\mathbf{i}\in\mathcal{J}}.
    \end{split}
\end{equation*}
Also, let $\mathfrak{S}$ denote the $\frac{1}{2}n(n+1)\times \frac{1}{2}n(n+1)$ covariance matrix of the random vector $\phi(\hat e_1)$, so that
\begin{equation}\label{eq:definitionofmathfrakS}
\begin{split}
    \mathfrak{S}& \ = \ \E(\phi(\hat e_1)\phi(\hat e_1)\ttop)\\
& \ = \ \E(\varphi(\hat e_1)\varphi(\hat e_1)\ttop)-\E(\varphi(\hat e_1))\E(\varphi(\hat e_1)\ttop).
    \end{split}
\end{equation}
Likewise, let $\lambda(\mathfrak{S})\in\mathbb{R}^{n(n+1)/2}$ denote the vector containing the eigenvalues of $\mathfrak{S}$, and define the random variable 
\begin{equation*}
    \xi_n \ = \ \frac{\sum_{j=1}^{n(n+1)/2}\lambda_j(\mathfrak{S})(Z_j^2-1)}{\sqrt{2}\var\big(h(\hat e_1, \hat e_2)\big)^{1/2} },
\end{equation*} 
where $Z_1,\dots,Z_{\frac{1}{2}n(n+1)}$ are independent standard Gaussian random variables.
It can be checked that $\E\big(h(\hat e_1, \hat e_2)|
\hat e_1\big)=0$, and applying Lemma \ref{lem:ustatistic} \citep[Proposition 9]{huang2023high} yields
\begin{align}\label{eq:weightedchi}
        d_{\mathrm{K}}\Big(\mathcal{L} \big(\ts \frac{ \sqrt{N(N-1)}}{ \sqrt{2\var(h(\hat e_1, \hat e_2))}}U\big), \mathcal{L}\big(\xi_n\big)\Big)  \ \lesssim \  N^{-1/5} \ + \  \Big(\frac{\E(|h(\hat e_1, \hat e_2)|^{3})}{\sqrt{N}\var(h(\hat e_1, \hat e_2))^{3/2}}\Big)^{1/7}.
\end{align}
Following the calculation in the proof of Lemma \ref{lem:sigma2}, we obtain
\begin{equation}
\label{eq:sigmau}
    \begin{split}
        \var\big(h(\hat e_1, \hat e_2)\big)
        \ = \ & \tr(L^2)  +  12\sum_{1 \leq i<j \leq n} L_{i j} ^2  +  o(\tr(L^2))\\
        \ \asymp \ & \tr(L^2),
    \end{split}
\end{equation}
and
\begin{equation}\label{eq:varuvar}
    \frac{2}{N(N-1)\sigma^2}\var\big(h(\hat e_1, \hat e_2)\big) \ \to \ 1.
\end{equation}
To complete the proof, it remains to establish an upper bound on $\E\big(| h(\hat e_1, \hat e_2)|^3\big)$. As a shorthand, we write $\hat e_1 \sim \hat e_2$ whenever the edges $\hat e_1$ and $\hat e_2$ share exactly one vertex,~$|\hat e_1\cap \hat e_2| = 1$. By noting the basic relation 
\begin{align*}
    |D_1\ttop D_2| \ = \ 1\{\hat e_1\sim \hat e_2\}+2\cdot 1\{\hat e_1=\hat e_2\},
\end{align*}
we have
\begin{align*}
    h(\hat e_1, \hat e_2) \ = \  1{\{\hat e_1\sim \hat e_2\}} +4 \cdot 1{\{\hat e_1=\hat e_2\}}- D_{1} \ttop L D_{1} -   D_{2} \ttop L D_{2} + \tr(L^2).
\end{align*}
Due to $D_1\ttop L D_1\leq 4\|L\|_{\infty}=o(1)$ (by Lemma~\ref{lem:rankofL}), we have $\E((D_1\ttop L D_1)^3)\lesssim \E(D_1\ttop L D_1)$ and
\begin{align*}
    \E\big(|h(\hat e_1, \hat e_2) | ^3\big) \ \lesssim \  &\E (  1{\{\hat e_1 \sim \hat e_2 \}}  ) + \E (  1{\{\hat e_1 = \hat e_2\}}  ) + \E\big((D_1\ttop L D_1)^3\big) +\tr(L^2)^3 \\
    \ \lesssim \ &  \sum_{1 \leq i<j \leq n} |L_{i j}| (L_{i i}+L_{jj}) \ + \  \sum_{1 \leq i<j \leq n}\!\! L_{i j}^2 \ + \  \E\big( D_{1} \ttop L D_{1}\big) +\tr(L^2)^3 \\
    \ \lesssim \  &  \tr(L^2),
\end{align*}
where the last step is based on Lemma \ref{lem:rankofL} and the fact that $L$ is diagonally dominant. Applying the above results and Lemma \ref{lem:rankofL} to \eqref{eq:weightedchi} implies
\begin{align*}
    d_{\mathrm{K}}\Big(\mathcal{L} \big(\ts \frac{ \sqrt{N(N-1)}}{ \sqrt{2\var(h(\hat e_1, \hat e_2))}}U\big), \mathcal{L}\big(\xi_n\big)\Big) \ \lesssim \  &  N^{-1/5} \ + \ \big(N\tr(L^2)\big)^{-1/14}\\
     \ \xrightarrow{} \  & 0.
\end{align*}
Also, Lemma \ref{lem:normalapproximation} and \eqref{eq:sigmau} imply $\xi_n \xrightarrow{\mathcal{L}} N(0,1)$.
Combining the above results with \eqref{eq:varuvar}, Slutsky's lemma completes the poof.
\qed

\begin{lemma}
    \label{lem:normalapproximation}
    Let $\{Z_j\,|\,1\leq j\leq n(n+1)/2\}$ be a collection of i.i.d.~$N(0,1)$ random variables. If the conditions in Theorem \ref{thm:Frobenius} hold, then as $n\to\infty$,
    \begin{align*}
        \frac{\sum_{j=1}^{n(n+1)/2}\lambda_j(\mathfrak{S})(Z_j^2-1) }{\sqrt {2(\tr(L^2)  +  12\sum_{1 \leq i<j \leq n} L_{i j} ^2}) } \ \xrightarrow{\mathcal{L}} \ N(0,1).
    \end{align*}
\end{lemma}
\proof
It follows from the Lindeberg CLT for triangular arrays given in \cite[Prop.~2.27]{van2000asymptotic} that if the condition $\frac{\|\lambda(\mathfrak{S})\|_{\infty}^2}{\|\lambda(\mathfrak{S})\|_2^2} \to 0$ holds, then as $n\to\infty$,
\begin{align*}
    \frac{1}{\sqrt{2}\|\lambda(\mathfrak{S})\|_2} \sum_{j=1}^{n(n+1)/2}\lambda_j(\mathfrak{S})(Z_j^2-1) \ \xrightarrow{\mathcal{L}} \ N(0,1).
\end{align*}
It remains to show 
\begin{equation}\label{eq:limitoftau}
    \frac{\|\lambda(\mathfrak{S})\|_2^2 }{\tr(L^2)  +  12\sum_{1 \leq i<j \leq n} L_{i j} ^2} \ \to \ 1 \ \ \ \ \ \ \ \ \ \ \text{ and } \ \ \ \ \ \ \ \ \ \ \frac{\|\lambda(\mathfrak{S})\|_{\infty}^2}{\|\lambda(\mathfrak{S})\|_2^2} \ \to \ 0.
\end{equation}
By noting that $\|\lambda(\mathfrak{S})\|_2^2 = \|\mathfrak{S}\|_F^2$, we need to calculate the sum of squares of the following entries
\begin{align*}
    \mathfrak{S}_{\mathbf{i}\mathbf{j}} \ = \ \E(\varphi_{\mathbf{i}}(\hat e_1) \varphi_{\mathbf{j}}(\hat e_1)) - \E(\varphi_{\mathbf{i}}(\hat e_1))\E(\varphi_{\mathbf{j}}(\hat e_1))
\end{align*}
for $\mathbf{i}, \mathbf{j} \in \mathcal{J}$, where $\mathcal{J} = \{(i,i')| 1\leq i\leq i' \leq n\}$.  
The equalities in \eqref{eq:expectaionofvarphi} give $\sum_{\mathbf{i} \in \mathcal{J}} (\E(\varphi_{\mathbf{i}}(\hat e_1)))^2 = \tr(L^2)$ and so
\begin{align*}
    \|\mathfrak{S}\|_F^2  
    \ = \ & \sum_{\mathbf{i}, \mathbf{j} \in \mathcal{J}} (\E(\varphi_{\mathbf{i}}(\hat e_1) \varphi_{\mathbf{j}}(\hat e_1)))^2 - 2\sum_{\mathbf{i}, \mathbf{j} \in \mathcal{J}} \E(\varphi_{\mathbf{i}}(\hat e_1))\E(\varphi_{\mathbf{j}}(\hat e_1)) \E(\varphi_{\mathbf{i}}(\hat e_1) \varphi_{\mathbf{j}}(\hat e_1)) + \tr(L^2)^2.
\end{align*}
Combining with Lemma \ref{lem:expectationofvarphi} yields 
\begin{align*}
    \sum_{\mathbf{i}, \mathbf{j} \in \mathcal{J}} \E(\varphi_{\mathbf{i}}(\hat e_1) \varphi_{\mathbf{j}}(\hat e_1))^2   \ = \ & \sum_{i=1}^{n} L_{i i}^2 +   \sum_{1 \leq i \neq j \leq n} L_{i j}^2 + 4  \sum_{1 \leq i < j \leq n} L_{i j}^2 + 4\cdot 2 \sum_{1 \leq i < j \leq n} L_{i j}^2  \\
     \  = \ &\tr(L^2)  +  12\sum_{1 \leq i<j \leq n} L_{i j} ^2,
\end{align*}
and
\begin{align*}
    \sum_{\mathbf{i}, \mathbf{j} \in \mathcal{J}} \big|\E(\varphi_{\mathbf{i}}(\hat e_1))\E(\varphi_{\mathbf{j}}(\hat e_1)) \E(\varphi_{\mathbf{i}}(\hat e_1) &\varphi_{\mathbf{j}}(\hat e_1))\big|\\
\lesssim \  & \  \sum_{i=1}^{n} L_{i i}^3 + \sum_{1 \leq i\neq j \leq n}|L_{i j}|L_{i i} L_{jj}+  \sum_{1 \leq i\neq j \leq n} |L_{i j}|^3 +  \sum_{1 \leq i\neq j \leq n} L_{i j}^2L_{i i}\\
     \  = \ & \ o(\tr(L^2)  ),
\end{align*}
where the last step is based on $L_{i i} = \sum_{j\neq i} |L_{i j}|$ and $\|L\|_{\infty} = o(1)$ given in Lemma \ref{lem:rankofL}.
By noting that Lemma \ref{lem:rankofL} shows $\tr(L^2)^2 = o(\tr(L^2))$, we obtain
\begin{equation*}
    \begin{split}
       \|\mathfrak{S}\|_F^2  \  = \ &\tr(L^2)  +  12\sum_{1 \leq i<j \leq n} L_{i j} ^2+ o( \tr(L^2)),
    \end{split}
\end{equation*}
which leads to the first limit in \eqref{eq:limitoftau}.

To complete the proof, we must show that $\|\lambda(\mathfrak{S})\|_{\infty}^2 = o(\|\lambda(\mathfrak{S})\|_{2}^2)$. Note that the previous calculation gives $\|\lambda(\mathfrak{S})\|_2^2 =\|\mathfrak{S}\|_F^2\asymp \tr(L^2)\geq \|\mathsf{d}\|_2^2$, and so it is sufficient to demonstrate  $\|\lambda(\mathfrak{S})\|_{\infty} \lesssim \|\mathsf{d}\|_{\infty}$, since $\|\mathsf{d}\|_{\infty}=o(\|\mathsf{d}\|_2)$ holds under the assumptions of Theorem~\ref{thm:Frobenius}. The definition of $\mathfrak{S}$ in \eqref{eq:definitionofmathfrakS} implies that $\E(\varphi(\hat e_1)\varphi(\hat e_1)\ttop)-\mathfrak{S}$ is positive semidefinite, and so
$\lambda_{\max}(\mathfrak{S}) \leq \lambda_{\max}(\E(\varphi(\hat e_1)\varphi(\hat e_1)\ttop))$. Since the inequality $\lambda_{\max}(A) \leq \max_{1 \leq i \leq d} \sum_{j=1}^d |A_{ij}|$ holds for any symmetric matrix $A \in \mathbb{R}^{d\times d}$, we have 
\begin{align*}
    \lambda_{\max}(\E(\varphi(\hat e_1)\varphi(\hat e_1)\ttop)) \ \leq \ \max_{\mathbf{i} \in \mathcal{J}}   \sum_{\mathbf{j} \in \mathcal{J}} \big|\E(\varphi_{\mathbf{i}}(\hat e_1)  \varphi_{\mathbf{j}}(\hat e_1))\big|.
\end{align*}
Letting $\mathbf{i} = (i,i') \in \mathcal{J}$, Lemma \ref{lem:expectationofvarphi} gives for $i = i'$, 
\begin{align*}
    \sum_{\mathbf{j} \in \mathcal{J}} \big|\E(\varphi_{\mathbf{i}}(\hat e_1)  \varphi_{\mathbf{j}}(\hat e_1))\big| \ = \ &L_{i i} + \sum_{j\neq i} |L_{i j}| + \sqrt{2} \sum_{j> i} |L_{i j}| + \sqrt{2} \sum_{j< i} |L_{i j}|\\
    \ \lesssim \ & \|\mathsf{d}\|_{\infty},
\end{align*}
and for $i < i'$,
\begin{align*}
    \sum_{\mathbf{j} \in \mathcal{J}} \big|\E(\varphi_{\mathbf{i}}(\hat e_1)  \varphi_{\mathbf{j}}(\hat e_1))\big| \ = \ &2|L_{i i'}| + \sqrt{2} |L_{i i'}| + \sqrt{2} |L_{i i'}|\\
    \ \lesssim \ & \|\mathsf{d}\|_{\infty}.
\end{align*}
Consequently, we conclude $\|\lambda(\mathfrak{S})\|_{\infty} \lesssim \|\mathsf{d}\|_{\infty}$.
\qed

\begin{lemma}
\label{lem:expectationofvarphi}
    Let $\varphi_{\mathbf{i} }(\hat e_1)$ be as defined in \eqref{eq:definitionofvarphi}, and let $\mathbf{i}=(i,i'), \mathbf{j}=(j,j') \in \{(i,i')| 1\leq i\leq i' \leq n\}$. If the conditions in Theorem \ref{thm:Frobenius} hold, then 
    \begin{align*}
     \E(\varphi_{\mathbf{i}}(\hat e_1) \varphi_{\mathbf{j}}(\hat e_1)) \ = \
        \begin{cases}
            L_{i i}  & i=i'=j= j'\\
            -L_{i j'} & i=i'\neq j= j'\\
            -2L_{i j'} & i=j< i'=j' \\
            \sqrt{2} L_{i' j'}  & i=i'=j< j', j=j'=i< i'\\
            \sqrt{2} L_{i j}  & i=i'=j'> j, j=j'=i'> i\\
            \ 0 & 
            \text{ otherwise.} 
        \end{cases}
    \end{align*}
    
\end{lemma}
\proof
Observe that 
\begin{align*}
    \varphi_{\mathbf{i} }(\hat e_1)\varphi_{\mathbf{j} }(\hat e_1)  \ = \  &1{\{i=i'  \in \hat e_1, j=j'  \in \hat e_1\}}  \ + \  2 \cdot 1{\{i< i', \hat e_1 = \{i, i'\}, j< j', \hat e_1 = \{j, j'\} \} } \\
    &- \sqrt{2} \cdot 1{\{i=i' \in \hat e_1, j< j', \hat e_1 = \{j, j'\} \} } \ - \ \sqrt{2}\cdot 1{\{j=j' \in \hat e_1, i< i', \hat e_1 = \{i,i'\} \} },
\end{align*}
and for any $\mathbf{i}, \mathbf{j}$, only one term on the right side is nonzero at most. To make the first term nonzero, there are two possible cases: $i=i'=j= j'$ and $i=i'\neq j= j'$. When $i=i'=j= j'$, we have 
\begin{align*}
    \E(\varphi_{\mathbf{i} }(\hat e_1)\varphi_{\mathbf{j} }(\hat e_1)) \  = \  \E(1{\{i \in \hat e_1\}}) \  = \  L_{i i}.
\end{align*}
For $i=i'\neq j= j'$, we obtain 
\begin{align*}
    \E(\varphi_{\mathbf{i} }(\hat e_1)\varphi_{\mathbf{j} }(\hat e_1)) \  = \  \E(1{\{\hat e_1 =\{i,j'\} \}}) \  = \  -L_{i j'}.
\end{align*}
To make the second term nonzero, we need $i=j< i'=j'$, and the corresponding expectation is 
\begin{align*}
    \E(\varphi_{\mathbf{i} }(\hat e_1)\varphi_{\mathbf{j} }(\hat e_1)) \  = \  2 \E(1{\{\hat e_1 = \{i,j'\} \}}) \  = \  -2L_{i j'}.
\end{align*}
To make the third term nonzero, there are also two possible ways: $i=i'=j< j'$ and $i=i'=j'> j$. When $i=i'=j< j'$, the corresponding term can be calculated as 
\begin{align*}
    \E(\varphi_{\mathbf{i} }(\hat e_1)\varphi_{\mathbf{j} }(\hat e_1)) \  = \   - \sqrt{2} \E(1{\{\hat e_1 = \{i',j'\} \}}) \  = \  \sqrt{2}L_{i' j'}.
\end{align*}
When $i=i'=j'> j$, the corresponding term is
\begin{align*}
    \E(\varphi_{\mathbf{i} }(\hat e_1)\varphi_{\mathbf{j} }(\hat e_1)) \  = \  - \sqrt{2} \E(1{\{\hat e_1 = \{i,j\} \}}) \  = \  \sqrt{2}L_{i j}.
\end{align*}
The fourth term on the right side can be handled in the same way as the third term. For the other cases, all four terms are zero, and thus, the corresponding expectation is zero.
\qed

\subsubsection{Consistency of the mean estimate}

\begin{lemma}
\label{lem:consistencyofexpectation}
    If the conditions in Theorem \ref{thm:Frobenius} hold, then as $n\to\infty$,
    \begin{equation*}
        \frac{\hat\mu-\mu}{\sigma} \ \xrightarrow{\P} \ 0.
    \end{equation*}
\end{lemma}
\proof
First recall that Lemma \ref{lem:sigma2} gives $\sigma^2 \asymp \tr(L^2)/N^2$. In light of the bias-variance decomposition for the mean-squared error of $\hat\mu$, it is sufficient to show that $N^2(E(\hat\mu)-\mu)^2/\tr(L^2)$ and $N^2\var(\hat\mu)/\tr(L^2)$ are both $o(1)$.
With regard to the bias, observe that 
\begin{equation*}
    \hat\mu-\mu \ = \ \frac{1}{B}\sum_{b=1}^B \Big( \|\hat L^*_{b} -\hat L\|_F^2 - \E(\|\hat L-L\|_F^2)\Big).
\end{equation*}
Combining the definitions of $\hat L^*$ and $\hat L$ with the calculation in \eqref{eq:expectationofF2} gives
\begin{equation}\label{eq:expectaionofF2|S}
    \begin{split}
        \E(\|\hat L^* -\hat L\|_F^2 |Q) \ = \  \frac{4}{N} - \frac{1}{N}  \tr(\hat L^2).
    \end{split}
\end{equation}
To deal with $\tr(\hat L^2)$, the identity $\tr(\hat L^2)=-\tr(L^2)+2\llangle \hat L, L\rrangle +\|\hat L-L\|_F^2 $ 
and the calculation in~\eqref{eq:expectationofF2} can be used to show that
\begin{equation}\label{eq:expectationoftrL2}
    \E(\tr(\hat L^2)) \ = \ \frac{4}{N} +  \frac{N-1}{N}\tr(L^2)
\end{equation}
and so
\begin{equation*}
    \begin{split}
        \E(\|\hat L^* -\hat L\|_F^2) \ = \  \frac{N-1}{N^2}(4- \tr(L^2)).
    \end{split}
\end{equation*}
Consequently, we have the following formula for the bias of $\hat\mu$,
\begin{equation*} 
    \begin{split}
        |\E(\hat \mu)-\mu| \ = \ 
        & \Big|\frac{\tr(L^2)-4}{N^2}\Big| \\
    \ = \ &o\big(\ts \frac{\sqrt{\tr(L^2) }}{N}\big),
    \end{split}
\end{equation*}
where the second step is due to Lemma \ref{lem:rankofL}.
To analyze the variance of $\hat\mu$, equations \eqref{eq:sdmean} and \eqref{eq:meansd} in the proof of Lemma \ref{lem:varofF2} give    
\begin{equation}\label{eq:varofmeanofF2}
    \begin{split}
        \var(\hat\mu) \ = \ &\var\Big(\frac{1}{B}\sum_{b=1}^B  \|\hat L^*_{b} -\hat L\|_F^2 \Big)\\
        \ = \ &\E\Big(\ts\frac{1}{B} \var\big(\|\hat L^* -\hat L\|_F^2 |Q\big)\Big) +\var\Big(\E\big(\|\hat L^* -\hat L\|_F^2 |Q\big)\Big)\\
        \ = \ &o\big(\ts \frac{ \tr(L^2)}{N^2}\big),
    \end{split}
\end{equation}
which completes the proof.
\qed

\subsubsection{Consistency of the variance estimate}

\begin{lemma} 
\label{lem:consistencyofvar}
    If the conditions in Theorem \ref{thm:Frobenius} hold, then as $n\to\infty$,
    \begin{equation*}
        \frac{\hat\sigma^2}{\sigma^2} \ \xrightarrow{\P} \ 1.
    \end{equation*}
\end{lemma}
\proof 
Due to the fact that variance is shift invariant, note that the estimate $\hat\sigma^2$ can be written as
\begin{equation}
            \hat \sigma^2   \ = \  \frac{1}{B} \sum_{b=1}^B \big (\|\hat L^*_b -\hat L\|_F^2 -  \E(\|\hat L^* -\hat L\|_F^2)\big)^2 - \Big(\frac{1}{B} \sum_{b=1}^B \|\hat L^*_b -\hat L\|_F^2 -  \E(\|\hat L^* -\hat L\|_F^2)\Big)^2.
\end{equation}
Lemma~\ref{lem:sigma2} ensures that $\sigma^2\asymp \tr(L^2)/N^2$, and so it suffices to show that the bias and variance of $\hat\sigma^2$ satisfy $|\E(\hat\sigma^2)-\sigma^2|=o(\tr(L^2)/N^2)$ and $\var(\hat\sigma^2)=o(\tr(L^2)^2/N^4)$.

Towards calculating the bias of $\hat\sigma^2$, we first calculate its expectation.
Due to the fact that each $\|\hat L^*_b -\hat L\|_F^2 -  \E(\|\hat L^* -\hat L\|_F^2)$ is centered, we have
\begin{align*}
    \E(\hat \sigma^2) \ = \ &\var\big(\|\hat L^* -\hat L\|_F^2\big) - \var\Big(\frac{1}{B}\sum_{b=1}^B  \|\hat L^*_{b} -\hat L\|_F^2 \Big)\\
     \ = \ & \frac{2}{N^2}\big(\tr(L^2)  +  12\sum_{1 \leq i<j \leq n} L_{i j} ^2\big) + o\big(\ts \frac{ \tr(L^2)}{N^2}\big),
\end{align*}
where the second step follows from Lemma \ref{lem:varofF2} and \eqref{eq:varofmeanofF2}. Next, Lemma \ref{lem:sigma2} shows that 
\begin{align*}
    \sigma^2 \ = \ &\frac{2}{N^2}\big(\tr(L^2)  +  12\sum_{1 \leq i<j \leq n} L_{i j} ^2\big) + o\big(\ts \frac{ \tr(L^2)}{N^2}\big)
\end{align*}
and so the bias of $\hat\sigma^2$ satisfies
\begin{equation*}
    |\E(\hat\sigma^2)-\sigma^2| \ = \ o\big(\ts \frac{ \tr(L^2)}{N^2}\big),
\end{equation*}
as needed.

Now we turn to the task of bounding the variance of $\hat\sigma^2$. For each $b=1,\ldots,B$, define the random variable
\begin{equation}\label{eq:xi}
    \xi_b^* \ = \ \|\hat L^*_b -\hat L\|_F^2 -  \E(\|\hat L^* -\hat L\|_F^2|Q).
\end{equation}
It will be helpful to note that $\hat\sigma^2$ can also be expressed as
\begin{equation}\label{eqn:varrep}
    \begin{split}
        \hat\sigma^2\ = \ & \frac{1}{B} \sum_{b=1}^B \big (\xi_b^*\big)^2 - \Big(\frac{1}{B} \sum_{b=1}^B \xi_b^*\Big)^2.
    \end{split}
\end{equation}
due shift invariance.
Using the bound $\var(X+Y)\leq 2\var(X)+2\var(Y)$ for generic random variables $X$ and $Y$, we have
\begin{align*}
    \var(\hat \sigma^2) \ \lesssim \ \ &\var\bigg(\frac{1}{B} \sum_{b=1}^B \big (\xi_b^*\big)^2 \bigg)  \ + \ \var\bigg(\frac{1}{B^2} \sum_{1\leq b\neq b'\leq B} \xi_b^*\xi_{b'}^*\bigg).
    \end{align*}
Using the fact that $\xi_1^*,\dots,\xi_B^*$ are conditionally i.i.d.~given $Q_1, \dots, Q_N$, we apply the law of total variance to each of the terms above, yielding
\begin{align*}
    \var(\hat \sigma^2) 
    \ \lesssim \ &\frac{1}{B} \E\Big(\var\big((\xi_1^*)^2 |Q\big) \Big) + \var\Big(\E\big((\xi_1^*)^2 |Q\big) \Big) + \frac{1}{B^2} \E\Big(\var\big(\xi_1^* \xi_2^* |Q\big) \Big) \\
    \ \lesssim \ &\frac{1}{B} \E\Big(\var\big((\xi_1^*)^2 |Q\big) \Big) + \E\Big(\Big(\E\big((\xi_1^*)^2 |Q\big)\Big)^2\Big) - \Big(\E\big((\xi_1^*)^2 \big)\Big)^2+ \frac{1}{B^2}\E\Big(\Big(\E\big((\xi_1^*)^2 |Q\big)\Big)^2\Big).
\end{align*}
Applying Lemmas \ref{lem:varofF2}, \ref{lem:boundofvarF22} and \ref{lem:boundofcovF22} implies $\var(\hat \sigma^2)= o(\frac{ \tr(L^2)^2}{N^4})$.
\qed

 \subsection{Conditional asymptotic normality of $\|\hat L^*-\hat L\|_F^2$}

Recall that $\hat{\mathsf{d}}\in\R^n$ is defined to contain the diagonal entries of $\hat L$. As explained on page~\pageref{explanation}, the proof of the limit~\eqref{eq:normalapprox1} reduces to the following lemma.

\begin{lemma}
\label{lem:conditionofS}
    If the conditions in Theorem \ref{thm:Frobenius} hold, then as $n\to\infty$,
    \begin{align*}
        \frac{\|\hat{\mathsf{d}}\|_{\infty}}{\|\hat{\mathsf{d}}\|_2} \ \xrightarrow{\P} \ 0.
    \end{align*}
\end{lemma}
\proof
By writing $\frac{\|\hat{\mathsf{d}}\|_{\infty}}{\|\hat{\mathsf{d}}\|_2}=
\frac{\|\hat{\mathsf{d}}\|_{\infty}}{\|\mathsf{d}\|_2}\frac{\|\mathsf{d}\|_{2}}{\|\hat{\mathsf{d}}\|_2}$, it is enough to establish the limits $\|\hat{\mathsf{d}}\|_{\infty}/\|\mathsf{d}\|_2\xrightarrow{\P}0$ and $\|\hat{\mathsf{d}}\|_{2}/\|\mathsf{d}\|_2\xrightarrow{\P}1$. 
With regard to the second limit, note that
\begin{align*}
    \big(\|\hat{\mathsf{d}}\|_2-\|\mathsf{d}\|_2\big)^2 & \ \leq \  \|\hat{\mathsf{d}}-\mathsf{d}\|_2^2
     \ \leq \ \|\hat L- L\|_F^2,
\end{align*}
and so equation \eqref{eq:expectationofF2} gives 
\begin{equation*}
\begin{split}
    \E\Big(\big(\|\hat{\mathsf{d}}\|_2-\|\mathsf{d}\|_2\big)^2\Big) & 
    %
    \ \lesssim \ \frac{1}{N}.
    \end{split}
\end{equation*}
Next, we will show that $\|\mathsf{d}\|_2^2\gtrsim \tr(L^2)$
so that Lemma~\ref{lem:rankofL} will imply
\begin{equation}\label{eqn:ratiolim}
\E\bigg(\Big(\textstyle\frac{\|\hat{\mathsf{d}}\|_2}{\|\mathsf{d}\|_2}-1\Big)^2\bigg) \ \lesssim \ \displaystyle \frac{1}{N \tr(L^2)} \ = \ o(1),
\end{equation}
yielding limit $\|\hat{\mathsf{d}}\|_{2}/\|\mathsf{d}\|_2\xrightarrow{\P}1$. To show the lower bound $\|\mathsf{d}\|_2^2\gtrsim \tr(L^2)$, observe that
\begin{equation*}
    \begin{split}
        \|\mathsf{d}\|_2^2 &= \sum_{i=1}^n \Big(\sum_{j\neq i} L_{i j}\Big)^2
        \geq \sum_{i=1}^n \sum_{j\neq i} L_{i j}^2
         = \tr(L^2)-\|\mathsf{d}\|_2^2,
    \end{split}
\end{equation*}
and so rearranging implies $\|\mathsf{d}\|_2^2\geq \frac{1}{2}\tr(L^2)$.

Now we turn to proving the limit $\|\hat{\mathsf{d}}\|_{\infty}/\|\mathsf{d}\|_2\xrightarrow{\P}0$. Consider the basic inequality
\begin{equation*}
    \frac{\|\hat{\mathsf{d}}\|_{\infty}}{\|\mathsf{d}\|_2} \ \leq \ \frac{\|\mathsf{d}\|_{\infty}}{\|\mathsf{d}\|_2} + \frac{\|\hat{\mathsf{d}}-\mathsf{d}\|_{\infty}}{\|\mathsf{d}\|_2}.
\end{equation*}
The conditions of Theorem~\ref{thm:Frobenius} ensure that the first term on the right is $o(1)$. Meanwhile, the second term is at most $\|\hat{\mathsf{d}}-\mathsf{d}\|_2/\|\mathsf{d}\|_2$, and our earlier work shows that the expectation of this quantity is $\mathcal{O}\big(\sqrt{\E(\|\hat L-L\|_F^2)/\|\mathsf{d}\|_2^2}\big)=\mathcal{O}(1/\sqrt{N\tr(L^2)})=o(1)$. \qed

\subsection{Moments}

\begin{lemma}\label{lem:sigma2}
    If the conditions in Theorem \ref{thm:Frobenius} hold, then 
    \begin{equation*}
            \var(\|\hat L-L\|_F^2) 
            \ = \ \frac{2}{N^2}\Big(\tr(L^2)  +  12\!\!\!\!\!\sum_{1 \leq i<j \leq n} L_{i j} ^2\Big) +o(\ts \frac{\tr(L^2)}{N^2}),
            \end{equation*}
            and in particular
            \begin{equation*}
           \var(\|\hat L-L\|_F^2)  \ \asymp \ \frac{\tr(L^2)}{N^2}.
    \label{eq:orderofF2}
    \end{equation*}
\end{lemma}
\proof
Note that 
\begin{align*}
    \|\hat L-L\|_F^2 \ = \ & \frac{1}{N^2}\sum_{1 \leq i\neq j \leq N}\big(D_{i}\ttop D_{j}\big)^2 \ -  \ \frac{2}{N}\sum_{i=1}^N D_{i}\ttop LD_{i}  \ +  \ \frac{4}{N} \ + \ \tr(L^2). 
\end{align*}
Since the last two terms on the right side are constants, we only need to handle the first two terms.
It follows that
\begin{equation*}
\begin{split}
   \var(\|\hat L-L\|_F^2)
    \ = \ &\frac{4}{N^4} \sum_{1\leq i\neq j\neq k \leq N} \cov\Big(\big(D_{i}\ttop D_{j}\big)^2, \big(D_{i}\ttop D_{k}\big)^2 \Big) \ + \ \frac{2}{N^4} \sum_{1 \leq i\neq j \leq N} \var\Big(\big(D_{i}\ttop D_{j}\big)^2\Big)\\
    & \ \ - \frac{8}{N^3} \sum_{1 \leq i\neq j \leq N} \cov\Big(\big(D_{i}\ttop D_{j}\big)^2,  D_{i}\ttop LD_{i} \Big) \ + \ \frac{4}{N^2} \sum_{i=1}^N \var\big(D_{i}\ttop LD_{i} \big).
\end{split}
\end{equation*}
When $i\neq j\neq k$, it follows from the independence of $D_i$, $D_j$, and $D_k$ that the quantities $\cov\big(\big(D_{i}\ttop D_{j}\big)^2, \big(D_{i}\ttop D_{k}\big)^2 \big)$, $\cov\big(\big(D_{i}\ttop D_{j}\big)^2,  D_{i}\ttop LD_{i} \big)$, and $\var\big(D_{i}\ttop LD_{i} \big)$ are all equal.
Consequently, we can obtain 
\begin{equation}\label{eq:sigma}
\begin{split}
   \var(\|\hat L-L\|_F^2)
    \ = \ & \frac{2(N-1)}{N^3} \var\Big(\big(D_{1}\ttop D_{2}\big)^2\Big) \ + \ \frac{4(-N+2)}{N^3}\cov\Big(\big(D_{1}\ttop D_{2}\big)^2, \big(D_{1}\ttop D_{3}\big)^2 \Big).
\end{split}
\end{equation}
It is direct to calculate
\begin{equation}\label{eq:definitionofa}
    \begin{split}
        \E\Big(\big(D_{1}\ttop D_{2}\big)^2\Big)
        \ = \ &\tr(L^2),\\
        \E\Big(\big(D_{1}\ttop D_{2}\big)^2\big(D_{1}\ttop D_{3}\big)^2\Big)
        \ = \ & \sum_{1 \leq i<j \leq n} |L_{i j}|\big(L_{i i} + L_{jj} + 2|L_{i j}|\big)^2,\\
        \E\Big(\big(D_{1}\ttop D_{2}\big)^4\Big)
        \ = \ &\tr(L^2)  +  12\sum_{1 \leq i<j \leq n} L_{i j} ^2.
    \end{split}
\end{equation}
Based on Lemma \ref{lem:rankofL}, the orders of the above three terms are 
\begin{equation} \label{eq:orderofD}
    \begin{split}
        \bigg( \E\Big(\big(D_{1}\ttop D_{2}\big)^2\Big)\bigg)^2
        \ = \ & o(\tr(L^2)),\\
        \E\Big(\big(D_{1}\ttop D_{2}\big)^2\big(D_{1}\ttop D_{3}\big)^2\Big)
        \ = \ & o(\tr(L^2)),\\
        \E\Big(\big(D_{1}\ttop D_{2}\big)^4\Big)
        \ \asymp \ &\tr(L^2).
    \end{split}
\end{equation}
Applying the above results to \eqref{eq:sigma} completes the proof of the lemma.
\qed

\begin{lemma}\label{lem:varofF2}
    Let $\xi_1^*$ be as defined in \eqref{eq:xi}. If the conditions in Theorem \ref{thm:Frobenius} hold, then
    \begin{align*}
        \var\big(\|\hat L^* -\hat L\|_F^2\big) \ = \  \frac{2}{N^2}\Big(\tr(L^2)  +  12\sum_{1 \leq i<j \leq n} L_{i j} ^2\Big) +o(\ts \frac{\tr(L^2) }{N^2})
    \end{align*}
    and 
    \begin{align*}
        \E\big((\xi_1^*)^2 \big) \ = \  \frac{2}{N^2}\Big(\tr(L^2)  +  12\sum_{1 \leq i<j \leq n} L_{i j} ^2\Big) +o(\ts \frac{\tr(L^2) }{N^2}).
    \end{align*}
\end{lemma}
\proof
We begin with the law of total variance
\begin{align}\label{eq:varF2*}
    \var\big(\|\hat L^* -\hat L\|_F^2\big) \ = \ \E\big(\var(\|\hat L^* -\hat L\|_F^2 |Q)\big)+ \var\big(\E(\|\hat L^* -\hat L\|_F^2 |Q)\big).
\end{align}
For the first term on the right side, following the proof of Lemma \ref{lem:sigma2} yields 
\begin{equation} \label{eq:varofF2}
\begin{split}
    \var(\|\hat L^* -\hat L\|_F^2 |Q) \ = \ &\frac{2(N-1)}{N^3} \bigg(\E\big(\langle D_{1}^* , D_{2}^* \rangle^4 |Q\big)  - \Big(\E\big(\langle D_{1}^* , D_{2}^* \rangle^2 |Q\big)\Big)^2\bigg)\\ &
    + \frac{4(2-N)}{N^3} \bigg(\E\big(\langle D_{1}^* , D_{2}^* \rangle^2 \langle D_{1}^* , D_{3}^* \rangle^2 |Q\big)-\Big(\E\big(\langle D_{1}^* , D_{2}^* \rangle^2 |Q\big)\Big)^2\bigg),
\end{split}
\end{equation}
where
\begin{equation}\label{eq:definitionofa*}
    \begin{split}
        \E\big(\langle D_{1}^* , D_{2}^* \rangle^2 |Q\big)
        \ = \ & \tr(\hat L^2)\\
        \E\big(\langle D_{1}^* , D_{2}^* \rangle^2 \langle D_{1}^* , D_{3}^* \rangle^2 |Q\big)
        \ = \ & \frac{1}{N^3}\sum_{i,j,k=1}^N \big(D_{i}\ttop  D_{j}\big) ^2 \big(D_{i}\ttop  D_{k}\big)^2\\
        \E\big(\langle D_{1}^* , D_{2}^* \rangle^4 |Q\big) \ = \ &\frac{1}{N^2}\sum_{1 \leq i\neq j \leq N} \big(D_{i}\ttop  D_{j}\big) ^4+\frac{16}{N}.
    \end{split}
\end{equation}
To bound $\E\big(\var(\|\hat L^* -\hat L\|_F^2 |Q)\big)$, we will analyze the order of the expectation of the above three terms. Lemmas \ref{lem:rankofL} and \ref{lem:moment} give $\E\Big(\Big(\E\big(\langle D_{1}^* , D_{2}^* \rangle^2 |Q\big)\Big)^2\Big) =o(\tr(L^2) )$. Lemma \ref{lem:rankofL} as well as equations \eqref{eq:definitionofa} and \eqref{eq:orderofD} imply
\begin{equation}\label{eq:expectationofa2a3*}
    \begin{split}
        \E\big(\langle D_{1}^* , D_{2}^* \rangle^2 \langle D_{1}^* , D_{3}^* \rangle^2\big) 
        \ \lesssim \ &  \E\Big( \big(D_{1}\ttop  D_{2}\big) ^2 \big(D_{1}\ttop  D_{3}\big)^2\Big)+ \frac{1}{N} \E\Big( \big(D_{1}\ttop  D_{2}\big) ^4\Big)  +\frac{1}{N} 
        \E\Big( \big(D_{1}\ttop  D_{2}\big)^2\Big) + \frac{1}{N^2}\\
        \ = \ & o(\tr(L^2))\\
        \E\big(\langle D_{1}^* , D_{2}^* \rangle^4\big) 
        \ = \ & \tr(L^2)  +  12\sum_{1 \leq i<j \leq n} L_{i j} ^2 +  o(\tr(L^2) ).
    \end{split}
\end{equation}
Therefore, we have 
\begin{equation}
\label{eq:sdmean}
    \E\big(\var(\|\hat L^* -\hat L\|_F^2 |Q)\big) \ = \  \frac{2}{N^2}\Big(\tr(L^2)  +  12\sum_{1 \leq i<j \leq n} L_{i j} ^2\Big) +o(\ts \frac{\tr(L^2) }{N^2}).
\end{equation}
For the second term on the right side of \eqref{eq:varF2*}, Lemmas \ref{lem:rankofL} and \ref{lem:moment} as well as \eqref{eq:expectaionofF2|S} lead to 
\begin{equation}\label{eq:meansd}
    \begin{split}
        \var\Big(\E(\|\hat L^* -\hat L\|_F^2 |Q)\Big) 
        \ = \ & \frac{1}{N^2} \var( \tr(\hat L^2))\\
        \ = \ & o(\ts \frac{\tr(L^2)}{N^2} ).
    \end{split}
\end{equation}
Combining the above results yields
\begin{align*}
    \var\big(\|\hat L^* -\hat L\|_F^2\big) \ = \  \frac{2}{N^2}\Big(\tr(L^2)  +  12\sum_{1 \leq i<j \leq n} L_{i j} ^2\Big) +o(\ts \frac{\tr(L^2) }{N^2})
\end{align*}
and 
\begin{align*}
    \E\big((\xi_1^*)^2 \big) \ = \ &\var\big(\|\hat L^* -\hat L\|_F^2\big) - \var\Big(\E(\|\hat L^* -\hat L\|_F^2|Q)\Big)\\
    \ = \ &\frac{2}{N^2}\Big(\tr(L^2)  +  12\sum_{1 \leq i<j \leq n} L_{i j} ^2\Big) +o(\ts \frac{\tr(L^2) }{N^2}).
\end{align*}
\qed

\begin{lemma}
\label{lem:boundofvarF22}
    Let $\xi_1^*$ be as defined in \eqref{eq:xi}. If the conditions in Theorem \ref{thm:Frobenius} hold, then
    \begin{align*}
        \E\Big(\var\big((\xi_1^*)^2 |Q\big) \Big) \ \lesssim \ \frac{ \tr(L^2)^2}{N^4}.
    \end{align*}
\end{lemma}

\proof
Letting $V_i^* = \frac{1}{N} ( D_{i}^* (D_{i}^*) \ttop -\hat L)$, equation \eqref{eq:expectaionofF2|S} implies the random variable $\xi_1^*$ can be decomposed as 
\begin{align*}
    \xi_1^* \ = \ &\sum_{i,j = 1}^N \llangle V_i^*,   V_j^*\rrangle- \frac{1}{N}(4- \tr( \hat L^2))
    \\ \ = \ & \gamma^*_1 + \gamma^*_2,
\end{align*}
where 
\begin{align} \label{eq:R}
    \gamma^*_1 \ = \ &\sum_{1 \leq i\neq j \leq N} \llangle V_i^*,   V_j^*\rrangle, \ \ \ \ \ \ \ \ \ \ \ \ \ \ \ \ \ \
    \gamma^*_2 \ = \ -\frac{2}{N^2}\sum_{i=1}^N   \big((D_{i}^*)\ttop \hat L D_{i}^*
    - \tr(\hat L^2)\big).
\end{align}
Note that the following relation holds almost surely,
\begin{equation}\label{er:boundofvar}
    \begin{split}
        \var((\xi_1^*)^2|Q) \ = \ &\var\Big((\gamma_1^*)^2 + (\gamma_2^*)^2 + 2\gamma^*_1 \gamma^*_2|Q\Big)\\
    \ \lesssim \ &\var((\gamma_1^*)^2|Q) + \var((\gamma_2^*)^2|Q)  + \var(\gamma^*_1 \gamma^*_2|Q) . 
    \end{split}
\end{equation}
Since Lemma \ref{lem:R1} proves $\E(\var ((\gamma_1^*)^2|Q)) \lesssim \frac{ \tr(L^2)^2}{N^4}$, it remains to show the expectations of the other terms on the right side are $\mathcal{O}(\frac{ \tr(L^2)^2}{N^4})$.
For $\E(\var ((\gamma_2^*)^2|Q))$, we will bound it through analyzing $\E((\gamma_2^*)^4)$. By noting that $(D_{1}^*)\ttop \hat L D_{1}^* \leq 4$, $\tr(\hat L^2) \leq 4$ and $|\gamma^*_2| \leq \frac{16}{N}$, Lemma \ref{lem:moment} gives
\begin{align*}
    \E((\gamma_2^*)^4) \ \lesssim \ & \frac{1}{N^2} \E\big( \E((\gamma_2^*)^2 |Q)\big)\\
    \ \lesssim \ & \frac{1}{N^6} \sum_{i,j=1}^N \E\Big( \E\Big(\big(   (D_{i}^*)\ttop \hat L D_{i}^* -  \tr(\hat L^2)\big) \big(   (D_{j}^*)\ttop \hat L D_{j}^* -  \tr(\hat L^2)\big) |Q\Big)\Big)\\
    \ \lesssim \ & \frac{1}{N^5} \E\Big( \E\Big(\big(   (D_{1}^*)\ttop \hat L D_{1}^* -  \tr(\hat L^2)\big)^2 |Q\Big)\Big)\\
    \ \lesssim \ & \frac{1}{N^5}\E \big(   (D_{1}^*)\ttop \hat L D_{1}^* +  \tr(\hat L^2) \big)\\
    \ \lesssim \ & \frac{\tr(L^2)}{N^5}.
\end{align*}
Using $N\tr(L^2)\to\infty$ as $n\to\infty$ from Lemma \ref{lem:rankofL}, it follows that 
\begin{equation}\label{eq:boundofvarb2}
    \E(\var ((\gamma_2^*)^2|Q)) \ \lesssim \ \frac{\tr(L^2)^2}{N^4}.
\end{equation}
It remains to analyze $\E(\var(\gamma^*_1\gamma^*_2|Q))$. By noting that 
\begin{align*}
    \gamma^*_1 \gamma^*_2 
    \ = \ -\frac{2}{N^2} \Big(&\sum_{1 \leq i \neq j \neq k \leq N } \llangle V_i^* , V_j^* \rrangle  \big((D_{k}^*)\ttop \hat L D_{k}^*   - \tr(\hat L^2)\big) 
    \\&+2  \sum_{1 \leq i\neq j \leq N} \llangle V_i^* , V_j^* \rrangle  \big((D_{i}^*)\ttop \hat L D_{i}^*   - \tr(\hat L^2)\big)\Big),
\end{align*}
the fact that $(D_{1}^*)\ttop \hat L D_{1}^* \leq 4$ and $\tr(\hat L^2) \leq 4$ hold almost surely implies
\begin{equation*}
    \begin{split}
        \var(\gamma^*_1\gamma^*_2 |Q)
        \ \lesssim \ & \frac{1}{N^4}\var\Big( \sum_{1 \leq i \neq j \neq k \leq N} \llangle V_i^* , V_j^* \rrangle  \big((D_{k}^*)\ttop \hat L D_{k}^*   - \tr(\hat L^2)\big) |Q\Big)\\
        & + \frac{1}{N^4} \var\big( \sum_{1 \leq i\neq j \leq N} \llangle V_i^* , V_j^* \rrangle  \Big((D_{i}^*)\ttop \hat L D_{i}^* - \tr(\hat L^2)\big)|Q\Big)\\
        \ \lesssim \ &\frac{1}{N}  \E\Big(\llangle V_1^* , V_2^* \rrangle ^2\Big( \big((D_{3}^*)\ttop \hat L D_{3}^*   - \tr(\hat L^2)\big)^2  + \big((D_{1}^*)\ttop \hat L D_{1}^*   - \tr(\hat L^2)\big)^2\Big) |Q\Big)\\
        \ \lesssim \ &\frac{1}{N} \E\Big(\llangle V_1^* , V_2^* \rrangle ^2  |Q\Big).
    \end{split}
\end{equation*}
Combining with the bound for $\E\big(\llangle V_1^* , V_2^* \rrangle ^2\big)$ given in \eqref{eq:boundoftr2} yields
\begin{equation}\label{eq:boundofvarb1b2}
    \E(\var(\gamma^*_1\gamma^*_2|Q))  \ \lesssim \  \frac{\tr(L^2)^2}{N^4}.
\end{equation}
Applying equations \eqref{eq:boundofvarb2} and \eqref{eq:boundofvarb1b2} to \eqref{er:boundofvar} yields the stated result.
\qed

\begin{lemma}
    \label{lem:R1}
    Let $\gamma^*_1$ be as defined in \eqref{eq:R}. If the conditions in Theorem \ref{thm:Frobenius} hold, then 
    \begin{align*}
        \E(\var ((\gamma_1^*)^2|Q)) \ \lesssim \ \frac{ \tr(L^2)^2}{N^4}.
    \end{align*}
\end{lemma}
\proof
To analyze $\E(\var ((\gamma_1^*)^2|Q))$, note that
\begin{align*}
    (\gamma_1^*)^2 
    \ = \ &\sum_{1 \leq i\neq j  \neq k \neq l \leq N} \llangle V_i^* , V_j^* \rrangle\llangle V_k^* , V_l^* \rrangle + 4\sum_{1 \leq i \neq j \neq k \leq N } \llangle V_i^* , V_j^* \rrangle\llangle V_i^* , V_k^* \rrangle  +2 \sum_{1 \leq i\neq j \leq N} \llangle V_i^* , V_j^* \rrangle^2\\
    \ =: \ &\gamma^*_{11} + 4 \gamma^*_{12} + 2 \gamma^*_{13},
\end{align*}
which implies
\begin{equation}\label{eq:varb1}
    \E(\var ((\gamma_1^*)^2|Q)) \ \lesssim \ \E(\var(\gamma^*_{11} |Q)) + \E(\var(\gamma^*_{12} |Q)) + \E(\var(\gamma^*_{13} |Q)).
\end{equation}
We know
\begin{equation*}
    \begin{split}
        \var(\gamma^*_{11} |Q) \ = \ & \sum_{1\leq i\neq j\neq k \neq l \leq N } 8 \var\Big( \llangle V_i^* , V_j^* \rrangle\llangle V_k^* , V_l^* \rrangle |Q\Big)\\ &+\sum_{1\leq i\neq j\neq k \neq l \leq N } 16 \cov\Big( \llangle V_i^* , V_j^* \rrangle\llangle V_k^* , V_l^* \rrangle,\llangle V_i^* , V_k^* \rrangle \llangle V_j^* , V_l^* \rrangle |Q\Big) \\
        \ \leq \ & 24 N^4 \E\Big( \llangle V_1^* , V_2^* \rrangle^2 \llangle V_3^* , V_4^* \rrangle ^2 |Q\Big) \\
        \ = \ & 24 N^4 \bigg(\E\Big( \llangle V_1^* , V_2^* \rrangle^2 |Q\Big)\bigg)^2.
    \end{split}
\end{equation*}
To bound $\E\Big( \llangle V_1^* , V_2^* \rrangle^2 |Q\Big)$, note that the following inequalities hold almost surely for all $i,j \in \{1,\ldots,N\}$:
\begin{equation*}
    \begin{split}
          \big( (D_{i}^*)\ttop D_{j}^*\big)^2 & \ \leq \ 4,\\
          (D_{i}^*)\ttop  \hat L D_{i}^* & \ \leq \ 4,\\
          \tr(\hat L^2) & \ \leq \ 4 .
    \end{split}
\end{equation*}
Hence, for any $i,j \in \{1,\ldots,N\}$, we have
\begin{equation} 
\label{eq:boundoftr}
    \begin{split}
        |\llangle V_i^* , V_j^* \rrangle| \ \leq \ &\frac{1}{N^2} \Big( \langle D_{i}^* , D_{j}^*\rangle^2  + (D_{i}^*)\ttop \hat L D_{i}^*  + (D_{j}^*)\ttop \hat L  D_{j}^* +\tr(\hat L^2) \Big)\\ \ \leq \ &  \frac{16}{N^2}
    \end{split}
\end{equation}
and
\begin{equation}\label{eq:expectationoftr}
    \begin{split}
        \E\big( |\llangle V_1^* , V_2^* \rrangle|  |Q\big)
        \ \leq \ & \frac{4}{N^2} \tr(\hat L^2)
    \end{split}
\end{equation}
hold almost surely. Combining above results with Lemma \ref{lem:moment} gives
\begin{equation}\label{eq:Evarb11}
    \begin{split}
        \E(\var(\gamma^*_{11} |Q)) \ \lesssim \ & N^4 \E\Big( \frac{\tr(\hat L^2)^2}{N^8}\Big) \\ \ \lesssim \ &\frac{\tr(L^2)^2}{N^4}.
    \end{split}
\end{equation}
To analyze $\E(\var(\gamma^*_{12} |Q))$, the definition of $\gamma^*_{12}$ gives
\begin{equation*}
    \begin{split}
        \var(\gamma^*_{12} |Q) \ = \ &\sum_{1\leq i\neq j\neq  k \neq l \leq N} 2 \cov\Big( \llangle V_i^* , V_j^* \rrangle\llangle V_i^* , V_k^* \rrangle,\llangle V_j^* , V_l^* \rrangle\llangle V_k^* , V_l^* \rrangle |Q\Big)  \\ &+\sum_{1\leq i\neq j\neq k \leq N} 2 \var\Big( \llangle V_i^* , V_j^* \rrangle\llangle V_i^* , V_k^* \rrangle |Q\Big) \\&+\sum_{1\leq i\neq j\neq k \leq N} 4 \cov\Big( \llangle V_i^* , V_j^* \rrangle\llangle V_i^* , V_k^* \rrangle,\llangle V_i^* , V_k^* \rrangle\llangle V_j^* , V_k^* \rrangle |Q\Big) \\
        \ \leq \ & 2 N^4 \E\Big(  \llangle V_1^* , V_2^* \rrangle \llangle V_1^* , V_3^* \rrangle \llangle V_2^* , V_4^* \rrangle \llangle V_3^* , V_4^* \rrangle  |Q\Big)  \\ &+ 6 N^3\E\Big(  \llangle V_1^* , V_2^* \rrangle^2 \llangle V_1^* , V_3^* \rrangle^2  |Q\Big) .
    \end{split}
\end{equation*}
Lemma \ref{lem:moment}, \eqref{eq:boundoftr} and \eqref{eq:expectationoftr} imply
\begin{align*}
    N^4 \Big|\E\Big( \llangle V_1^* , V_2^* \rrangle \llangle V_1^* , V_3^* \rrangle \llangle V_2^* , V_4^* \rrangle \llangle V_3^* , V_4^* \rrangle \Big)\Big|  \ \lesssim \ & \E\Big(\E\big(  |\llangle V_1^* , V_2^* \rrangle|  |Q\big)\E\big(  |\llangle V_3^* , V_4^* \rrangle|  |Q\big)\Big) \\ 
    \ \lesssim \ & \frac{1}{N^4} \E\big( \tr(\hat L^2)^2 \big) \\ \ \lesssim \ & \frac{\tr(L^2)^2}{N^4}.
\end{align*}
Since Lemmas \ref{lem:rankofL} and \ref{lem:moment} as well as \eqref{eq:expectationofa2a3*} and \eqref{eq:boundoftr} imply
\begin{equation}\label{eq:boundoftr2}
    \begin{split}
        \E\Big(  \llangle V_1^* , V_2^* \rrangle^4  \Big)   \ \lesssim \ &     \frac{1}{N^4} \E\Big(  \llangle V_1^* , V_2^* \rrangle^2 \Big) 
        \\ \ \lesssim \ &   \frac{1}{N^8}\E\Big( \langle D_{1}^* , D_{2}^* \rangle^4  +  ((D_{1}^*)\ttop \hat L D_{1}^*)^2  + \tr(\hat L^2)^2   \Big) 
        \\ \ \lesssim \ &   \frac{\tr(L^2)^2}{N^7},
    \end{split}
\end{equation}
we have
\begin{align*}
     N^3 \E\Big(  \llangle V_1^* , V_2^* \rrangle^2 \llangle V_1^* , V_3^* \rrangle^2 \Big)
    \ \lesssim \ &\frac{\tr(L^2)^2}{N^4}.
\end{align*}
Combining the above results yields
\begin{equation}\label{eq:Evarb12}
    \E(\var(\gamma^*_{12} |Q)) 
    \ \lesssim \ \frac{\tr(L^2)^2}{N^4}.
\end{equation}
To analyze $ \E(\var(\gamma^*_{13} |Q))$, note that 
\begin{equation*}\label{eq:varb13}
    \begin{split}
        \var(\gamma^*_{13} |Q) \ = \ &\sum_{1\leq i\neq j\neq k \leq N} 4 \cov\Big( \llangle V_i^* , V_j^* \rrangle^2,\llangle V_i^* , V_k^* \rrangle^2 |Q\Big)  +\sum_{1 \leq i\neq j \leq N} 2 \var\Big( \llangle V_i^* , V_j^* \rrangle^2 |Q\Big)\\
        \ \leq \ & 6 N^3 \E\Big(  \llangle V_1^* , V_2^* \rrangle^4  |Q\Big).
    \end{split}
\end{equation*}
Equation \eqref{eq:boundoftr2} gives
\begin{equation}\label{eq:Evarb13}
    \E(\var(\gamma^*_{13} |Q)) \ \lesssim \ \frac{\tr(L^2)^2}{N^4}.
\end{equation}
Applying \eqref{eq:Evarb11}, \eqref{eq:Evarb12} and \eqref{eq:Evarb13} to \eqref{eq:varb1} completes the proof.
\qed

\begin{lemma}\label{lem:boundofcovF22}
    Let $\xi_1^*$ be as defined in \eqref{eq:xi}. If the conditions in Theorem \ref{thm:Frobenius} hold, then 
    \begin{align*}
        \E\Big(\Big(\E\big((\xi_1^*)^2 |Q\big)\Big)^2\Big) \ = \ &\frac{4}{N^4}\big(\tr(L^2)  +  12\sum_{1 \leq i<j \leq n} L_{i j} ^2\big)^2 +  o( \ts \frac{ \tr(L^2)^2}{N^4})\\
            \ \asymp \ &\frac{\tr(L^2)^2}{N^4}.
    \end{align*}
\end{lemma}
\proof
Note that the definition of $\xi_1^*$ and equation \eqref{eq:varofF2} give
\begin{equation*}
    \begin{split}
        \E\big((\xi_1^*)^2 |Q\big)
        \ = \ & \var\big(\|\hat L^* -\hat L\|_F^2 |Q\big)  \\
        \ = \ & \frac{2(N-1)}{N^5}  \sum_{ i, j =1 }^N \big(D_{i}\ttop  D_{j}\big) ^4 + \frac{4(2-N)}{N^6} \sum_{i,j,k=1}^N \big(D_{i}\ttop  D_{j}\big) ^2 \big(D_{i}\ttop  D_{k}\big)^2 + \frac{2 N-6}{N^3}\tr(\hat L^2)^2.
    \end{split}
\end{equation*}
If we can show 
\begin{align*}
    &\E\Big(\big(\sum_{ i, j =1 }^N \big(D_{i}\ttop  D_{j}\big) ^4\big)^2\Big)  \ = \  N^4\big(\tr(L^2)  +  12\sum_{1 \leq i<j \leq n} L_{i j} ^2\big)^2 +  o( N^4\tr(L^2)^2)\\
    &\E\Big(\big(\sum_{i,j,k=1}^N \big(D_{i}\ttop  D_{j}\big) ^2 \big(D_{i}\ttop  D_{k}\big)^2\big)^2\Big)  \ = \ o( N^6 \tr(L^2)^2)\\
    &\E\big(\tr(\hat L^2)^4\big)  \ = \ o( \tr(L^2)^2),
\end{align*}
applying Hölder's inequality completes the proof of the lemma. Since the last statement is implied by Lemmas \ref{lem:rankofL} and \ref{lem:moment}, we will show the first two equalities. Combining the fact $|D_{i}\ttop  D_{j} |\leq 2$ with Lemma \ref{lem:rankofL}, \eqref{eq:definitionofa} and \eqref{eq:orderofD}  implies  
\begin{equation*}\label{eq:expectationofa3*}
    \begin{split}
        \E\bigg(\Big(\sum_{ i, j =1 }^N \big(D_{i}\ttop  D_{j}\big)^4\Big)^2\bigg)   \ = \ &\E\Big(\big(\sum_{ 1 \leq i\neq j \leq N } \big(D_{i}\ttop  D_{j}\big) ^4\big)^2\Big) + 32 N \E\Big(\sum_{ 1 \leq i\neq j \leq N } \big(D_{i}\ttop  D_{j}\big) ^4\Big) +216N^2\\
        \ = \ &\E\Big(\sum_{1 \leq i \neq j \neq k \neq l \leq N} \big(D_{i}\ttop  D_{j}\big) ^4 \big(D_{k}\ttop  D_{l}\big) ^4\Big)   +2 \E\Big(\sum_{1 \leq i\neq j \leq N} \big(D_{i}\ttop  D_{j}\big) ^8\Big) \\&+4 \E\Big( \sum_{1 \leq i\neq j\neq k \leq N} \big(D_{i}\ttop  D_{j}\big) ^4 \big(D_{i}\ttop  D_{k}\big) ^4\Big)+ o( N^4\tr(L^2)^2)\\
        \ = \ &  N^4\big(\tr(L^2)  +  12\sum_{1 \leq i<j \leq n} L_{i j} ^2\big)^2  + o( N^4\tr(L^2)^2)
    \end{split}
\end{equation*}
and 
\begin{align*}
    \E\bigg(\Big(\sum_{i,j,k=1}^N \big(D_{i}\ttop  D_{j}\big) ^2 \big(D_{i}\ttop  D_{k}\big)^2\Big)^2\bigg) 
    \ \leq \ &\E\bigg(\Big(\mathcal{O}(N^2)  + \sum_{1 \leq i \neq j \neq k \leq N} \big(D_{i}\ttop  D_{j}\big) ^2 \big(D_{i}\ttop  D_{k}\big)^2\Big)^2\bigg)\\
    \ \lesssim \ & \sum_{1 \leq i \neq j \neq k  \neq  l \neq r \neq s \leq N}  \big(D_{i}\ttop  D_{j}\big) ^2 \big(D_{i}\ttop  D_{k}\big) ^2 \big(D_{s}\ttop  D_{l}\big) ^2 \big(D_{s}\ttop  D_{r}\big)^2\Big)\\
    &+ N^5\E\Big(\big(D_{1}\ttop  D_{2}\big) ^4 \big(D_{1}\ttop  D_{3}\big) ^4\Big) + N^4\\
    \ = \ & o(N^6\tr(L^2)^2).
\end{align*}
\qed

\begin{lemma}
\label{lem:moment}
If the conditions in Theorem \ref{thm:Frobenius} hold, then for any fixed integer $l\geq 1$ not depending on $n$, we have
    \begin{align*}
        \E(\|\hat L\|_F^{2l}) \ \lesssim \ &\|L\|_{F}^{2l}.
    \end{align*}
\end{lemma}
\proof
For $l = 1$, Lemma \ref{lem:rankofL} and \eqref{eq:expectationoftrL2} imply $\E(\|\hat L\|_F^{2}) \lesssim \|L\|_{F}^{2}$.
We will use strong induction to complete the proof. Let $l > 1$ and assume $\E(\|\hat L\|_F^{2k}) \lesssim \|L\|_{F}^{2k}$ holds for all integer $0\leq k <l$. By noting that 
\begin{align*}
    \E(\|\hat L\|_F^{2l}) \ = \ &\frac{1}{N^{2l}}  \sum_{i_1,\ldots,i_l,j_1,\ldots,j_l = 1}^N \E\Big( \Pi_{h=1}^l\big(D_{i_h}\ttop D_{j_h}\big)^2\Big)
\end{align*}
and $\E\big( \Pi_{h=1}^l\big(D_{i_h}\ttop D_{j_h}\big)^2\big) = \|L\|_{F}^{2l}$ for $1 \leq i_1\neq \ldots\neq i_l\neq j_1\neq \ldots\neq j_l \leq N$, we only need to show that the summation over  $i_1,\ldots,i_l,j_1,\ldots,j_l$ which are not all distinct can be bounded by $N^{2l}\|L\|_{F}^{2l}$. Due to $|D_{i}\ttop D_{j}|  \leq 2$ for any $i,j =1,\ldots N$, if there are terms involving a same index, we will only keep one of them and use the bound $|D_{i}\ttop D_{j}|  \leq 2$ for the others. Finally, there are $d$ pairs remaining and the others are changed to be a constant, where $0 \leq d < l$. Without loss of generality, we consider the first $d$ pairs to all be distinct, while the remaining pairs, which involve at least one index with repetition, are changed to 2. Combining with Lemma \ref{lem:rankofL} gives
\begin{align*}
     \sum_{|\{i_1,\ldots,j_l\}| < 2l} \E\Big( \Pi_{h=1}^l\big(D_{i_h}\ttop D_{j_h}\big)^2\Big) \ \lesssim \ & \sum_{d=0}^{l-1} \bigg(\sum_{1\leq i_1 \neq  \ldots \neq j_d \leq N}  N^{l-d}\E\Big( \Pi_{h=1}^d\big(D_{i_h}\ttop D_{j_h}\big)^2\Big)\bigg)\\
    \ \lesssim \ &\sum_{d=0}^{l-1} N^{l+d} \E(\|\hat L\|_F^{2d})  \\
    \ \lesssim \ & N^{2l}\|L\|_{F}^{2l},
\end{align*}
which completes the proof.
\qed

\section{Background results} 
\label{sec:background}
\setcounter{lemma}{0}
\renewcommand{\thelemma}{G.\arabic{lemma}}

\begin{lemma}[\cite{chernozhuokov2022improved}, Theorem 2.1]
\label{lem:CCK}
    Let $X_1,\dots,X_n$ be independent random vectors in $\mathbb{R}^p$ such that $\E(X_{ij}) = 0$ for all $i =1,\ldots n$ and $j = 1, \ldots, p$. Let $(G_1,\ldots,G_p)$ be a Gaussian random vector in $\mathbb{R}^p$ with mean 0 and covariance matrix $\frac{1}{n}\sum_{i=1}^n \E(X_i X_i\ttop)$, and define $M(G) = \max_{1\leq j \leq p} |G_i|$. Let $b_1$ and $b_2$ be some strictly positive constants such that $b_1 \leq b_2$ and let $\{c_n\}_{n\geq 1}$ be a sequence of constants such that $c_n \geq 1$. If for all $i=1,\ldots, n$ and $j = 1, \ldots, p$, we have 
    \begin{align*}
        \E\Big(\exp\big(\ts \frac{|X_{ij}|}{c_n}\big)\Big) \leq 2, \ \ \ \ \ \ \ \ \  b_1^2 \leq \frac{1}{n} \sum_{i=1}^n \E(X_{ij}^2)  \ \ \ \ \ \ \ \ \  \text{ and }  \ \ \ \ \ \ \ \ \  \frac{1}{n} \sum_{i=1}^n \E(X_{ij}^4) \leq  b_2^2 c_n^2,
    \end{align*}
    then 
    \begin{align*}
        \sup_{t\in\R}\bigg|\P\Big( \max_{1\leq j \leq p} \big|\ts \frac{1}{\sqrt{n}} \sum_{i=1}^n X_{ij}\big|\leq t\Big)-\P\big(M( G)\leq t\big)\bigg| \ \leq  \ \kappa \bigg(\displaystyle\frac{c_n^2 \log(2pn)^5}{n}\bigg)^{1/4},
    \end{align*}
    where $\kappa$ is a constant depending only on $b_1$ and $b_2$.
\end{lemma}

\begin{lemma}[\cite{chernozhuokov2022improved}, Proposition 2.1]
\label{lem:gaussianappro}
    Let $V$ and $W$ be mean-zero Gaussian vectors in $\R^p$ with respective covariance matrices $\Sigma$ and $\tilde \Sigma$. Also, assume that $\min_{1\leq j\leq p}\Sigma_{jj} \geq \varsigma$ for some positive constant $\varsigma$. Then,
    \begin{align*}
        \sup_{t \in \mathbb{R}}\bigg|\P\Big(\max_{1\leq j\leq p}|V_j| \leq t\Big)-\P\Big(\max_{1\leq j\leq p} |W_j| \leq t \Big)\bigg| \ \leq \ C\log(2p) \|\Sigma - \tilde \Sigma\|_{\infty}^{1/2}
    \end{align*}
    where $C$ is a positive constant depending only on $\varsigma$.
\end{lemma}

\begin{lemma}[\cite{nazarov2003maximal}, Nazarov’s inequality]
\label{lem:Nazarov}
    Let $\epsilon>0$, and let $V$ be a mean-zero Gaussian random vector in $\mathbb{R}^p$ with $\E(V_j^2) = 1$ for all $j = 1, \ldots, p$. Then,
    \begin{align*}
        \sup_{r \in \mathbb{R} } \P\big(|\max_{1 \leq j \leq p} |V_j| -r| \leq \epsilon \big) \  \leq \ 2 \epsilon\Big(\sqrt{2 \log(2p)}+2\Big).
    \end{align*}
\end{lemma}

\begin{lemma}
\label{lem:anticon}
     If $V$ and $W$ are random variables, then for any $\delta > 0$,
    \begin{align*}
        \sup_{t \in \mathbb R} \P(|V-t|\leq \delta) \  \leq \  \sup_{t \in \mathbb R} \P(|W-t|\leq \delta) +2d_{\mathrm{K}}(\mathcal{L}(V),\mathcal{L}(W)).
    \end{align*}
\end{lemma}

\begin{lemma}[\cite{huang2023high}, Proposition 9]
\label{lem:ustatistic}
Let $X_1,\dots,X_n$ be i.i.d.~random vectors in $\mathbb{R}^p$, and let $h$ be a function $h: \mathbb{R}^p \times \mathbb{R}^p \to \mathbb{R}$ satisfying $\E(h(X_1,X_2)|X_1) = 0$. Suppose there is a sequence of functions $\phi_1,\dots,\phi_K:\mathbb{R}^p\to\mathbb{R}$ such that $h$ can be represented as 
\begin{equation*}
h(x,x') = \sum_{k=1}^{K} \phi_k(x)\phi_k(x')
\end{equation*}
for all $x,x'\in\mathbb{R}^p$. Also, let $\phi(X_1) = (\phi_1(X_1),\ldots,\phi_{K}(X_1))$ and let $\Sigma\in \mathbb{R}^{K\times K}$ denote the covariance matrix of $\phi(X_1)$. Lastly, let $\tau^2=\var(h(X_1,X_2))$, and let $Z_1,\dots,Z_K$ denote independent standard normal random variables.
Then, 
\begin{align*}
   \sup_{t\in\R}\Bigg|\P\bigg( \sum_{1\leq i \neq j \leq n} \frac{h(X_i,X_j)}{\sqrt{\tau^2n(n-1)}}\leq t\bigg) \,-\,\P\bigg( \sum_{k=1}^K \textstyle\frac{1}{\tau}\lambda_k(\Sigma)&(Z_k^2-1)\leq t\bigg)\Bigg| 
      \ \lesssim \  n^{-\frac{1}{5}} + n^{-\frac{1}{14}} \Big(\ \ts\frac{1}{\tau}\|h(X_1, X_2)\|_{L^{3}}\Big)^{\frac{3}{7}}.
\end{align*}
\end{lemma}

\end{document}